\documentclass{article}

\usepackage{jmlr2e-arxiv} 

\usepackage{algorithm}
\usepackage{algorithmic}
\usepackage{bbm}

\usepackage{rotating}
\usepackage{rotfloat}
\usepackage{graphicx}  
\usepackage{subfigure} 
\usepackage{multirow}

\usepackage{bbold} 
\usepackage{hyperref}

\usepackage{amsmath}
\usepackage{amssymb} 
\usepackage{color}
\usepackage{esint}
\usepackage{verbatim}

\usepackage[utf8]{inputenc}
\usepackage[T1]{fontenc}

\newcommand{\mmp}[1]{\textcolor{red}{\em MMP: {#1}}}

\renewcommand{\comment}[1]{}

\newcommand{\riememb}{{\sc LearnMetric}}

\newcommand{\computegraph}{{\sc GraphLaplacian}}
\newcommand{\myemb}{{\sc GenericEmbed}} 
\newcommand{\pseudoinv}{{\sc PseudoInverse}}

\newcommand{\beq}{\begin{equation}}
\newcommand{\eeq}{\end{equation}}
\newcommand{\beqa}{\begin{eqnarray}}
\newcommand{\eeqa}{\end{eqnarray}}
\newcommand{\beqas}{\begin{eqnarray*}}
\newcommand{\eeqas}{\end{eqnarray*}}

\newcommand{\bit}{\begin{itemize}}
\newcommand{\eit}{\end{itemize}}
\newcommand{\bits}{\begin{itemize*}}
\newcommand{\eits}{\end{itemize*}}
\newcommand{\benum}{\begin{enumerate}}
\newcommand{\eenum}{\end{enumerate}}
\newcommand{\benums}{\begin{enumerate*}}
\newcommand{\eenums}{\end{enumerate*}}

\newcommand{\nsamp}{n}
\newcommand{\bw}{\sqrt{\epsilon}} 
\newcommand{\epps}{\epsilon}

\newcommand{\lam}{\lambda}

\newcommand{\1}{\mathbf{1}}

\newcommand{\G}{\mathcal{G}}
\newcommand{\M}{\mathcal{M}}

\newcommand{\N}{{\mathcal N}}
\newcommand{\Ln}{\tilde{\mathcal{L}}_{\epps_\nsamp,\nsamp}}
\newcommand{\Lbw}{\tilde{\mathcal{L}}_{\epps,\nsamp}}
\newcommand{\Lbwnot}{\mathcal{L}_{\epps,\nsamp}} 
\newcommand{\LB}{\Delta_{\M}}

\newcommand{\dataset}{\mathcal{P}}

\newcommand{\mbx}{x} 
\newcommand{\mby}{y} 

\newcommand{\rrr}{{\mathbb{R}}}

\newcommand{\density}{\pi}

\newcommand{\fracpartial}[2]{\frac{\partial {#1}}{\partial {#2}}}
\newcommand{\bigOO}{\mathcal{O}}

\newcommand{\diag}[1]{{\rm diag}\{\,#1\}}
\newcommand{\dembed}{s}
\newcommand{\dintri}{d}  
\newcommand{\dhigh}{r}

\newcommand{\ihigh}{\delta_{\dhigh}}  
\newcommand{\iembed}{\delta_{\dembed}}  

\newcommand{\matrh}{{\mathbf H}}

\newcommand{\myprod}{\cdot} 

\newlength{\picwi}  

\newlength{\backitem}
\newenvironment{boxit}{\begin{lrbox}{\savestuff}
	\begin{minipage}[b]{\linewidth}}
{\end{minipage}\end{lrbox}\fbox{\usebox{\savestuff}}}
\newsavebox{\savestuff}
\comment{

\newcommand{\BlackBox}{\rule{1.5ex}{1.5ex}}  
\newenvironment{proof}{\par\noindent{\bf Proof\ }}{\hfill\BlackBox\\[2mm]}
 
\newtheorem{theorem}{Theorem}
 
\newtheorem{proposition}[theorem]{Proposition}

\newtheorem{definition}[theorem]{Definition}

}

\ShortHeadings{}{Perrault-Joncas Meil\u{a}} 
\firstpageno{1}

\begin{document}
 

\title{Non-linear Dimensionality Reduction: Riemannian Metric Estimation and the
Problem of Geometric Recovery}

\author{\name Dominique Perrault-Joncas \email joncas@amazon.com \\
       Amazon.com
       \AND
       \name Marina Meil\u{a} \email mmp@stat.washington.edu \\
       \addr Department of Statistics\\
       University of Washington\\
       Seattle, WA 98195-4322, USA}



\maketitle

\begin{abstract}
In recent years, manifold learning has become increasingly popular as a tool for performing non-linear dimensionality reduction. This has led to the development of numerous algorithms of varying degrees of complexity that aim to recover manifold geometry using either local or global features of the data. 

Building on the Laplacian Eigenmap and Diffusionmaps framework, we propose a new paradigm that offers a guarantee, under reasonable assumptions, that any manifold learning algorithm will preserve the geometry of a data set. Our approach is based on augmenting the output of embedding algorithms with geometric information embodied in the Riemannian metric of the manifold. We provide an algorithm for estimating the Riemannian metric from data and demonstrate possible applications of our approach in a variety of examples.
\end{abstract}

\comment{The ability to determine when an embedding algorithm preserves the geometry of the data is of practical and theoretical importance in manifold learning. In this paper, we propose a new paradigm that offers a guarantee, under reasonable assumptions, that a manifold learning algorithm preserves the geometry of a data set. Our approach is based on augmenting the ouput of embedding algorithms with geometric information embodied in the {\em Riemannian metric} of the manifold. The Riemannian metric allows one to compute geometric quantities of interest (such as angles, path lengths, or volume)  on a manifold in any coordinate system. We provide an algorithm for estimating the metric from data, consider its consistency, and demonstrate the advantages of our approach in a variety of examples.} 


\section{Introduction}
\label{sec:intro}

When working with large sets of high-dimensional data, one is regularly confronted with the problem of tractability and interpretability of the data. An appealing approach to this problem is the method of dimensionality reduction: finding a low-dimensional representation of the data that preserves all or most of the important ``information''. One popular idea for Euclidean data is to appeal to the manifold hypothesis, whereby the data is assumed to lie on a low-dimensional smooth manifold embedded in the high dimensional space. The task then becomes to recover the low-dimensional manifold so as to perform any statistical analysis on the lower dimensional representation of the data. 

A common technique for performing dimensionality reduction is Principal Component Analysis, which assumes that the low-dimenisional manifold is an affine space. The affine space requirement is generally violated in practice and this has led to the development of more general techniques which perform non-linear dimensionality reduction. Although not all non-linear dimensionality reduction techniques are based on the manifold hypothesis, manifold learning has been a very popular approach to the problem. This is in large part due to the easy interpretability and mathematical elegance of the manifold hypothesis.   

The popularity of manifold learning has led to the development 
of numerous algorithms that aim to recover the geometry of the low-dimensional
manifold $\mathcal{M}$ using either local or global features of the
data. These algorithms are of varying degrees of complexity, but all
have important shortcomings that have been documented
in the literature~\citep{GolZakKusRit08,Wit05}.  Two important criticisms are that 
1) the algorithms fail to recover the geometry of the manifold in many instances and 2) no coherent framework yet exists in which the multitude of existing algorithms can easily be compared and selected for a given
application. 

It is customary to evaluate embedding algorithms by how well they
``recover the geometry'', i.e. preserve the important information of
the data manifold, and much effort has been devoted to finding
embedding algorithms that do so. While there is no uniformly accepted
definition of what it means to ``recover the geometry'' of the data,
we give this criterion a mathematical interpretation, using the
concepts of {\em Riemannian metric} and {\em isometry}. The criticisms
noted above reflect the fact that the majority of manifold learning
algorithms output embeddings that are not isometric to the original
data except in special cases.

Assuming that recovering the geometry of the data is an important goal, we offer a new perspective: rather than contributing yet another embedding algorithm that strives to achieve isometry, we provide a way to augment {\em any} reasonable embedding so as to allow for the correct computation of geometric values of interest in the embedding's own coordinates.

The information necessary for reconstructing the geometry of the
manifold is embodied in its Riemannian metric, defined in Section
\ref{sec:g-math}. We propose to recover a {\em Riemannian manifold}
$(\M,g)$ from the data, that is, a manifold and its
Riemannian metric $g$, and express $g$ in any desired coordinate
system. Practically, for any given mapping produced by an existing
manifold learning algorithm, we will add an estimation of the
Riemannian metric $g$ in the new data coordinates, that makes the
geometrical quantities like distances and angles of the mapped data
(approximately) equal to their original values, in the raw data.
\comment{
Our paradigm is illustrated schematically by Figure~\ref{fig:flowchart}.
}

We start with a brief discussion of the literature and an introduction
to the Riemannian metric in Sections~\ref{sec:Prob} and
~\ref{sec:Metric}.  The core of our paper is the demonstration of how to obtain the Riemannian metric from the mathematical, algorithmic and statistical points of view. These are presented in
Sections~\ref{sec:g-math} and \ref{sec:Algo}. Finally, we offer some examples and applications in
Section~\ref{sec:App} and conclude with a discussion in Section \ref{sec:discussion}.

\section{The Task of Manifold Learning}
\label{sec:Prob}


In this section, we present the problem of manifold learning. We focus on formulating coherently and explicitly a two properties
that cause a manifold learning algorithm to ``work well'', or have intuitively
desirable properties.

The first desirable property is that the algorithm produces a {\em smooth}
map, and Section \ref{sec:Metric} defines this concept in differential geometry
terms. This property is common to a large number of algorithms, so it
will be treated as an assumption in later sections. 

\comment{
The second desirable property is {\em consistency}. Existing consistency results will be discussed briefly in Section \ref{sub:quick-cons} below, and then a more detailed framework will be presented in Section
\ref{sec:Consistency}. 
}

The second property is the preservation of the
{\em intrinsic geometry} of the manifold. This property is of central interest to this article. 

We begin our survey of manifold learning algorithms by discussing a well-known method for linear dimensionality reduction: Principal Component Analysis. PCA is a simple but very powerful technique that projects data onto a linear space of a fixed dimension that explains the highest proportion of variability in the data. It does so by performing an eigendecomposition of the data correlation matrix and selecting the eigenvectors with the largest eigenvalues, i.e. those that explain the most variation. Since the projection of the data is linear by construction, PCA cannot recover any curvature present in the data. 

In contrast to linear techniques, manifold learning algorithms assume that the data lies near or along a non-linear, smooth, submanifold of dimension $\dintri$ called the {\em data manifold} $\M$, embedded in the
original high-dimensional space $\rrr^\dhigh$ with $\dintri \ll \dhigh$, and attempt to uncover this low-dimensional ${\M}$. If they succeed in doing so, then each high-dimensional observation can accurately be described by a small number of parameters, its {\em embedding coordinates} $f(p)$ for all $p\in\M$.

Thus, generally speaking, a {\em manifold learning} or {\em manifold
embedding} algorithm is a method of non-linear 
dimension reduction. Its input is a set of points $\dataset=\{p_1,\ldots
p_\nsamp\}\subset\rrr^\dhigh$, where $\dhigh$ is typically high. These are 
assumed to be sampled from a low-dimensional manifold $\M\subset \rrr^\dhigh$ and are mapped into vectors $\{f(p_1),\ldots f(p_\nsamp)\}\subset \rrr^\dembed$, with $\dembed\ll \dhigh$ and $\dintri \leq \dembed$. This terminology, as well as other differential geometry terms used in this section, will later be defined formally. 

\comment{In the case where the map $f$ is an embedding of $\M$ into $rrr^\dembed$ with $\dembed=\text{dim}(\M)$, $f$ can also} 
\comment{ for $\M$   vectors in $\rrr^\dembed$ are denoted by $\mbx$, and we will use $f(p)$ and
$x(p)$ interchangeably for the image of point $p\in \rrr^\dhigh$, calling
it the {\em coordinates} of $p$}

\comment{\mmp{take this out... some more on the same was taken out from here}
As the map $f$ will be used to define coordinate chart, we will
sometime denote $f$ with $\mbx$ in keeping standard differential
geometry notation and clarify in what context $f(p)$ is thought as the
{\em coordinates} of $p$.} 

\comment{
Sometimes the dimension of the output $x(p)$ is equal to $\dintri$, in
which case $x(p)$ represents a {\em coordinate chart}. This is
examplified in Figure \ref{fig:chart} (left). Other times, the
dimension $m$ of $x(p)$ is larger than $\dintri$, in which case we
call $x(p)$ and {\em embedding}. Such a situation, typical of closed
manifolds, is shown in Figure \ref{fig:chart} (right).
}

\subsection{Nearest Neighbors Graph} \label{ssec:nng}

Existing manifold learning algorithms pre-process the data by first
constructing a \textit{neighborhood graph} $\G = (V,E)$, where $V$ are 
the vertices and $E$ the edges of $\G$. While the vertices are generally taken
to be the observed points $V = \{p_1,\dots,p_\nsamp\}$, there are three
common approaches for constructing the edges. 

The first approach is to construct the edges by connecting the $k$ nearest neighbors for each vertex.
Specifically, $(p_i,p_j) \in E_k$ if $p_i$ is one of the k-nearest neighbors of $p_j$ or if $p_j$ is one of the k-nearest neighbors of $p_i$. $\G_k =(V,E_k)$ is then known as the $k-$nearest neighbors graph. While it may be relatively easy to choose the neighborhood parameter $k$ with this method, it is not very intuitive in a geometric sense. 

The second approach is to construct the edges by finding all the neighborhoods of radius $\bw$ so that
$E_{\epps} = \{\1[||p_i-p_j||^2 \leq \epps]: i,j = 1,\dots,\nsamp\}$. This is known as the $\epps$-neighborhood
graph $\G_{\epps}=(V,E_{\epps})$. The advantage of this method is that it is geometrically motivated; however, it can be difficult to choose $\bw$, the bandwidth parameter. Choosing a $\bw$ that is too small may lead to disconnected components, while choosing a $\bw$ that is too large fails to provide locality information - indeed, in the extreme limit, we obtain a complete graph. Unfortunately, this does not mean that the range of values between these two extremes necessarily constitutes an appropriate middle ground for any given learning task. 

The third approach is to construct a complete weighted graph where the weights represent the \textit{closeness}
or \textit{similarity} between points. A popular approach for constructing the weights, and the one we
will be using here, relies on kernels \cite{TingHJ10}. For example, weights defined by the heat kernel are given by  
\beq
w_{\epps}(i,j) = \exp{\left(\frac{-\left|\left|p_i-p_j\right|\right|^2}{\epps}\right)} \label{eq:heat_kernel_w}
\eeq
such that $E_{w_{\epps}} = \{w_{\epps}(i,j):i,j=1,\dots,\nsamp\}$. The weighted neighborhood graph $G_{w_{\epps}}$ has the 
same advantage as the $\epps$-neighborhood graph in that it is geometrically motivated; however, it can be difficult to work with given that any computations have to be performed on a complete graph. This computational complexity can partially be alleviated by truncating for very small values of $w$ (or, equivalently, for a large multiple of $\bw$), but not without reinstating the risk of generating disconnected components. However, using a truncated weighted neighborhood graph compares favorably with using an $\epps'$-neighborhood graph with large values of $\epps'$ since the truncated weighted neighborhood graph $G_{w_{\epps}}$ - with $\epps < \epps'$ - preserves locality information through the assigned weights.


In closing, we note that some authors distinguish between the step of creating the nearest neighbors graph using any one of the methods we discussed above, and the step of creating the similarity graph (\cite{BelNiy02}). In practical terms, this means that one can improve on the $k$ nearest neighbors graph by applying the heat kernel on the existing edges, generating a weighted $k$ nearest neighbors graph.


\subsection{Existing Algorithms}

Without attempting to give a thorough overview of the existing
manifold learning algorithms, we discuss two main categories.  One category uses only local
information, embodied in $\G$ to construct the embedding. Local Linear Embedding (LLE) (\cite{SauRow03}), 
Laplacian Eigenmaps (LE) (\cite{BelNiy02}), Diffusion Maps (DM) (\cite{CoiLaf06}), and Local Tangent Space Alignment (LTSA) (\cite{ZhangZ:04}) are in this category.

Another approach is to use global information to construct the embedding, and the foremost example in this category is Isomap (\cite{TenDeS00}). Isomap estimates the shortest
path in the neighborhood graph $\G$ between every pair of data points $p,p'$,
then uses the Euclidean Multidimensional Scaling (MDS) algorithm
(\cite{BorGro05}) to embed the points in $\dembed$ dimensions with minimum distance distortion all at once. 

We now provide a short overview of each of these algorithms.

\begin{itemize}

\item \textbf{\textit{LLE: Local Linear Embedding}} is one of the algorithms that constructs $\G$ by connecting the $k$ nearest neighbors of each point. In addition, it assumes that the data is linear in each neighborhood $\G$, which means that any point $p$ can be approximated by a weighted average of its neighbors. The algorithm finds weights that minimize the cost of representing the point by its neighbors under the $L_2$-norm. Then, the lower dimensional representation of the data is achieved by a map of a fixed dimension that minimizes the cost, again under the $L_2$-norm, of representing the mapped points by their neighbors using the weights found in the first step.

\item \textbf{\textit{LE: The Laplacian Eigenmap}} is based on the random walk graph Laplacian, henceforth referred to as graph Laplacian, defined formally in Section \ref{sec:Algo} below. The graph Laplacian is used because its eigendecomposition can be shown to preserve local distances while maximizing the smoothness of the embedding. Thus, the LE embedding is obtained simply by keeping the first $\dembed$ eigenvectors of the graph Laplacian in order of ascending eigenvalues. The first eigenvector is omitted, since it is necessarily constant and hence non-informative. 

\item \textbf{\textit{DM: The Diffusion Map}} is a variation of the LE that emphasizes the deep connection between the graph Laplacian and heat diffusion on manifolds. The central idea remains to embed the data using an eigendecomposition of the graph Laplacian. However, DM defines an entire family of graph Laplacians, all of which correspond to different diffusion processes on $\M$ in the continuous limit. Thus, the DM can be used to construct a graph Laplacian whose asymptotic limit is the Laplace-Beltrami operator, defined in (\ref{sec:g-math}), independently of the sampling distribution of the data. This is the most important aspect of DM for our purposes. 

\item \textbf{\textit{LTSA: The Linear Tangent Space Alignment}} algorithm, as its name implies, is based on estimating the tangent planes of the manifold $\M$ at each point in the data set using the $k$-nearest neighborhood graph $\G$ as a window to decide which points should be used in evaluating the tangent plane. This estimation is acheived by performing a singular value decomposition of the data matrix for the neighborhoods, which offers a low-dimensional parameterization of the tangent planes. The tangent planes are then pieced together so as to minimize the reconstruction error, and this defines a global low-dimensional parametrization of the manifold provided it can be embedded in $\rrr^\dintri$. One aspect of the LTSA is worth mentioning here even though we will not make use of it: by obtaining a parameterization of all the tangent planes, LTSA effectively obtains the Jacobian between the manifold and the embedding at each point. This provides a natural way to move between the embedding $f(\M)$ and $\M$. Unfortunately, this is not true for all embedding algorithms: more often then not, the inverse map for out-of-sample points is not easy to infer. 

\item \textbf{\textit{MVU: Maximum Variance Unfolding}} (also known as Semi-Definite Embedding) (\cite{weinberger06unsupervised}) represents the input and output data in terms of Gram matrices. The idea is to maximize the output variance, subject to exactly preserving the distances between neighbors. This objective can be expressed as a semi-definite program.
 
\item \textbf{\textit{ISOMAP}}: This is an example of a non-linear global algorithm. The idea is to embed $\M$ in $\rrr^\dembed$ using the minimizing geodesics between points. The algorithm first constructs $\G$ by connecting the $k$ nearest neighbors of each point and computes the distance between neighbors. Dijkstra's algorithm is then applied to the resulting local distance graph in order to approximate the minimizing geodesics between each pair of points. The final step consists in embedding the data using Multidimensional Scaling (MDS) on the computed geodesics between points. Thus, even though Isomap uses the linear MDS algorithm to embed the data, it is able to account for the non-linear nature of the manifold by applying MDS to the minimizing geodesics. 

\item \textbf{\textit{MDS}}: For the sake of completeness, and to aid in understanding the Isomap, we also provide a short description of MDS. MDS is a spectral method that finds an embedding into $\rrr^\dembed$ using dissimilarities (generally distances) between data points. Although there is more than one flavor of MDS, they all revolve around minimizing an objective function based on the difference between the dissimilarities and the distances computed from the resulting embedding.
\end{itemize}


\subsection{Manifolds, Coordinate Charts and Smooth Embeddings}
\label{sub:embed}

Now that we have explained the task of manifold learning in general terms and presented the most common embedding algorithms, we focus on formally defining manifolds, coordinate charts and smooth embeddings. These formal definitions set the foundation for the methods we will introduce in Sections \ref{sec:Metric} and \ref{sec:g-math}, as well as in later sections.

We first consider the geometric problem of manifold and metric representation, and define a smooth manifold in terms of coordinate charts. 

\begin{definition}[Smooth Manifold with Boundary \label{def:manifold_bnd}] A $\dintri$-dimensional
\textbf{manifold} $\M$ \textbf{with boundary} is a topological (Hausdorff) space such that every point
has a neighborhood homeomorphic to an open subset of $\mathbb{H}^d \equiv \{(x^{1},...,x^{\dintri})\in\rrr^{\dintri} |x^{1}\geq0\}$.  A
\textbf{chart} $(U,\mbx)$, or coordinate chart, of manifold $\M$ is an
open set $U\subset\M$ together with a homeomorphism
$\mbx:U\rightarrow V$ of $U$ onto an open subset
$V \subset \mathbb{H}^d$. 
A $C^{\infty}$-\textbf{Atlas} $\mathcal{A}$ is a collection of charts, \[
\mathcal{A}\equiv\cup_{\alpha\in I}\{(U_{\alpha},\mbx_{\alpha})\}\,,\]
where $I$ is an index set, such that $\M=\cup_{\alpha\in I}U_{\alpha}$
and for any $\alpha,\beta\in I$ the corresponding transition map,\begin{equation}
\mbx_{\beta}\circ\mbx_{\alpha}^{-1}:\mbx_{\alpha}(U_{\alpha}\cap U_{\beta})\rightarrow\rrr^{\dintri}\,,\label{eq:TransMap}\end{equation}
is continuously differentiable any number of times. Finally, a \textbf{smooth manifold} $\M$ \textbf{with boundary} is a manifold with boundary with a $C^{\infty}$-Atlas.\end{definition} 

Note that to define a manifold without boundary, it suffices to replace $\mathbb{H}^d$ with $\rrr^\dintri$ in Definition \ref{def:manifold_bnd} . 
For simplicity, we assume throughout that the manifold is smooth, but it is commonly sufficient to have a
$\mathcal{C}^2$ manifold, i.e. a manifold along with a $\mathcal{C}^2$ atlas. Following \cite{Lee03}, we will identify local coordinates of an open set $U\subset\M$ by the image coordinate
chart homeomorphism.  That is, we will identify $U$ by $\mbx(U)$
and the coordinates of point $p\in U$ by
$\mbx(p)=(x^{1}(p),...,x^{d}(p))$.

This definition allows us to reformulate the goal of manifold
learning: assuming that our (high-dimensional) data set
$\dataset=\{p_1,\ldots p_\nsamp\}\subset \rrr^\dhigh$ comes from a smooth
manifold with low $d$, the goal of manifold learning is to find a corresponding collection of $d$-dimensional coordinate charts for these data. 

The definition also hints at two other well-known facts.  First, the
coordinate chart(s) are not uniquely defined, and there are
infinitely many atlases for the same manifold $\M$
(\cite{Lee03}). Thus, it is not obvious from coordinates alone whether two atlases represent the same manifold or not. In particular, to compare the outputs of a manifold learning algorithm with the original data, or with the
result of another algorithm on the same data, one must resort to {\em
intrinsic}, coordinate-independent quantities. As we shall see later
in this chapter, the framework we propose takes this observation into account.

The second remark is that a manifold cannot be represented in general
by a {\em global} coordinate chart. For instance, the sphere is a
2-dimensional manifold that cannot be mapped homeomorphically to
$\rrr^2$; one needs at least two coordinate charts to cover the
2-sphere. It is also evident that the sphere is naturally embedded in
$\rrr^3$. 

One can generally circumvent the need for multiple charts by mapping
the data into $\dembed>\dintri$ dimensions as in this
example. Mathematically, the grounds for this important fact are
centered on the concept of {\em embedding}, which we introduce next.

Let $\M$ and $\mathcal{N}$ be two manifolds, and
$f:\M\rightarrow\mathcal{N}$ be a $C^{\infty}$ (i.e {\em smooth}) map
between them. Then, at each point $p\in \M$, the Jacobian $df_p$ of
$f$ at $p$ defines a linear mapping between the tangent plane to $\M$
at $p$, denoted $T_p(\M)$, and the tangent plane to $\mathcal{N}$ at
$f(p)$, denoted $T_{f(p)}(\mathcal{N})$.

\begin{definition}[Rank of a Smooth Map]
A smooth map $f:\M\rightarrow\mathcal{N}$ has rank $k$ if the Jacobian
$df_{p}:T_{p}\M\rightarrow T_{f(p)}\mathcal{N}$ of the map has rank $k$
for all points $p\in\M$. Then we write
$rank\left(f\right)=k$. \end{definition}

\begin{definition}[Embedding \label{def:emb}] Let $\M$ and $\mathcal{N}$
be smooth manifolds and let $f:\M\rightarrow\mathcal{N}$
be a smooth injective map with $rank(f)=dim(\M)$, then
$f$ is called an immersion. If $f$ is a homeomorphism onto its image, 
then $f$ is an embedding of $\M$ into $\mathcal{N}$. 
\end{definition}

The Strong Whitney Embedding Theorem (\cite{Lee03}) states that any
$\dintri$-dimensional smooth manifold can be embedded into $\rrr^{2\dintri}$. It follows from this fundamental result that if the {\em intrinsic
dimension} $\dintri$ of the data manifold is small compared to the
observed data dimension $\dhigh$, then very significant dimension reductions
can be achieved, namely from $\dhigh$ to $\dembed\leq 2\dintri$\footnote{In practice, it may be more practical to consider $\dembed \leq 2\dintri+1$, since any smooth map $f:\M \rightarrow \rrr^{2\dintri+1}$ can be perturbed to be an embedding. See Whitney Embedding Theorem in \cite{Lee03} for details.} with a single map
$f:\M\rightarrow \rrr^\dembed$.

\comment{Clearly, an embedding into $\mathcal{N}=\rrr^{n}$
will not constitue a $C^{\infty}$-Atlas when $n>d$. At the same time,
it would be overly restrictive to consider only Riemannian manifolds
of dimension $d$ that can be embedded into $\rrr^{d}$, and hence
define a one chart $C^{\infty}$-Atlas. Indeed, any closed
$d$-dimensional manifold, such as the $d$-sphere, requires at least
$n=d+1$ to be embedded into $\rrr^{n}$. More generally,}

Whitney's result is tight, in the sense that some manifolds, such as
real projective spaces, need all $2\dintri$ dimensions. However, the
$\dhigh=2\dintri$ upper bound is probably pessimistic for most
data sets. Even so, the important point is that the existence of an
embedding of $\M$ into $\rrr^{\dintri}$ cannot be relied upon; at the same
time, finding the optimal $\dembed$ for an unknown manifold might be more
trouble than it is worth if the dimensionality reduction from the
original data is already significant, i.e. $2\dintri\ll \dhigh$.

In light of these arguments, for the purposes of our work, we
set the objective of manifold learning to be the recovery of an
embedding of $\M$ into $\rrr^{\dembed}$ subject to $\dintri \leq \dembed \leq 2\dintri$ and
with the additional assumption that $\dembed$ is sufficiently large to
allow a smooth embedding. That being said, the choice of $\dembed$ will only be discussed tangentially in this article and even then, the constraint $\dembed \leq 2\dintri$ will not be enforced. 

\comment{ In special cases, when the manifold $\M$ allows it,
$\dembed$ will equal $\dintri$ the dimension of the manifold, and the
embedding $f$ will be a coordinate chart. In this case, we will 
sometimes denote $f$ with $\mbx$ in keeping standard differential
geometry notation.}

\comment{
\begin{figure}
\begin{tabular}{cc}
{\small ``Swiss roll with hole'' $\dintri=2,\dhigh=3$} &
{\small ``1D Dumbbell''   $\dintri=1,\dhigh=2$}\\

{\small Chart for ``Swiss roll with hole'' $\dembed=\dintri=2$} &
{\small Embedding for ``1D Dumbbell''   $\dembed=2>\dintri=1$}\\

\end{tabular}
\caption{\label{fig:chart}
Examples of charting and embedding a manifold. The results were
obtained by the LTSA algorithm for ``Swiss Roll with hole'' and by the
LE algorithm for ``1D Dumbbell''. Note that an arbitrary number of
ambient dimensions could be added to $\dhigh$ without essentially
changing the algorithms' results.
 }
\end{figure}
} 

\setlength{\picwi}{0.32\textwidth}
\begin{figure}
\begin{tabular}{ccc}
Original data & LE & LLE  \\
$\dhigh=3$ & $\dembed=2$ & $\dembed=3$ \\
 \includegraphics[width=\picwi]{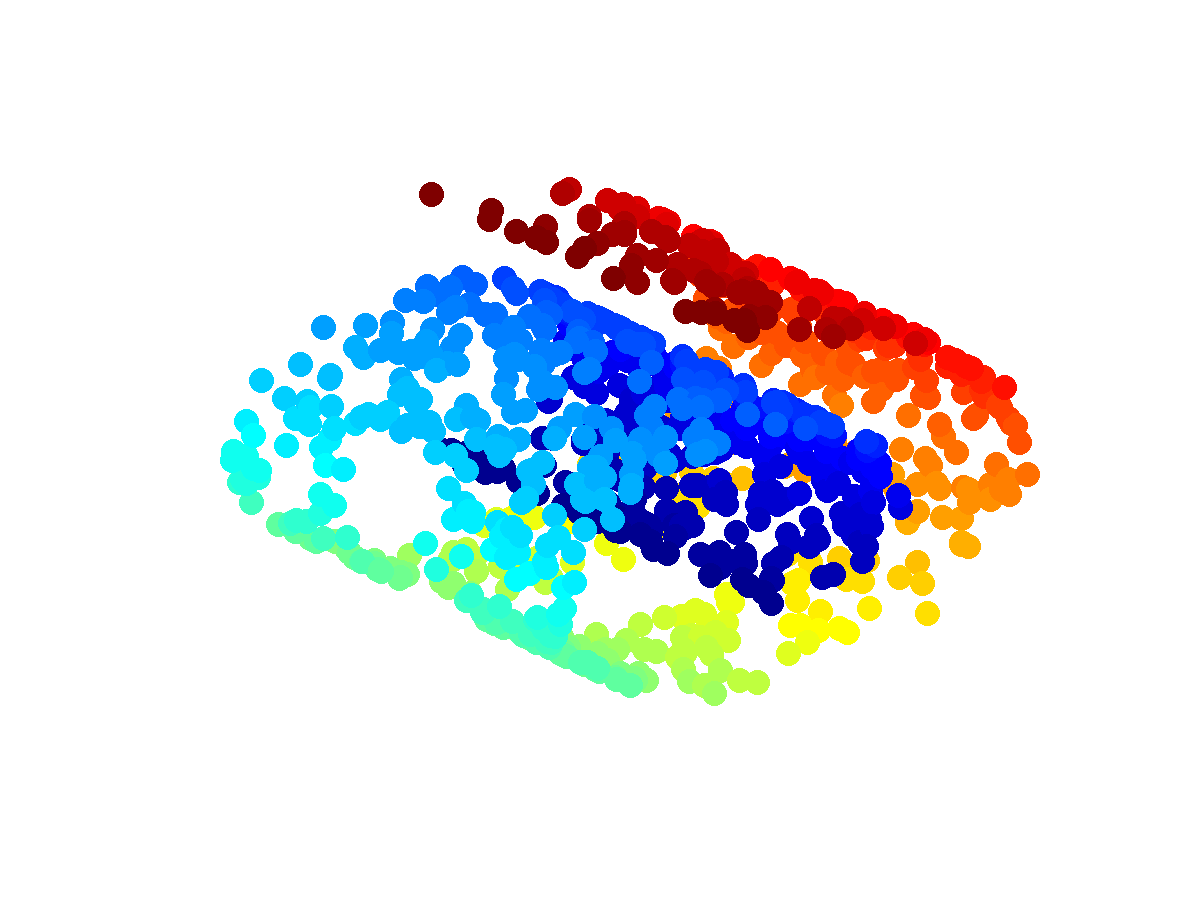}&
\includegraphics[width=\picwi]{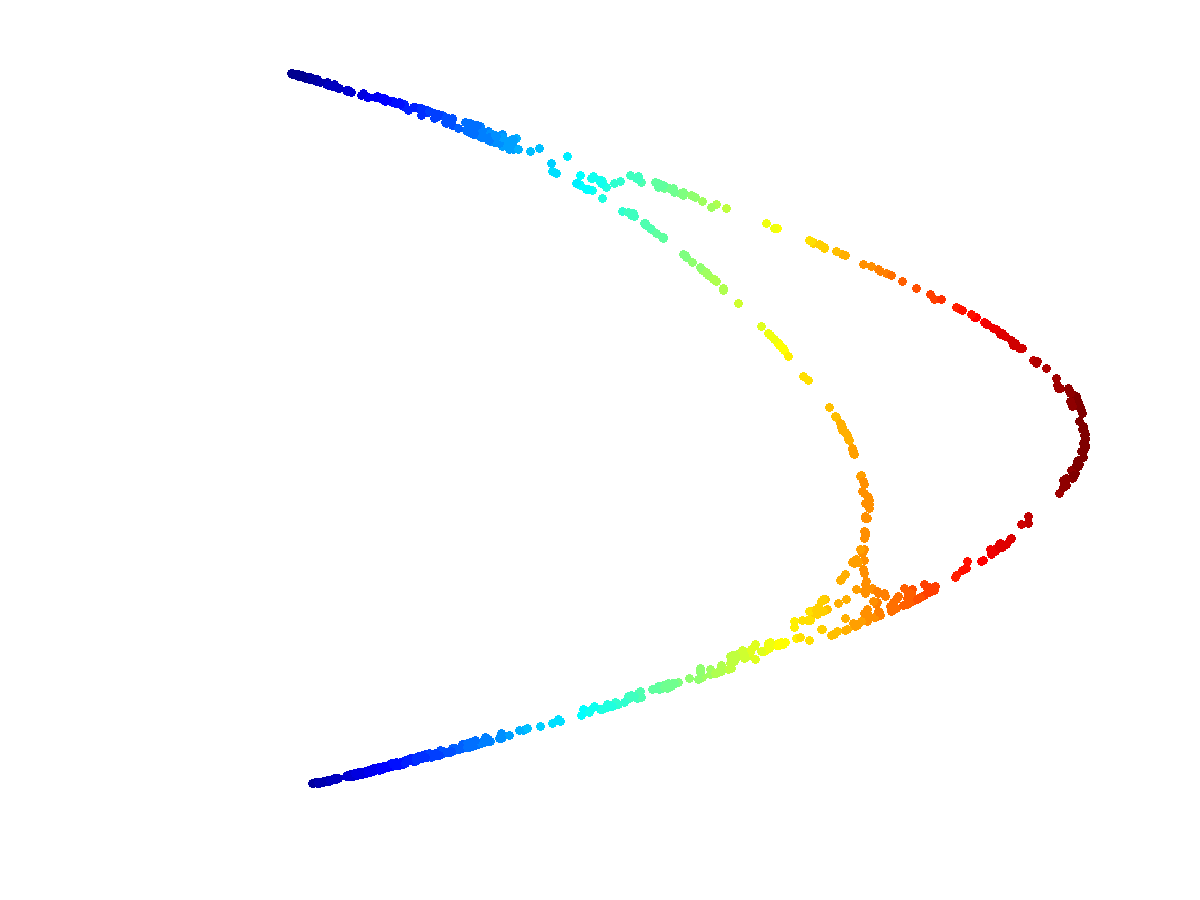}&
\includegraphics[width=\picwi]{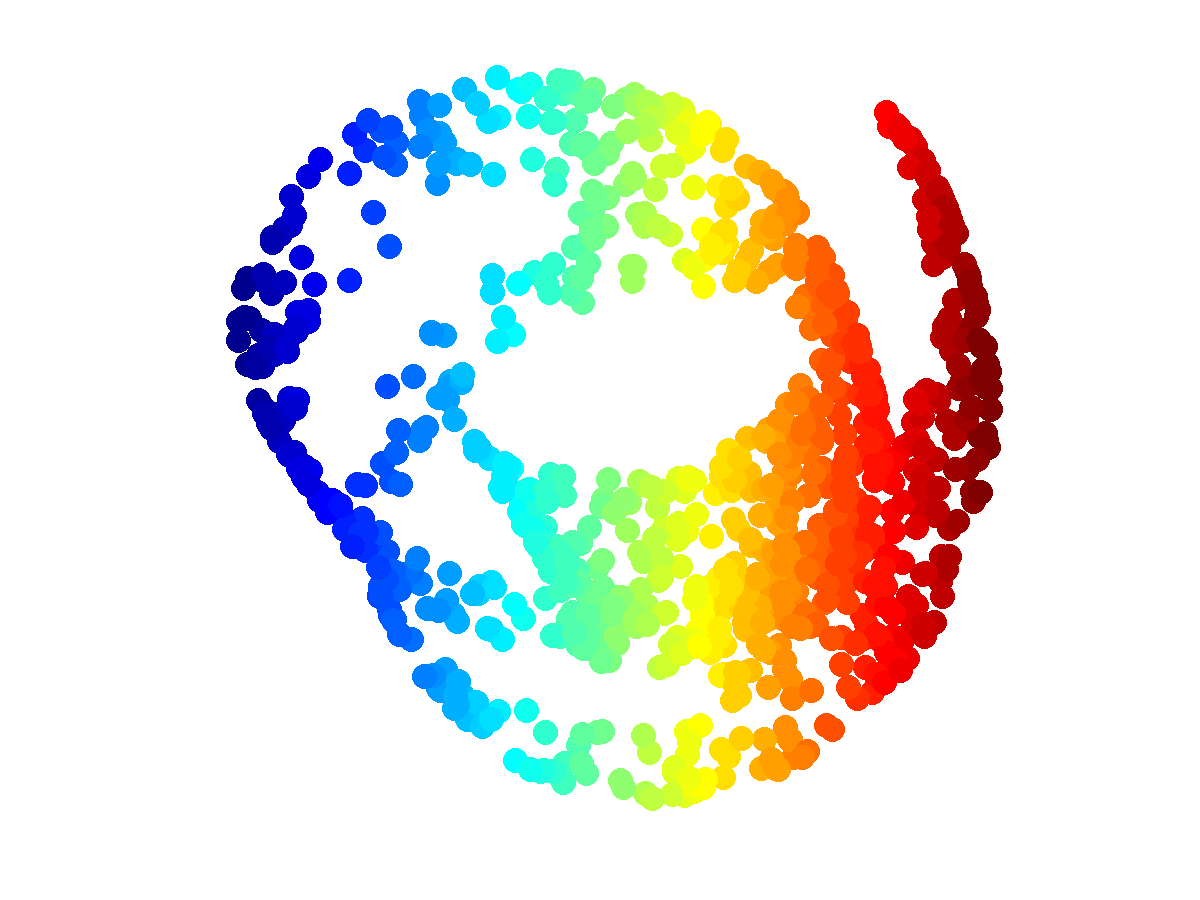}\\
\\
 LTSA & Isomap $k=12$ & Isomap $k=8$ \\
 $\dembed=2$ & $\dembed=2$ & $\dembed=2$ \\
\includegraphics[width=\picwi]{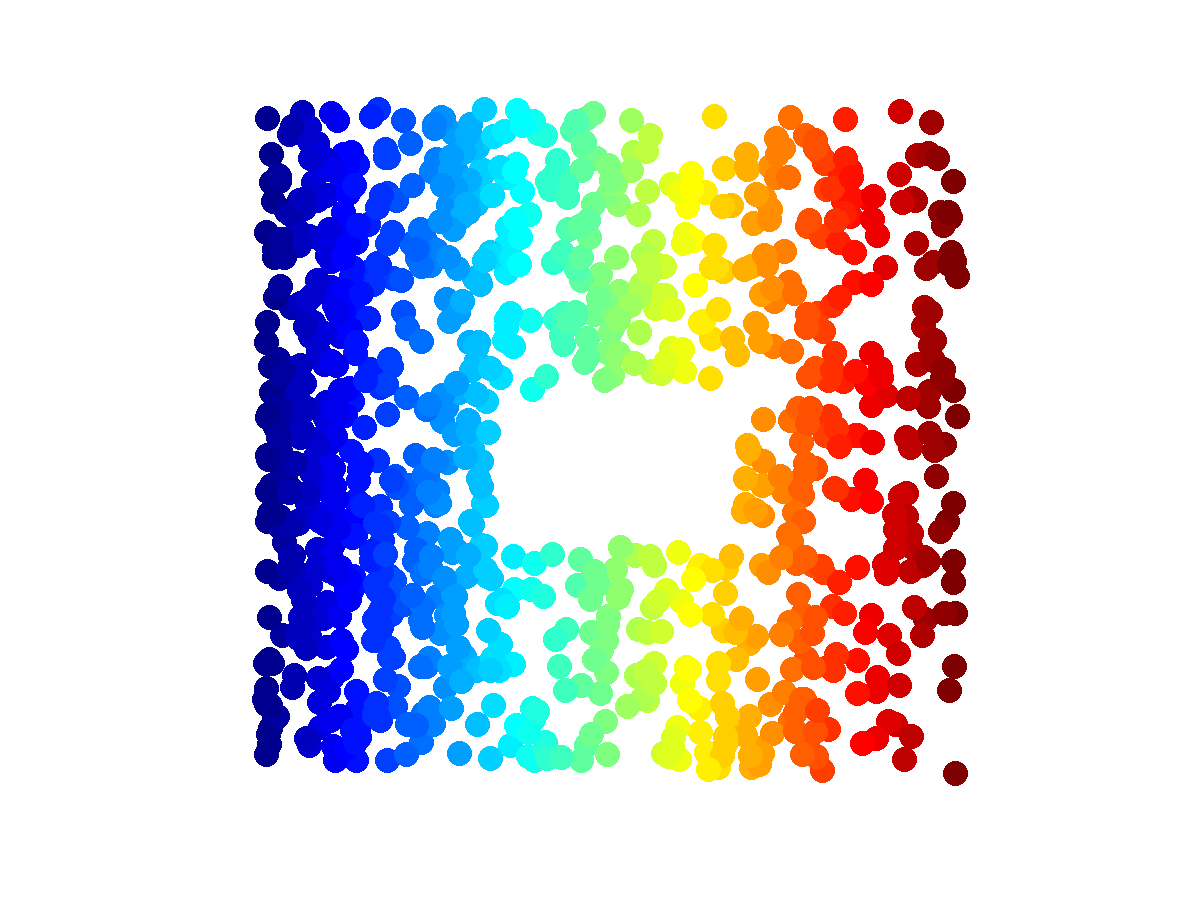}&
\includegraphics[width=\picwi]{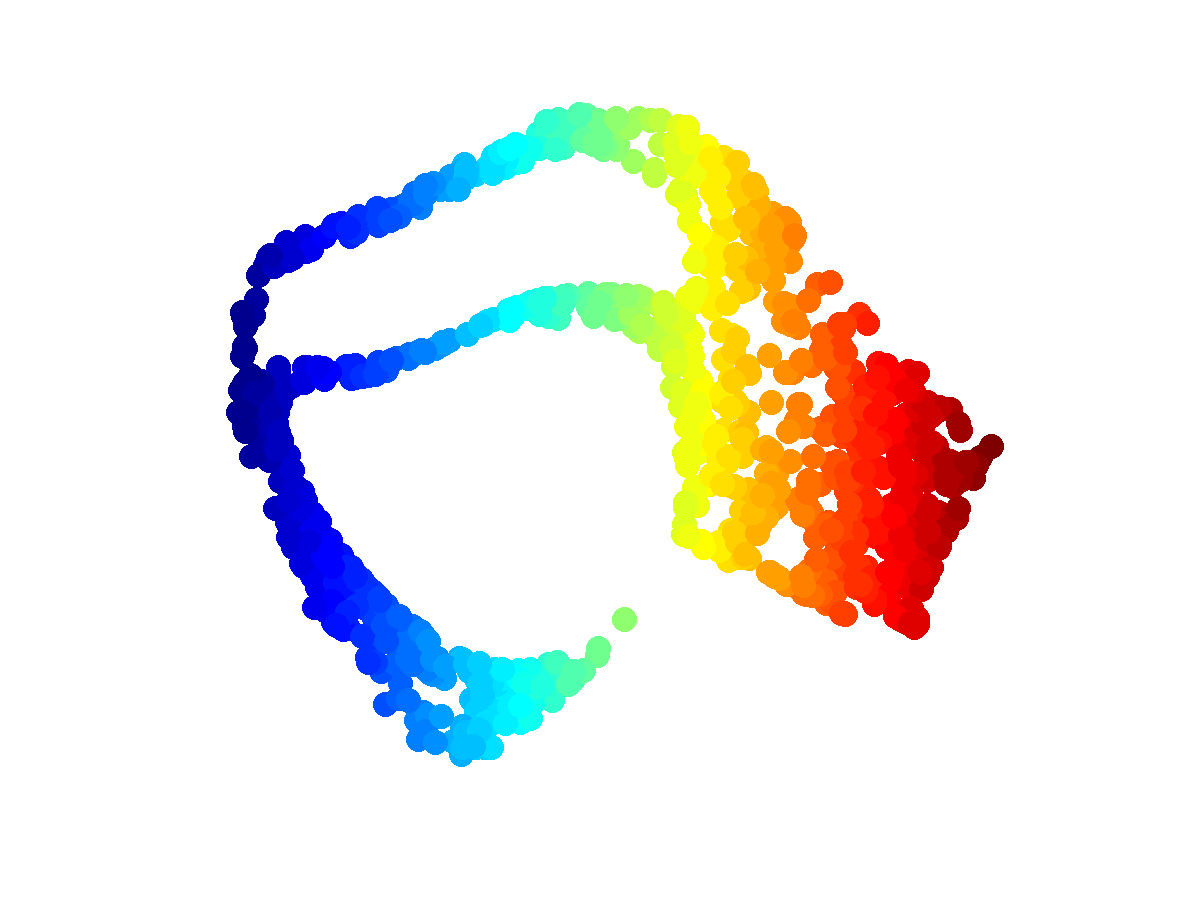}&
\includegraphics[width=\picwi]{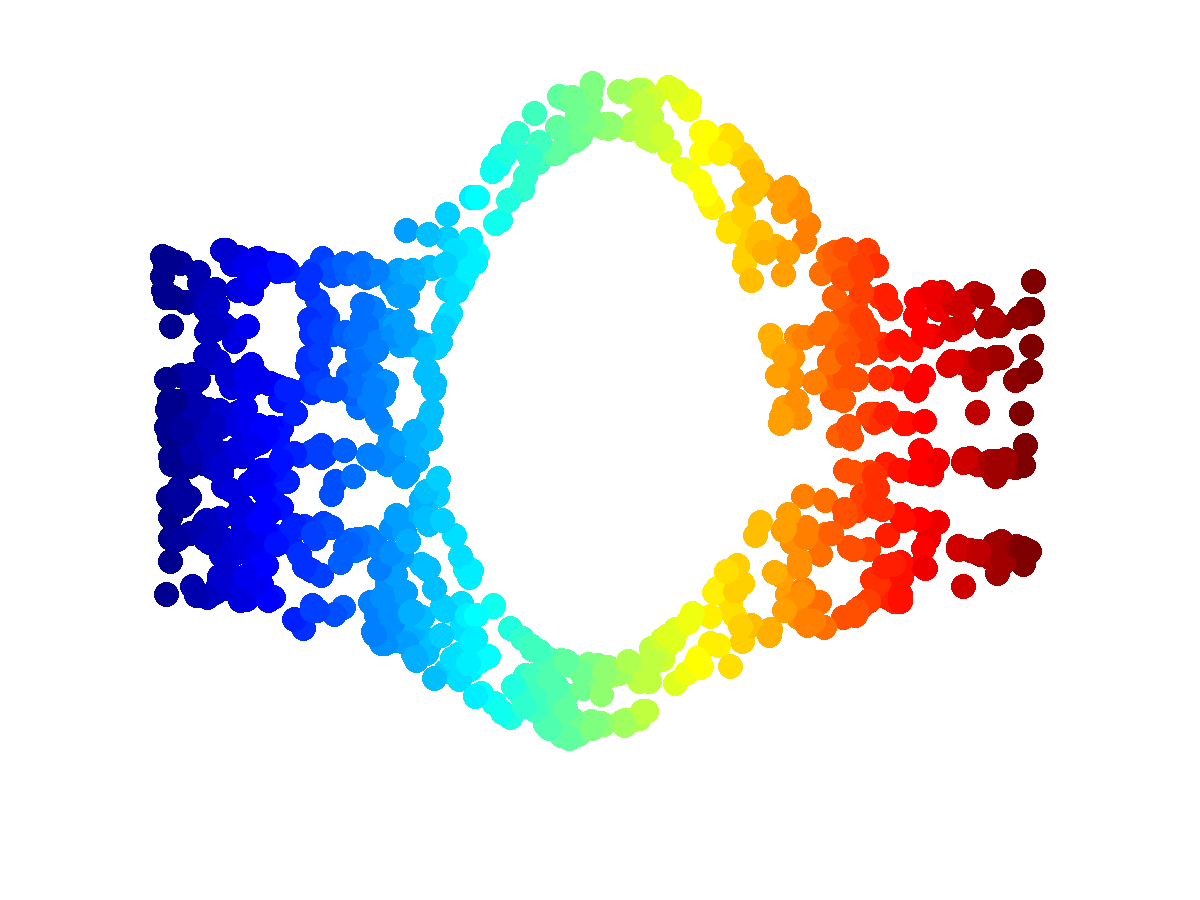}\\
\end{tabular}
\caption{\label{fig:swiss-fail}
Manifold learning algorithms distort the geometry of the data. The classical ``Swiss roll'' example is shown here embedded via a variety of manifold learning algorithms. For clarity, the original data is in $\dhigh=3$ dimensions; it is obvious that adding extra dimensions does not affect the resulting embeddings.
}
\end{figure}


\subsection{Consistency} \label{sub:quick-cons}

The previous section defined smoothness of the embedding in the ideal,
continuous case, when the ``input data'' covers the whole manifold $\M$
and the algorithm is represented by the map
$f:\M\rightarrow \mathbb{R}^\dembed$. This analysis is useful in order
to define what is mathematically possible in the limit. 
 
Naturally, we would hope that a real algorithm, on a
real finite data set $\dataset$, behaves in a way similar to its continuous
counterpart. In other words, as the sample size
$\nsamp=|\dataset|\rightarrow \infty$, we want the output of the
algorithm $f_{\nsamp}(\dataset)$  to converge to the output $f(\M)$ of the continous algorithm, irrespective of the particular sample, in a  probabilistic sense. This is what is generally understood as {\em consistency}
of the algorithm.

\comment{
Consistency is a very important property. It ensures that the
inferences we draw from a finite sample carry over to the generating
process that produced the sample. It also lets us study the idealized
continuous case in order to make predictions about an algorithm's
behavior for finite samples.
}

Proving consistency of various manifold-derived quantities has
received considerable attention in the literature (\citep{BernEtAL:00},
\citep{LuxBelBou08}). However, the meaning of consistency in the context of manifold learning remains unclear. For example, in the case of the Isomap algorithm, the convergence proof focuses on establishing that the graph that estimates the distance between two sampled points converges to the minimizing geodesic distance 
on the manifold $\M$ (\cite{BernEtAL:00}). Unfortunately, the proof does
not address the question of whether the empirical embedding $f_{\nsamp}$ is consistent for $f$ or whether $f$ defines a proper embedding. 

Similarly, proofs of consistency for other popular algorithms do not address 
these two important questions, but instead focus on showing that the linear
operators underpinning the algorithms converge to the appropriate differential operators (\cite{CoiLaf06,HeiAudLux07,GinKol06,TingHJ10}). Although this is an important problem in itself, it still falls short of establishing that $f_\nsamp \rightarrow f$. The exception to this are the results in \cite{LuxBelBou08,belniy07} that prove the convergence of the eigendecomposition of the graph Laplacian to that of the Laplace-Beltrami operator (defined in Section \ref{sec:g-math}) for a uniform sampling density on $\M$. These results also allow us to assume, by extension, the consistency of the class of algorithms that use the eigenvectors of the Laplace-Beltrami operator to construct embeddings - Laplacian Eigenmaps and Diffusion Maps.
Though incomplete in some respects, these results allow us to assume when necessary that an embedding algorithm is consistent and in the limit produces a smooth embedding. 
\comment{
Nonetheless, obtaining a
consistent estimator for the Riemannian metric will require us to
revisit the issue of consistency in Section
\ref{sec:Consistency}. 
}

We now turn to the next desirable property, one for which negative results abound.

\subsection{Manifold Geometry Preservation} \label{sub:failalgo}

Having a consistent smooth mapping from
$f:\M\;\rightarrow\;\rrr^\dembed$ guarantees that
neighborhoods in the high dimensional ambient space will be mapped into neighborhoods in the embedding space with some amount of ``stretching'', and vice versa. A reasonable question, therefore, is whether we
can reduce this amount of ``stretching'' to a minimum, even to zero.
In other words, can we preserve not only neighborhood relations, but
also distances within the manifold? Or, going one step further, could
we find a way to simultaneously preserve distances, areas, volumes,
angles, etc. - in a word, the {\em intrinsic geometry} - of the manifold?

Manifold learning algorithms generally fail at preserving the
geometry, even in simple cases. We illustrate this with the well-known
example of the ``Swiss-roll with a hole'' (Figure
\ref{fig:swiss-fail}), a two dimensional strip with a rectangular
hole, rolled up in three dimensions, sampled uniformly. Of course, no matter how the original sheet is rolled without
stretching, lengths of curves within the sheet will be
preserved. So will areas, angles between curves, and other geometric
quantities. However, when this data set is embedded using various algorithms, 
this does not occur. The LTSA algorithm recovers the original strip up to an
affine coordinate transformation (the strip is turned into a square);
for the other algorithms, the ``stretching'' of the original manifold
varies with the location on the manifold. As a consequence, distances,
areas, angles between curves - the intrinsic geometric
quantities - are not preserved between the original manifold and the
embeddings produced by these algorithms.

These shortcomings have been recognized and discussed in the
literature (\citep{GolZakKusRit08,ZhaZ:03}). More illustrative examples
can easily be generated with the software in \citep{Wit05}. 

The problem of geometric distortion is central to this article: the 
main contribution is to offer a constructive solution to it. The
definitions of the relevant concepts and the rigorous statement of the
problem we will be solving is found in the next section.

We conclude this section by stressing that the consistency of an
algorithm, while being a necessary property, does not help alleviate
the geometric distortion problem, because it merely guarantees that
the {\em mapping} from a set of points in high dimensional space to a
set of points in $\dembed$-space induced by a manifold learning
algorithm converges. It will not guarantee that the mapping recovers
the correct geometry of the manifold. In other words, even with
infinite data, the distortions observed in Figure \ref{fig:swiss-fail}
will persist.  
\section{Riemannian Geometry} \label{sec:Metric}

In this section, we will formalize what it means for an embedding
$f:\M\rightarrow\rrr^m$ to preserve the geometry of $\M$.  

\subsection{The Riemannian Metric \label{ssub:rmetric}}

The extension of Euclidean geometry
to a manifold $\M$ is defined mathematically via the Riemannian metric.
\begin{definition}[Riemannian Metric]
A Riemannian metric $g$ is a symmetric and positive definite tensor
field which defines an inner product $<,>_{g}$ on the tangent space
$T_{p}\M$ for every $p\in\M$. \end{definition}
\begin{definition}[Riemannian Manifold] A Riemannian manifold $(\M,g)$ is a smooth manifold $\M$ with a Riemannian metric $g$ defined at
every point $p\in\M$.
\end{definition}

The inner product $<u,v>_{g}=g_{ij}u^{i}v^{j}$ (with the Einstein
summation convention\footnote{This convention assumes implicit
summation over all indices appearing both as subscripts and
superscripts in an expression. E.g in $g_{ij}u^{i}v^{j}$ the symbol
$\sum_{i,j}$ is implicit.}) for $u,v\in T_{p}\M$ is used to define
usual geometric quantities such as the norm of a
vector$||u||=\sqrt{<u,v>_{g}}$ and the angle between two vectors
$\cos(\theta)=\frac{<u,v>_{g}}{||u||||v||}$. Thus, in any coordinate
representation of $\M$, $g$ at point $p$ is represented as a
$\dintri\times\dintri$ symmetric positive definite matrix.

The inner product $g$ also defines infinitesimal quantities such as
the line element $dl^{2}=g_{ij}dx^{i}dx^{j}$ and the volume element
$dV_g=\sqrt{\det(g)}dx^{1}\dots dx^{d}$, both expressed in local
coordinate charts. The length $l$ of a curve $c:[a,b]\rightarrow\M$
parametrized by $t$ then becomes
\beq \label{eq:l-curve}
 l(c) \; = \;
\int_{a}^{b}\sqrt{g_{ij}\frac{dx^{i}}{dt}\frac{dx^{j}}{dt}}dt,
\eeq
where $(x^{1},...,x^{d})$ are the coordinates of chart
$(U,\mathbf{x})$ with $c([a,b])\subset U$. Similarly, the volume of
$W\subset U$ is given by
\beq \label{eq:Vol}
 \text{Vol}(W)=\int_{W}\sqrt{\det(g)}dx^{1}\dots
dx^{d}\,.
\eeq 
Obviously, these definitions are trivially extended to
overlapping charts by means of the transition
map~\eqref{eq:TransMap}. For a comprehensive treatement of calculus on
manifolds, the reader is invited to consult \citep{Lee97}. 

\subsection{Isometry and the Pushforward Metric}

Having introduced the Riemannian metric, we can now formally discuss what 
it means for an embedding to preserve the geometry of 
$\M$. 

\begin{definition}[Isometry] Let $f:\M\rightarrow {\cal
N}$ be a diffeomorphism between two Riemannian manifolds $(\M,g),\,({\cal N},h)$ is called
an {\em isometry} iff for all $p\in \M$ and all $u,w\in T_p(\M)$
\[
<u,w>_{g(p)}\;=\;<df_pu, d_pfw>_{h(f(p))}
\]
\end{definition}
In the above, $df_p$ denotes the Jacobian of $f$ at $p$, i.e. the map
$df_p:T_p\M\rightarrow T_{f(p)}\mathcal{N}$.
An embedding will be isometric if $(f(\M),h|_{f(\M)})$ is 
isometric to $(\M,g)$, where $h|_{f(\M)}$ is the restriction of $h$, the metric of the embedding space $\mathcal{N}$, to the tangent space $T_{f(p)}f(\M)$. An isometric embedding obviously preserves path lengths, angles, areas and volumes. It is then natural to take isometry as the strictest notion of what it means for an algorithm to ``preserve geometry''. 

We also formalize what it means to carry the geometry over from a Riemannian manifold $(\M,g)$ via an embedding $f$.
\begin{definition}[Pushforward Metric] 
Let $f$ be an embedding from the Riemannian manifold $(\M,g)$ to another manifold $\N$. Then the pushforward $h=\varphi^{*}g$ of the metric $g$ along $\varphi \equiv f^{-1}$ is given by \begin{eqnarray*}
\left\langle u,v\right\rangle _{\varphi^{*}g_{p}} & = & \left\langle df_{p}^{-1}u,df_{p}^{-1}v\right\rangle _{g_{p}}\,,\end{eqnarray*}
for $u,v\in T_{f(p)}\N$ and where $df_{p}^{-1}$ denotes
the Jacobian of $f^{-1}$.
\end{definition}
This means that, by construction, $(\N,h)$ is isometric to $(\M,g)$. 

As the defintion implies, the superscript $-1$ also refers to the fact
that $df_{p}^{-1}$ is the matrix inverse of the jacobian $df_p$. This
inverse is well-defined since $f$ has full rank $d$. In the next
section, we will extend this definition by considering the case where
$f$ is no longer full-rank. 

%

\subsection{Isometric Embedding vs. Metric Learning}

Now consider a manifold embedding algorithm, like Isomap or Laplacian
Eigenmaps. These algorithms take points $p\in \rrr^\dhigh$ and map them through some function $f$ into $\rrr^\dembed$. The geometries in the two representations
are given by the induced Euclidean scalar products in $\rrr^\dhigh$
and $\rrr^\dembed$, respectively, which we will denote by $\ihigh,\,\iembed$. In matrix form, these are represented by unit matrices\footnote{The actual metrics for $\M$ and $f(\M)$ are  $\ihigh |_\M$ and $\iembed |_{f(\M)}$, the restrictions of $\ihigh$ and $\iembed$ to the tangent bundle $T\M$ and $Tf(\M)$.}. In view of the
previous definitions, the algorithm will preserve the geometry of the
data only if the new manifold $(f(\M), \iembed)$ is isometric to the
original data manifold $(\M,\ihigh)$. 

The existence of an isometric embedding of a manifold into $\rrr^\dembed$ for 
some $\dembed$ large enough is guaranteed by Nash's theorem (\citep{Nas56}), reproduced here for completeness.

\begin{theorem}
If $\M$ is a given $\dintri$-dimensional Riemannian manifold of class
$C^k,\, 3\leq k \leq\infty$ then there exists a number
$\dembed\leq\dintri(3\dintri+11)/2$ if $\M$ is compact, or
$\dembed\leq\dintri(\dintri+1)(3\dintri+11)/2$ if $\M$ is not compact,
and an injective map $f:\M\,\longrightarrow\,\rrr^\dembed$ of class
$C^k$, such that
\[ <u,v>\,=\,<df_p(v),\,df_p(v)>
\]
for all vectors $u,\, v$ in $T_p\M$.
\end{theorem}

The method developed by Nash to prove the existence of an isometric
embedding is not practical when it comes to finding an isometric
embedding for a data manifold. The problem is that the method involves
tightly wrapping the embedding around extra dimensions, which, as observed by \citep{DreKir07}, may
not be stable numerically\footnote{Recently, we became aware of a yet unpublished paper, which introduces
an algorithm for an isometric embedding derived from Nash's theorem. We
are enthusiastic about this achievement, but we note that achieving an isometric embedding via Nash does
not invalidate what we propose here, which is an alternative approach
in pursuit of the desirable goal of ``preserving geometry''.}.


Practically, however, as it was shown in Section \ref{sub:failalgo},
manifold learning algorithms do not generally define isometric
embeddings. The popular approach to resolving this problem is to try
to correct the the resulting embeddings as much as possible
(\citep{GolRit09,DreKir07,BehRai10,ZhaZ:03}). \comment{Although this approach is
interesting, in practice these corrections are usually approximate and
tied to the specific algorithm they aim to correct.}

We believe that there is a more elegant solution to this problem,
which is to carry the geometry over along with $f$ instead of trying
to correct $f$ itself. Thus, we will take the coordinates $f$ produced
by any reasonable embedding algorithm, and augment them with the
appropriate (pushforward) metric $h$ that makes $(f(\M),h)$ isometric
to the original manifold $(\M,g)$. We call this procedure {\em metric
learning}. 


\section{Recovering the Riemannian Metric: The Mathematics}
\label{sec:g-math}

We now establish the mathematical results that will allow us to
estimate the Riemannian metric $g$ from data. The key to obtaining $g$ for
any $C^{\infty}$-Atlas is the Laplace-Beltrami operator $\LB $
on $\M$, which we introduce below. Thereafter, we extend the solution to manifold embeddings, where the embedding dimension
$\dembed$ is, in general, greater than the dimension of $\M$, $\dintri$. 

\subsection{The Laplace-Beltrami Operator\label{sub:lp-op} and $g$}

\begin{definition}[Laplace-Beltrami Operator]
The Laplace-Beltrami operator $\LB $ acting on a
twice differentiable function $f:\M \rightarrow \rrr$
is defined as $\LB f\equiv \text{div} \cdot \nabla f$.\end{definition}

In local coordinates, for chart $(U,\mbx)$, the Laplace-Beltrami
operator $\LB $ is expressed by means of $g$ as per \cite{Ros97}
\begin{equation}
\LB f=\frac{1}{\sqrt{\det(g)}}\frac{\partial}{\partial x^{l}}\left(\sqrt{\det(g)}g^{lk}\frac{\partial}{\partial x^{k}}f\right)\,.\label{eq:lb-op}\end{equation}
In~\eqref{eq:lb-op}, $g^{lk}$ denotes the $l,k$ component of the
inverse of $g$ and Einstein summation is assumed.

The Laplace-Beltrami operator has been widely used in the context of
manifold learning, and we will exploit various existing results about
its properties. We will present those results when they
become necessary. For more background, the reader is invited to
consult \cite{Ros97}. In particular, methods for
estimating $\LB$ from data exist and are well
studied (\cite{CoiLaf06,HeiAudLux07,BelSunWan08}). This makes using \eqref{eq:lb-op} ideally suited to recover $g$. The simple but powerful proposition below is the key to achieving this.

\begin{proposition}\label{prop:L-g}
Given a coordinate chart $(U,\mbx)$ of a smooth Riemannian manifold $\M$ and $\Delta_\M$ defined on $\M$, then the $g(p)^{-1}$, the inverse of the Riemannian metric, or dual metric, at point $p\in U$ as expressed in local coordinates $\mbx$, can be derived from
\beq \label{eq:g_inv}
g^{ij}\;=\;\frac{1}{2}\LB \left(x^{i}-x^{i}(p)\right)\left(x^{j}-x^{j}(p)\right)|_{x^{i}=x^{i}(p),x^{j}=x^{j}(p)}
\eeq
with $i,j=1,\dots,\dintri$. 
\end{proposition}

\begin{proof}
This follows directly from~\eqref{eq:lb-op}. Applying $\Delta_\M$ to the coordinate products of $x^i$ and $x^j$ centered at $\mbx(p)$, i.e.
$\frac{1}{2}\left(x^{i}-x^{i}(p)\right)\left(x^{j}-x^{j}(p)\right)$, and evaluating this expression at $\mbx=\mbx(p)$ using~\eqref{eq:lb-op} gives 
\[
g^{lk}\fracpartial{}{x^{l}}\left(x^{i}-x^{i}(p)\right)
 \times\fracpartial{}{x^{k}}\left(x^{j}-x^{j}(p)\right)|_{x^{i}=x^{i}(p),x^{j}=x^{j}(p)}
\;=\;g^{ij}\, , 
\]
since all the first order derivative terms vanish. The superscripts $i,j$ in the equation above and in~\eqref{eq:g_inv} refer to the fact that $g^{ij}$ is the inverse, i.e. dual metric, of $g$ for coordinates $x^i$ and $x^j$.
\end{proof}
With all the components of $g^{-1}$
 known, it is straightforward to compute its inverse and obtain $g(p)$. The power of Proposition~\ref{prop:L-g} resides in the fact that the coordinate chart is arbitrary. Given a coordinate chart (or embeddding, as will be shown below), one can apply the {\em coordinate-free} Laplace-Beltrami operator as in~\eqref{eq:g_inv} to recover $g$ for that coordinate chart. 

\subsection{Recovering a Rank-Deficient Embedding Metric} \label{sub:H}
 
In the previous section, we have assumed that we are given a
coordinate chart $(U,\mbx)$ for a subset of $\M$, and have shown how to obtain the Riemannian metric of $\M$ in that coordinate chart via the Laplace-Beltrami operator.

Here, we will extend the method to work with any embedding of $\M$. The
main change will be that the embedding dimension $\dembed$ may be larger
than the manifold dimension $\dintri$. In other words, there will be $\dembed \geq \dintri$ embedding coordinates for each point $p$, while $g$ is only 
defined for a vector space of dimension $d$. An obvious solution to this 
is to construct a coordinate chart around $p$ from the embedding $f$. 
This is often unnecessary, and in practice it is simpler to work directly from $f$ until the coordinate chart representation is actually required. In fact, once we have the correct metric for $f(\M)$, it becomes relatively easy to construct coordinate charts for $\M$. 

Working directly with the embedding $f$ means that at each embedded
point $f_p$, there will be a corresponding $\dembed\times \dembed$
matrix $h_p$ defining a scalar product. The matrix $h_p$ will have
rank $\dintri$, and its null space will be orthogonal to the tangent
space $T_{f(p)}f(\M)$. We define $h$ so that $(f(\M), h)$ is isometric
with $(\M, g)$. Obviously, the tensor $h$ over $ T_{f(p)}f(\M)
\bigoplus T_{f(p)}f(\M)^\perp \cong \rrr^\dembed$ that achieves this
is an extension of the {\em pushforward} of the metric $g$ of $\M$.

\begin{definition}[Embedding (Pushforward) Metric] \label{def:h} 
For all \begin{equation}
u,v\in T_{f(p)}f(\M) \bigoplus T_{f(p)}f(\M)^\perp, \nonumber
\end{equation} the \emph{embedding} pushforward metric $h$, or shortly the \emph{embedding metric}, of an embedding
$f$ at point $p\in\M$ is defined by the inner product\begin{eqnarray}
\label{eq:push-f}
<u,v>_{h(f(p))} & \equiv & <df_{p}^{\dagger}\left(u\right),df_{p}^{\dagger}\left(v\right)>_{g(p)}\,,\end{eqnarray}
where \begin{equation}
df_{p}^{\dagger}:  T_{f(p)}f(\M) \bigoplus T_{f(p)}f(\M)^\perp \rightarrow T_{p}\M \nonumber
\end{equation}
is the pseudoinverse of the Jacobian $df_{p}$ of $f:\M\rightarrow \rrr^\dembed$
\end{definition} 

In matrix notation, with $ df_{p}\equiv J$, $g\equiv G$ and $h\equiv
H$, \eqref{eq:push-f} becomes
\beq
u^{t}J^{t}HJv\; = \; u^{t}Gv
\eeq
 Hence,
\beq
H \;\equiv\; \left(J^{t}\right)^{\dagger}GJ^{\dagger}
\eeq

In particular, when $\M\subset\rrr^{\dhigh}$, with the metric inherited from the ambient Euclidean space, as is often the case for manifold learning, we have that $G=\Pi^{t}I_\dhigh\Pi$, where $I_\dhigh$ is the Euclidean metric in $\rrr^{\dhigh}$ and $\Pi$ is the orthogonal projection of $v\in\rrr^{\dhigh}$ onto $T_{p}\M$. Hence, the embedding metric $h$ can then be expressed as 
\begin{equation} H(p)=\left(J^{t}\right)^{\dagger}\Pi(p)^{t}I_\dhigh\Pi(p)J^{\dagger}\, .\label{eq:pushforward}
\end{equation}

The constraints on $h$ mean that $h$ is symmetric semi-positive definite
(positive definite on $T_{p}f(\M)$ and null on
$T_{p}f(\M)^{\perp}$, as one would hope), rather than symmetric positive definite like $g$.

One can easily verify that $h$ satisfies the following proposition: 
\begin{proposition} \label{prop:h}
Let $f$ be an embedding of $\M$ into $\rrr^\dembed$; then $(\M,g)$ and $(f(\M),h)$ are isometric, where $h$ is the embedding  metric $h$ defined in Definition \ref{def:h}. Furthermore, $h$ is null over $T_{f(p)}f(\M)^\perp$.
\end{proposition}

\begin{proof}
Let $u\in T_{p}\M$, then the map $df_{p}^{\dagger}\circ df_{p}:T_{p}\M\rightarrow T_{p}\M$ satisfies $df_{p}^{\dagger}\circ df_{p}(u)=u$, since $f$ has rank
$d=dim(T_{p}\M)$. So $\forall u,v\in T_{p}\M$ we have
\beq
<df_{p}(u),df_{p}(v)>_{h(f(p))}\; = \;<df_{p}^{\dagger}\circ df_{p}(u),df_{p}^{\dagger}\circ df_{p}(v)>_{g(p)}
\;=\;<u,v>_{g(p)}
\eeq
Therefore, $h$ ensures that the embedding is isometric. Moreover,
the null space of the pseudoinverse is ${\rm Null}(df_{p}^{\dagger})
= {\rm Im}(df_{p})^{\perp}=T_{p}f\left(\M\right)^{\perp}$, hence
$\forall u\in T_{p}f\left(\M\right)^{\perp}$ and $v$ arbitrary, the inner product defined by $h$ satisfies
\beq
<u,v>_{h(f(p))}\;=\;<df_{p}^{\dagger}\left(u\right),df_{p}^{\dagger}\left(v\right)>_{g(p)}\; =\;<0,df_{p}^{\dagger}\left(v\right)>_{g(p)}\; =\; 0 \, .
\eeq
By symmetry of $h$, the same holds true if $u$ and $v$ are interchanged. 
\end{proof}

Having shown that $h$, as defined, satisfies the desired properties,
the next step is to show that it can be recovered using $\LB $,
just as $g$ was in Section~\ref{sub:lp-op}.

\begin{proposition}
Let $f$ be an embedding of $\M$ into $\rrr^{\dembed}$, and
$df$ its Jacobian. 
Then, the embedding metric $h(p)$ is given by the pseudoinverse of $\tilde{h}$, where
\beq\label{eq:H}
\tilde{h}^{ij}\;=\;
\LB \frac{1}{2}\left(f^{i}-f^{i}(p)\right)\left(f^{j}-f^{j}(p)\right)|_{f^{i}=f^{i}(p),f^{j}=f^{j}(p)}
\eeq
\end{proposition}
\begin{proof} 
We express $\LB $ in a coordinate chart $(U,\mbx)$. $\M$ 
being a smooth manifold, such a coordinate chart always 
exists. Applying $\Delta_\M$ to the centered product of coordinates of the embedding, i.e.
$\frac{1}{2}\left(f^{i}-f^{i}(p)\right)\left(f^{j}-f^{j}(p)\right)$,
then \eqref{eq:lb-op} means that
\begin{eqnarray*}
\LB \frac{1}{2}\left(f^{i}-f^{i}(p)\right)\left(f^{j}-f^{j}(p)\right)|_{f^{i}=f^{i}(p),f^{j}=f^{j}(p)}
 & = & g^{lk}\frac{\partial}{\partial
 x^{l}}\left(f^{i}-f^{i}(p)\right) \\
 &   &\times\frac{\partial}{\partial x^{k}}\left(f^{j}-f^{j}(p)\right)|_{f^{i}=f^{i}(p),f^{j}=f^{j}(p)}\\
 & = & g^{kl}\frac{\partial f^{i}}{\partial x^{l}}\frac{\partial
 f^{j}}{\partial x^{k}}
\end{eqnarray*}
Using matrix notation as before, with $J\equiv df_p,\,G\equiv g(p),\, H\equiv h$, $\tilde{H}\equiv \tilde{h}$, the above results take the form
\beq
g^{kl}\frac{\partial f^{i}}{\partial x^{l}}\frac{\partial
 f^{j}}{\partial x^{k}} = (JG^{-1}J^t)_{ij} = \tilde{H}_{ij} \, . \label{eq:H_tilde}
\eeq
Hence, $\tilde{H} = JG^{-1}J^t$ and 
it remains to show that $H\;=\;\tilde{H}^\dagger$, i.e. that
\beq
\left(J^{t}\right)^{\dagger}GJ^{\dagger} = \left (JG^{-1}J^t \right )^\dagger.
\eeq
This is obviously straightforward for square invertible matrices, but if $\dintri<\dembed$, this might not be the case. Hence, we need an additional technical fact:
guaranteeing that
\beq \label{eq:dagger_prop}
\left(AB\right)^{\dagger}=B^{\dagger}A^{\dagger}
\eeq
requires $C=AB$ to constitute a full-rank decomposition of $C$,
i.e. for $A$ to have full column rank and $B$ to have full row
rank (\cite{BenGre03}).  In the present case, $G^{-1}$ has full rank, $J$ has
full column rank, and $J^{t}$ has full row rank. All these ranks are
equal to $d$ by virtue of the fact that $dim(\M)=d$ and $f$ is an
embedding of $\M$. Therefore, applying~\eqref{eq:dagger_prop}
repeatedly to $JG^{-1}J^t$, implicitly using the fact that
$\left(G^{-1}J^{t}\right)$ has full row rank since $G^{-1}$ has full
rank and $J$ has full row rank, proves that $h$ is the pseudoinverse of $\tilde{h}$. 
\end{proof}

\subsection*{Discussion}

Computing the pseudoinverse of $\tilde{h}$ generally means performing a Singular Value Decomposition ({SVD}). It is interesting to note that this decomposition offers very useful insight into the embedding. Indeed, we know from Proposition \ref{prop:h} that $h$ is positive definite over $T_{f(p)}f(\M)$ and null over $T_{f(p)}f(\M)^\perp$. This means that the singular vector(s) with non-zero singular value(s) of $h$ at $f(p)$ define an orthogonal basis for $T_{f(p)}f(\M)$, while the singular vector(s) with zero singular value(s) define an orthogonal basis for  $T_{f(p)}f(\M)^\perp$ (not that the latter is of particular interest). Having an orthogonal basis for $T_{f(p)}f(\M)$ provides a natural framework for constructing a coordinate chart around $p$. The simplest option is to project a small neighborhood $f(U)$ of $f(p)$ onto $T_{f(p)}f(\M)$, a technique we will use in Section \ref{sec:App} to compute areas or volumes. An interesting extension of our approach would be to derive the exponential map for $f(U)$. However, computing all the geodesics of $f(U)$ is not practical unless the geodesics themselves are of interest for the application. In either case, computing $h$ allows us to achieve our set goal for manifold learning, i.e. construct a collection of coordinate charts for $\dataset$. We note that it is not always necessary, or even wise, to construct an Atlas of coordinate charts explicitly; it is really a matter of whether charts are required to perform the desired computations.  

Another fortuitous consequence of computing the pseudoinverse is that the non-zero singular values yield a measure of the distortion induced by the embedding. Indeed, if the embedding were isometric to $\M$ with the metric inherited from $\rrr^\dembed$, then the embedding metric $h$ would have non-zero singular values equal to 1. This can be used in many ways, such as getting a global distortion for the embedding, and hence as a tool to compare various embeddings. It can also be used to define an objective function to minimize in order to get an isometric embedding, should such an embedding be of interest. From a local perspective, it gives insight into what the embedding is doing to specific regions of the manifold and it also prescribes a simple linear transformation of the embedding $f$ that makes it locally isometric to $\M$ with respect to the inherited metric $\iembed$. This latter attribute will be explored in more detail in Section \ref{sec:App}. 


\comment{
Note that, having proved \eqref{eq:H_tilde}, we can now take it as the
definition of the pseudoinverse for the square, symmetric, rank
defective matrix $\tilde{H}^\dagger$. Equation \eqref{eq:H_tilde} also
indicates that $H$ can be computed by a Singular Value Decomposition
(SVD) of $\tilde{H}$.
} 

\section{Recovering the Riemannian Metric: The Algorithm} \label{sec:Algo}

The results in the previous section apply to any embedding of $\M$ and
can therefore be applied to the output of any embedding algorithm,
leading to the estimation of the corresponding $g$ if $\dintri
=\dembed$ or $h$ if $ \dintri <\dembed$. In this section, we present
our algorithm for the estimation procedure, called \riememb. Throughout, we assume that
an appropriate embedding dimension $\dembed \geq \dintri$ is already
selected and $\dintri$ is known.

\subsection{Discretized Problem} \label{ssec:discret_problem}

Prior to explaining our method for estimating $h$ for an embedding algorithm, it is important to discuss the discrete version of the problem.

As briefly explained in Section \ref{sec:Prob}, the input data for a manifold learning algorithm is a set of points $\dataset = \{p_1,\dots,p_{\nsamp}\}\subset \M$ where $\M$ is a compact Riemannian manifold. These points are assumed to be an i.i.d. sample with distribution  $\density $ on $\M$, which is absolutely continuous with respect to the Lebesgue measure on $\M$. From this sample, manifold learning algorithms construct a map $f_n:\dataset\rightarrow\rrr^\dembed$, which, if the algorithm is consistent, will converge to an embedding $f:\M\rightarrow\rrr^\dembed$. 

Once the map is obtained, we go on to define the embedding metric $h_n$. Naturally, it is relevant to ask what it means to define the embedding metric $h_n$ and how one goes about constructing it. Since $f_n$ is defined on the set of points $\dataset$, $h_n$ will be defined as a positive semidefinite matrix over $\dataset$. With that in mind, we can hope to construct $h_n$ by discretizing equation~\eqref{eq:H}. In practice, this is acheived by replacing $f$ with $f_n$ and $\LB$ with some discrete estimator $\Lbw$ that is consistent for $\LB$.  

\comment{
In what sense the estimators $f_n$, $h_n$, and $\Lbw$ can be said to be consistent will be discussed in Section \ref{sec:Consistency} below. In the meantime,
} 

We still need to clarify how to obtain $\Lbw$. The most common approach, and the one we favor here, is to start by considering the ``diffusion-like'' operator $\tilde{\mathcal{D}}_{\epps,\lam}$ defined via the heat kernel $w_{\epps}$ (see \eqref{eq:heat_kernel_w}):
\beqa      
\tilde{\mathcal{D}}_{\epps,\lam}(f)(x) &=& \int_\M \frac{\tilde{w}_{\epps,\lam}(x,y)}{
    \tilde{t}_{\epps,\lam}}) f(y)\density(y)dV_{g}(y) \, , \,\, \text{with } \mbx\in\M \text{ and where} \label{eq:cont_D_1}  \\
     \tilde{t}_{\epps,\lam}(x) & = & \int_\M \tilde{w}_{\epps,\lam}(x,y)) \density(y)dV_{g}(y)
     \, , \,\, \text{and} \, \, w_{\epps,\lam} = \frac{w_{\epps}(x,y)}{t_\epps^\lam(x) t_\epps^\lam(y)} 
     \, , \,\, \text{while} \nonumber \\
     \,\, t_{\epps}(x) & =& \int_\M w_\epps(x,y)\density(y)dV_{g}(y) \, , \nonumber  
\eeqa 
\cite{CoiLaf06} showed that $\tilde{\mathcal{D}}_{\epps,\lam}(f) = f+\epps c\LB f +\bigOO(\epps^2)$ provided
$\lam = 1$, $f\in \mathcal{C}^3(\M)$, and where $c$ is a constant that depends on the choice of kernel $w_{\epps}$\footnote{In the case of heat kernel \eqref{eq:heat_kernel_w}, $c=1/4$, which - crucially - is independent of the dimension of $\M$.}. Here, $\lambda$ is introduced to guarantee the appropriate limit in cases where the sampling density $\density$ is not uniform on $\M$.

Now that we have obtained an operator that we know will converge to $\LB$, i.e.
\beq
\tilde{\mathcal{L}}_{\epps,1}(f) \equiv \frac{\tilde{\mathcal{D}}_{\epps,1}(f) - f}{c\epps} = \LB f + \bigOO(\epps) \, , \label{eq:cont_L_1}
\eeq
we turn to the discretized problem, since we are dealing with a finite sample of points from $\M$. 

Discretizing \eqref{eq:cont_D_1} entails using the finite sample approximation: 
\beqa      
\tilde{\mathcal{D}}_{\epps,\nsamp}(f)(x) &=& \sum_{p_i\in\dataset} \frac{w_{\epps,\lam}(\mbx,p_i)}{
    \tilde{t}_{\epps,\lam,n}(\mbx)}f(p_i)  \, , \,\,  \text{with } \mbx\in\M \text{ and where} \label{eq:discret_D_1}  \\
     \tilde{t}_{\epps,\lam,n}(x) & = & \sum_{p_i\in\dataset} \tilde{w}_{\epps,\lam}(\mbx,p_i)
     \, , \,\, \text{and} \, \, \tilde{w}_{\epps,\lam} = \frac{w_{\epps}(\mbx,\mby)}{t_\epps^\lam(\mbx) d_\epps^\lam(\mby)} 
     \, , \,\, \text{while} \nonumber \\
     \,\, t_{\epps}(x) & =&  \sum_{p_i\in\dataset} w_\epps(\mbx,p_i) \, , \nonumber  
\eeqa 
and \eqref{eq:cont_L_1} now takes the form
\beq
\Lbw(f) \equiv \frac{\tilde{\mathcal{D}}_{\epps,1}(f) - f}{c\epps} = \LB f + \bigOO(\epps) \, . \label{eq:discret_L_1}
\eeq
Operator $\Lbw$ is known as the \textbf{\textit{geometric graph Laplacian}} (\cite{ZhouBelkin11}). We will refer to it simply as graph Laplacian in our discussion, since it is the only type of graph Laplacian we will need.

Note that since is $\M$ unknown, it is not clear when $\mbx \in \M$ and when $\mbx \in \rrr^\dhigh \setminus \M$. The need to define $\Lbw(f)$ for all $\mbx$ in $\M$ is mainly to study its asymptotic properties. In practice however, we are interested in the case of $\mbx \in \dataset$. The kernel $w_\epps(\mbx,\mby)$ then takes the form of weighted neighborhood graph $\G_{w_{\epps}}$, which we denote by the weight matrix $W_{i,j} = w_{\epps}(p_i,p_j), \, p_i,p_j \in \dataset$. In fact, \eqref{eq:discret_D_1} and \eqref{eq:discret_L_1} can be expressed in terms of matrix operations when $\mbx \in \dataset$ as done in Algorithm \ref{alg:graph_laplacian}.


\begin{algorithm}[tb]
   \caption{\computegraph}
   \label{alg:graph_laplacian}
\begin{algorithmic}
   \STATE {\bfseries Input:} Weight matrix $W$, bandwidth $\bw$, and $\lambda$
   \STATE $D \gets \text{diag}(W \1)$
   \STATE $\tilde{W} \gets D^{-\lambda} W D^{-\lambda}$
   \STATE $\tilde{D} \gets \text{diag}(\tilde{W} \1)$
   \STATE $L \gets (c\epps)^{-1}(\tilde{D}^{-1} \tilde{W} - I_n)$ 
   \STATE {\bfseries Return} $L$   
\end{algorithmic}
\end{algorithm}

\comment{
We note that the heat kernel is not the only option available for defining the weight matrix $W_n$ (see for example \cite{TingHJ10}). An important variation to the heat kernel is the truncated heat kernel discussed in Subsection \ref{ssec:nng}, where $k_{\epps}(p,p')=0$ if $||p-p'||^2>C$ where $C$ is usually taken to be a multiple of $\epps$. This procedure really defines two graphs, the ${0,1}$ neighborhood graph indexed by $||p-p'||^2>\epps$ and the similarity graph with weights $\exp(-||p-p'||^2/\epps)$ when two points are neighbors. }

\comment{ 
The use of a truncated kernel has two important consequences. First, the convergence of $\Lbw$ can be extended to non-compact manifolds \cite{HeiAudLux07} (provided additional conditions are imposed on the curvature of $\M$).
Second, the truncation induces sparsity on $K_n$ and $\Lbw$, which substantially reduces the computational complexity involved in estimating $h$.
} 

With $\Lbw$, the discrete analogue to \eqref{eq:lb-op}, clarified, we are now ready to introduce the central algorithm of this article.

\comment{This appraoch has three important advantages:
it is algorithm-agnostic; it immediately quantifies the distortion
resulting from the emebdding; and it recovers the intrinsic geometry
exactly as $\nsamp\rightarrow \infty$.}

\comment{An alternative to finding an isometric embedding of $(\M,g)$ is to recover the pushforward $\varphi^{*}g_{p}$ of $g$ through $f$ so that
$(f(\M),f^{*}g)$ retains all the intrinsic geometry of $(\M,g)$.
}

\comment{Although appealing in theory, it is very impractical to use Nash's
theorem to find $f$~\citep{Nas56}, as no constructive proof for the
theorem is known to date. Unless an alternate proof to Nash's theorem
is found, we cannot expect to use it to obtain an algorithm that will
achieve the isometric embedding of any data manifold into
$(\rrr^{d},\delta)$.}

\comment{\em Another point to make somewhere is that not all algorithms are
guaranteed to give an Atlas (examples?), but might still offer a
number of charts allowing for geometrical computations on part of
$\M$.}


\subsection{The \riememb~ Algorithm}

The input data for a manifold learning algorithm is a set of points
$\dataset = \{p_1,\dots,p_\nsamp\}\subset \M$ where $\M$ is an unknown
Riemannian manifold. Our \riememb~ algorithm takes as input, along with 
dataset $\dataset$, an embedding dimension $\dembed$ and an
embedding algorithm, chosen by the user, which we will denote by
\myemb. \riememb~ proceeds in four steps, the first three being
preparations for the key fourth step.

\benum
\item construct a weighted neighborhood graph $\G_{w_{\epps}}$
\item calculate the graph Laplacian $\Lbw$
\item map the data $p\in\dataset$ to $f_\nsamp(p)\in\rrr^\dembed$ by \myemb
\item apply the graph Laplacian $\Lbw$ to the coordinates $f_\nsamp$ to obtain the embedding metric $h$
\eenum
Figure \ref{fig:algo} contains the \riememb~ algorithm in pseudocode.
The subscript $\nsamp$ in the notation indicates that these are
discretized, ``sample'' quantities, (i.e. $f_n$ is a vector and $\Lbw$
is a matrix) as opposed to the continuous quantities (functions,
operators) that we were considering in the previous sections. 
We now move to describing each of the steps of the algorithm.

\comment{Our algorithm proceeds in four steps. First, we find a set of neighbors
for each data point $p$, and evaluating a similarity $S_\nsamp(p,p')$ for
each pair of points $p,p'$ that are neighbors. Second, we use these
pairwise similarities to obtain $\Lbw$, an estimate of the
Laplace-Beltrami operator $\LB$, by following the procedure described
in ~\cite{HeiAudLux07} (see also
~\cite{CoiLaf06,GinKol06,BelSunWan08}). Third, we select an embedding
for the manifold of interest using any one of the many existing
manifold learning algorithms. Any learning algorithm that provides a
smooth injective map $f:\rrr^\dhigh\rightarrow \rrr^\dembed$ can be
used. Finally, we compute the embedding metric of the manifold in the
coordinates of the chosen embedding.

The details of the algorithm are given in Figure \ref{fig:algo}.
} 

The first two steps, common to many manifold learning
algorithms, have already been described in subsections \ref{ssec:nng} and \ref{ssec:discret_problem} respectively.
\comment{
consists of finding a set of neighbors for each data point
$p$, and evaluating a weight matrix $W_\nsamp(p,p')$ for each pair of points
$p,p'$ that are neighbors. Here, we stress the point that the
bandwidth parameter $\bw$ determines the size of the neighborhood for
each point, and, hence, the density of the graph. The
adjacency/similarity matrix $W_\nsamp$ represents the heat kernel applied
to point differences $||p-p'||^2$.

Second, we use the weight matrix $W_\nsamp$ to obtain $\Lbw$ \eqref{eq:rwL2}, the estimator of the Laplace-Beltrami operator $\LB$, by following the procedure described in \cite{HeiAudLux07} (see also \cite{CoiLaf06,GinKol06,BelSunWan08}). This procedure includes renormalizing $K_\nsamp$ (Step 2, c, d) in order for $\Lbw$ to be a consistent estimator of $\Delta_\M$, even in the presence of non-uniformly sampled data.
} 
The third step calls for obtaining an embedding of the data points
$p\in\dataset$ in $\rrr^\dembed$. This can be done using any one of
the many existing manifold learning algorithms (\myemb), such as the
Laplacian Eigenmaps, Isomap or Diffusion Maps.

\comment{, provided that it
satisfies the requirements we discuss in \ref{sec:Consistency}. In
short, the algorithm must have a well-defined $h$, i.e. be an
embedding of $\M$, and $h_\nsamp$ must be consistent for $h$. 
smoothness?}

\comment{ The algorithm
may take other parameters than the data $\dataset$, $\dintri$ and $\dembed$ such as the bandwidth parameter $\bw$.
} 

At this juncture, we note that there may be overlap in the computations involved in the first three steps. Indeed, a large number of the common embedding algorithms, including
Laplacian Eigenmaps, Diffusion Maps, Isomap, LLE, and LTSA  use a
neighborhood graph and/or similarities in order to obtain an
embedding. In addition, Diffusion Maps and Eigemaps obtain an embedding for the eigendecomposition $\Lbw$ or a similar operator. While we define the steps of our algorithm in their most complete form, we encourage the reader to take advantage of any efficiencies that may result from avoiding to compute the same quantities multiple times. 

The fourth and final step of our algorithm consists of computing the
embedding metric of the manifold in the coordinates of the chosen
embedding. Step 4, applies the $\nsamp\times \nsamp$ Laplacian
matrix $\Lbw$ obtained in Step 2 to pairs
$f^i_{\nsamp},\,f^j_{\nsamp}$ of embedding coordinates of the data
obtained in Step 3. We use the symbol $\myprod$ to refer to the
elementwise product between two vectors. Specfically, for two vectors
$x,y\in \rrr^\nsamp$ denote by $x \myprod y$ the vector $z \in
\rrr^\nsamp$ with coordinates $z=(x_1y_1,\dots,x_\nsamp
y_\nsamp)$. This product is simply the usual function multiplication
on $\M$ restricted to the sampled points $\dataset\subset\M$. Hence,
equation \eqref{eq:H_tilde_numeric} is equivalent to applying equation
\eqref{eq:H} to all the points of $p\in\dataset$ at once. The result
are the vectors $\tilde{h}_{\nsamp}^{ij}$, each of which is an
$\nsamp$-dimensional vector, with an entry for each
$p\in\dataset$. Then, in Step 4, b, at each embedding point $f(p)$
the embedding metric $h_{\nsamp}(p)$ is computed as the matrix
(pseudo) inverse of $[\tilde{h}_{\nsamp}^{ij}(p)]^{ij=1:\dembed}$.

If the embedding dimension $\dembed$ is larger than the
manifold dimension $\dintri$, we will obtain the rank $\dintri$
embedding metric $h_n$; otherwise, we will obtain the Riemannian
metric $g_n$. For the former, $h_n^\dagger$ will have a theoretical 
rank $\dintri$, but numerically it might have rank between $\dintri$ and 
$\dembed$. As such, it is important to set to
zero the $\dembed-\dintri$ smallest singular values of
$h_n^\dagger$ when computing the pseudo-inverse. This is the key reason why $\dembed$ and $\dintri$ need
to be known in advance. Failure to set the smallest singular values to
zero will mean that $h_n$ will be dominated by noise. Although
estimating $\dintri$ is outside the scope of this work, it is
interesting to note that the singular values of $h_n^\dagger$ may offer
a window into how to do this by looking for a ``singular value gap''.

In summary, the principal novelty in our \riememb~ algorithm is its
last step: the estimation of the embedding metric $h$. The embedding
metric establishes a direct correspondence between geometric
computations performed using $(f(\M),h)$ and those performed directly
on $(\M,g)$ {\em for any embedding} $f$. Thus, once augmented with
their corresponding $h$, all embeddings become geometrically
equivalent to each other, and to the orginal data manifold $(\M,g)$.

\comment{
\benum
\item \label{it:graph0} Preprocessing
\item \label{it:L0} Estimate the discretized Laplace-Beltrami $\Lbw$ operator for the data
\item \label{it:embed0} Find an embedding $f_\nsamp$ of the data into $\rrr^\dembed$, with embedding coordinates $f_\nsamp = (f^1_{\nsamp},\dots,f^\dembed_{\nsamp})$ 
\item \label{it:h0} obtain estimates of $h_\nsamp^\dagger(p)$ and $h_\nsamp(p)$ for all data points $p$ as shown in Section \ref{sec:g-math}
\item[]\hspace{\backitem}{\bf Output} $(f_\nsamp(p),h_\nsamp(p))$ for all data points $p$
\eenum
} 

\begin{algorithm}[tb]
   \caption{{\pseudoinv}}
   \label{alg:pseudoin}
\begin{algorithmic}
   \STATE {\bfseries Input:} Embedding metric $\tilde{h}_\nsamp(p)$ and intrinsic dimension $\dintri$
   \STATE $[U,\Lambda]  \gets$ {\sc EigenDecomposition}($\tilde{h}_\nsamp(p)$) where $U$ is the matrix of column eigenvectors of $\tilde{h}_\nsamp(p)$ ordered by their eigenvalues $\Lambda$ from largest to smallest.
   \STATE $\Lambda \gets \Lambda(1:\dintri)$ (keep $\dintri$ largest eigenvalues) 
   \STATE $\Lambda^{\dagger} \gets \text{diag}(1/{\Lambda})$
   \STATE $h_\nsamp(p) \gets U \Lambda^{\dagger} U^t$ (obtain rank $\dintri$ pseudo-inverse of $\tilde{h}_\nsamp(p)$)
   \STATE {\bfseries Return} $h_\nsamp(p)$   
\end{algorithmic}
\end{algorithm}

\begin{algorithm}[tb]
   \caption{\riememb}
   \label{fig:algo}
\begin{algorithmic}
\STATE {\bfseries Input:}  $\dataset$ as set of $\nsamp$ data points in $\rrr^\dhigh$, $\dembed$ the number of dimensions of the embedding, $\dintri$ the intrinsic dimension of the manifold, $\bw$ the bandwidth parameter, and \myemb$(\dataset,\,\dembed,\bw)$ a manifold learning algorithm, that outputs $\dembed$ dimensional embedding coordinates
\benum 
\item \label{it:sim} Construct the weighted neighborhood graph with weight matrix $W$ given by
$W_{i,j}=\exp{(-\frac{1}{\epps}||p_i-p_j'||^2)}$ for every points $p_i,p_j \in \dataset$.
\item Construct the Laplacian matrix $\Lbw$ using Algorithm \ref{alg:graph_laplacian} with input $W$, $\bw$, and $\lam=1$.
\item Obtain the {\em  embedding coordinates} $f_n(p)=(f^1_{n}(p),\ldots,f^\dembed_{n}(p))$ of each point $p\in \dataset$ by 
\[
[f_n(p)]_{p\in \dataset}\;=\;\mbox{\myemb}(\dataset,\dembed,\dintri,\bw)
\]
\item \label{it:emb} Calculate the {\em embedding metric} $h_n$ for each point 
	\benum

	\item \label{it:H_d} For $i$ and $j$ from 1 to $\dembed$ calculate the column vector $\tilde{h}_\nsamp^{ij}$ of dimension $\nsamp = |\dataset|$ by 
	\beq \label{eq:H_tilde_numeric}
	\tilde{h}_n^{ij}\;=\;\frac{1}{2}\left[\Lbw(f^i_{n}\myprod f^j_{n})-f^i_{n}\myprod(\Lbw f^j_{n})-f^j_{n}\myprod(\Lbw f^i_{n})\right]
	\eeq 
	\item For each data point $p\in \dataset$, form the matrix $\tilde{h}_{\nsamp}(p)=[\tilde{h}_{\nsamp}^{ij}(p)]_{i,j\in {1,\dots,\dembed}}$. The embedding metric at $p$ is then given by $h_\nsamp(p)=\mbox{\pseudoinv}(\tilde{h}_\nsamp(p),\dintri)$ 
	\eenum
\item[]\hspace{\backitem}{\bf Return} $(f_\nsamp(p),h_\nsamp(p))_{p\in \dataset}$
\eenum
\end{algorithmic}
\end{algorithm}


\comment{ In the algorithm
description, we adopt a pseudocode-like notation, that emphasizes the
data type of each variable. Scalars, vectors and matrices are denoted
respectively by lower case, upper case and bold upper case
literals. In particular, the embedding coordinates of point $p$ are
denote $f_\nsamp(p)$, and the estimate of the Riemannian metric at $p$ is
$\matrh_p$. We deviate from this notation in one respect: we do not
introduce an additional notation for the coordinates of the input
data. Thus, we let $p$ denote both the index of of a data point in the
data set $\dataset$, and its coordinates in $\rrr^\dhigh$. The latter meaning is only used in Step \ref{it:sim} of the \riememb~ algorithm.} 



\subsection{Computational Complexity \label{ss_algo_comp_compl}}

Obtaining the neighborhood graph involves computing $\nsamp^2$
distances in $\dhigh$ dimensions. If the data is high- or very
high-dimensional, which is often the case, and if the sample size is
large, which is often a requirement for correct manifold recovery,
this step could be by far the most computationally demanding of the
algorithm. However, much work has been devoted to speeding up this
task, and approximate algorithms are now available, which can run in
linear time in $\nsamp$ and have very good accuracy
(\cite{ram2010ltaps}). In any event, this computationally intensive
preprocessing step is required by all of the well known embedding
algorithms, and would remain necessary even if one's goal were solely to
embed the data, and not to compute the Riemannian metric.

Step 2 of the algorithm operates on a sparse $\nsamp \times \nsamp$ matrix. If the neighborhood
size is no larger than $k$, then it will be of order $\bigOO(\nsamp k)$, and $\bigOO(\nsamp^2)$ otherwise. 

The computation of the embedding in Step 3 is algorithm-dependent. For
the most common algorithms, it will involve eigenvector computations.
These can be performed by Arnoldi iterations that each take
$\bigOO(\nsamp^2\dembed)$ computations, where $\nsamp$ is the sample size, and $\dembed$ is the embedding dimension or, equivalently, the number of eigenvectors used. This step, or a variant thereof, is also a component of many embedding algorithms. 
\comment{A notable exception is Isomap, which iterates over this step, 
and is thus significantly slower than e.g Diffusion Map. 
}

Finally, the newly introduced Step 4 involves obtaining an
$\dembed\times \dembed$ matrix for each of the $\nsamp$ points, and
computing its pseudoinverse. Obtaining the $\tilde{h}_n$
matrices takes $\bigOO(\nsamp^2)$ operations ($\bigOO(\nsamp k)$ for sparse $\Lbw$ matrix) 
times $\dembed \times \dembed$ entries, for a total of
$\dembed^2\nsamp^2$ operations.  The $\nsamp$ SVD and pseudoinverse calculations
take order $\dembed^3$ operations.

Thus, finding the Riemannian metric makes a small contribution to the computational burden of finding the embedding. The overhead is quadratic in the data set size $\nsamp$ and embedding dimension $\dembed$, and cubic in the intrinsic dimension $\dintri$. 


\section{Experiments~\label{sec:App}}

The following experiments on simulated data demonstrate the
\riememb~ algorithm and highlight a few of its immediate applications.

\subsection{Embedding Metric as a Measure of Local Distortion}

\begin{figure}
\setlength{\picwi}{0.5\textwidth}

\begin{tabular}{cc}
{\small Original data} & {\small Embedding with $h$ estimates}\\
\includegraphics[width=0.9\picwi]{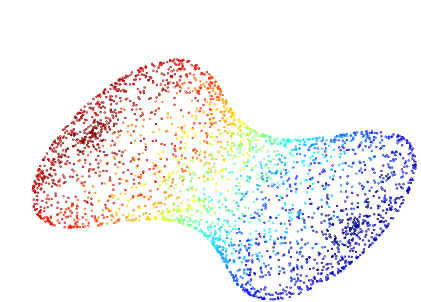}
&
\multirow{4}{*}{
\includegraphics[width=\picwi,height=0.85\picwi,trim=0 0 0 100]{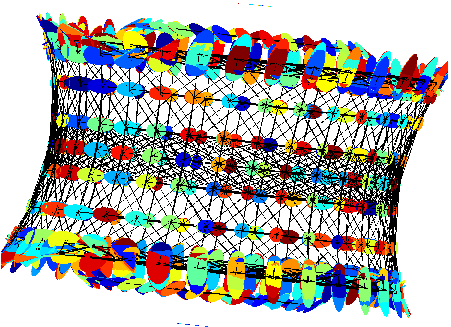}
}
\\
\\
{\small $h$ estimates, detail}&\\
\includegraphics[width=\picwi]{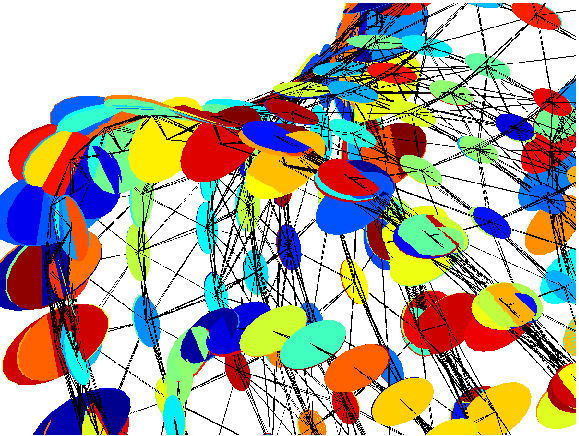}&\\
\end{tabular}
\caption{\label{fig:hourglass-distortion}
Estimation of $h$ for a 2D hourglass-shaped manifold in 3D space. The
embedding is obtained by Diffusion Maps. The ellipses attached to
each point represent the embedding metric $h_{\nsamp}$ estimate for this
embedding. At each data point $p\in\dataset$, $h_{\nsamp}(p)$ is a $3\times 3$ symmetric semi-positive definite matrix of rank
2.  Near the ``girth'' of the hourglass, the ellipses are round,
showing that the local geometry is recovered
correctly. Near the top and bottom of the hourglass, the elongation of the
ellipses indicates that distances are larger along the direction of elongation than the embedding suggests. For clarity, in the embedding
displayed, the manifold was sampled regularly and sparsely. The black
edges show the neigborhood graph $\G$ that was used. For all images in this figure, the color code has no particular meaning.}
\end{figure}

The first set of experiments is intended to illustrate the output of the \riememb~
algorithm. Figure \ref{fig:hourglass-distortion} shows the embedding
of a 2D hourglass-shaped manifold. Diffusion Maps, the embedding algorithm we used (with $\dembed=3$, $\lambda = 1$)
distorts the shape by excessively flattening the top and bottom. \riememb~ outputs a $\dembed\times \dembed$ quadratic
form for each point $p\in \dataset$, represented as
ellipsoids centered at $p$. 
Practically, this means that
the ellipsoids are flat along one direction $T_{f_{\nsamp}(p)}f_{\nsamp}(\M)^\perp$, and two-dimensional because $\dintri=2$, i.e. $h_{\nsamp}$ has rank 2. If the embedding correctly recovered the local geometry, $h_{\nsamp}(p)$ would equal $I_3|_{f_{\nsamp}(\M)}$, the identity matrix restricted to  $T_{f_{\nsamp}(p)}f_{\nsamp}(\M)$: it would define a circle in the tangent plane of $f_{\nsamp}(\M)$, for each $p$. We see that this is the case in the girth area of the
hourglass, where the ellipses are circular. Near the top and bottom,
the ellipses' orientation and elongation points in the direction
where the distortion took place and measures the amount of (local)
correction needed. 

The more the space is compressed in a given direction, the more
elongated the embedding metric ``ellipses'' will be, so as to make each
vector ``count for more''. Inversely, the more the space is stretched,
the smaller the embedding metric will be. This is illustrated in
Figure~\ref{fig:hourglass-distortion}.

We constructed the next example to demonstrate how our method applies to the popular Sculpture Faces data set. This data set was introduced by \cite{TenDeS00} along with Isomap as a prototypical example of how to recover a simple low dimensional manifold embedded in a high dimensional space. 
Specifically, the data set consists of $\nsamp = 698$ $64 \times 64$ gray images of faces. The faces are allowed to vary in three ways: the head can move up and down; the head can move right to left; and finally the light source can move right to left. With only three degrees of freedom, the faces define a three-dimensional manifold in the space of all $64 \times 64$ gray images. In other words, we have a three-dimensional manifold $\M$ embedded in $[0,1]^{4096}$. 

\begin{sidewaysfigure}
\begin{center}
\begin{tabular}{cc}
\includegraphics[trim= 1.5cm 1.5cm 1.5cm 1.5cm,clip=true,width=0.5\textwidth]{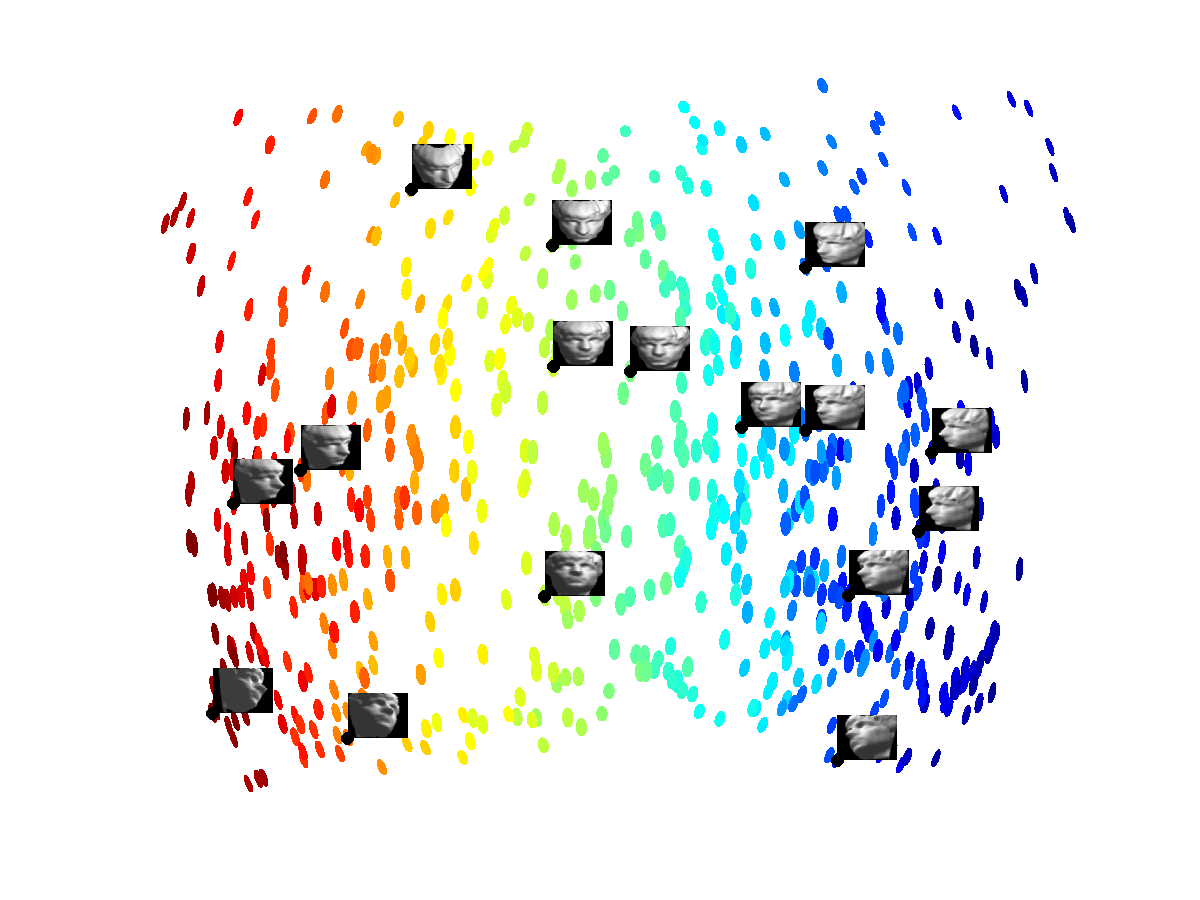}  &
\includegraphics[trim= 1.5cm 1.5cm 1.5cm 1.5cm,clip=true,width=0.5\textwidth]{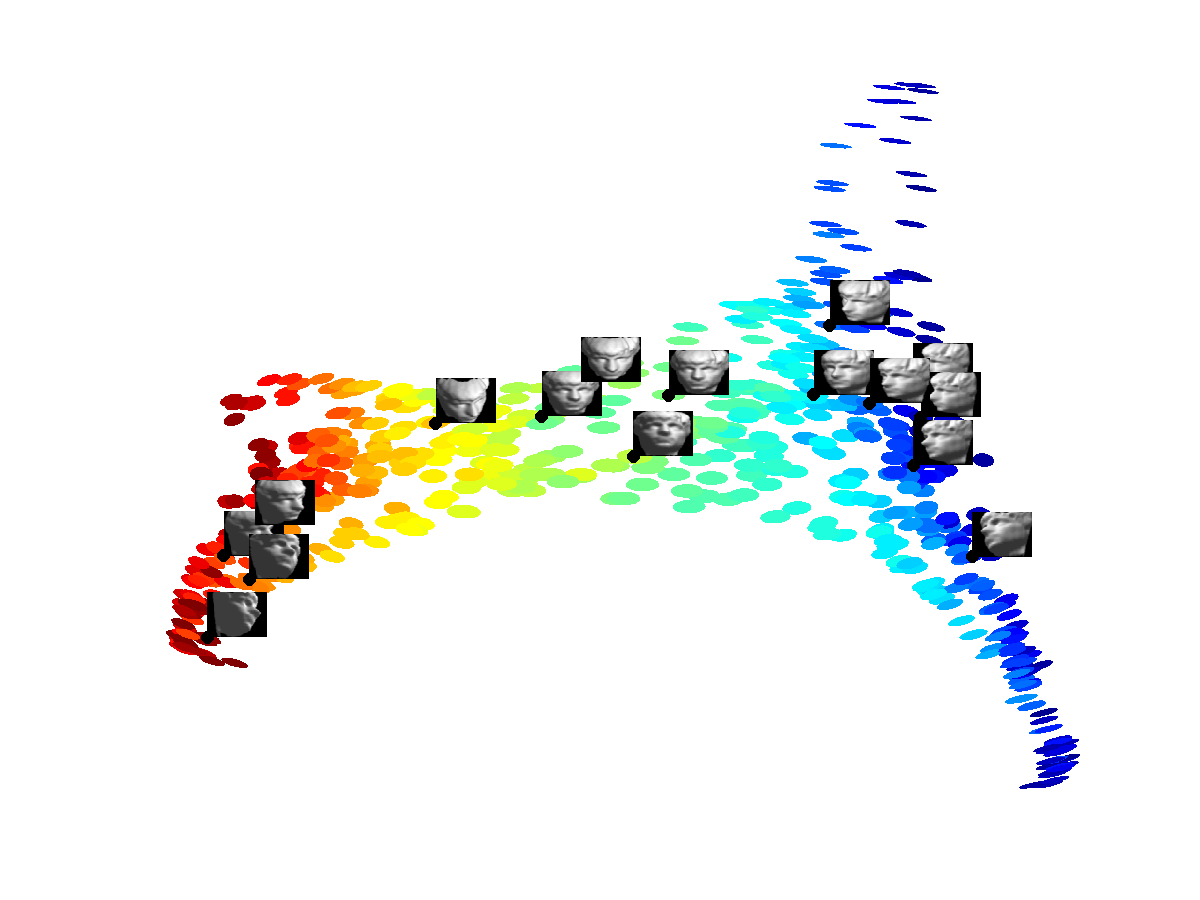} \\
(a) & (b) \\ 

\multicolumn{2}{c}{
 \includegraphics[trim= 1.5cm 1.5cm 1.5cm 1.5cm,clip=true,width=0.5\textwidth]{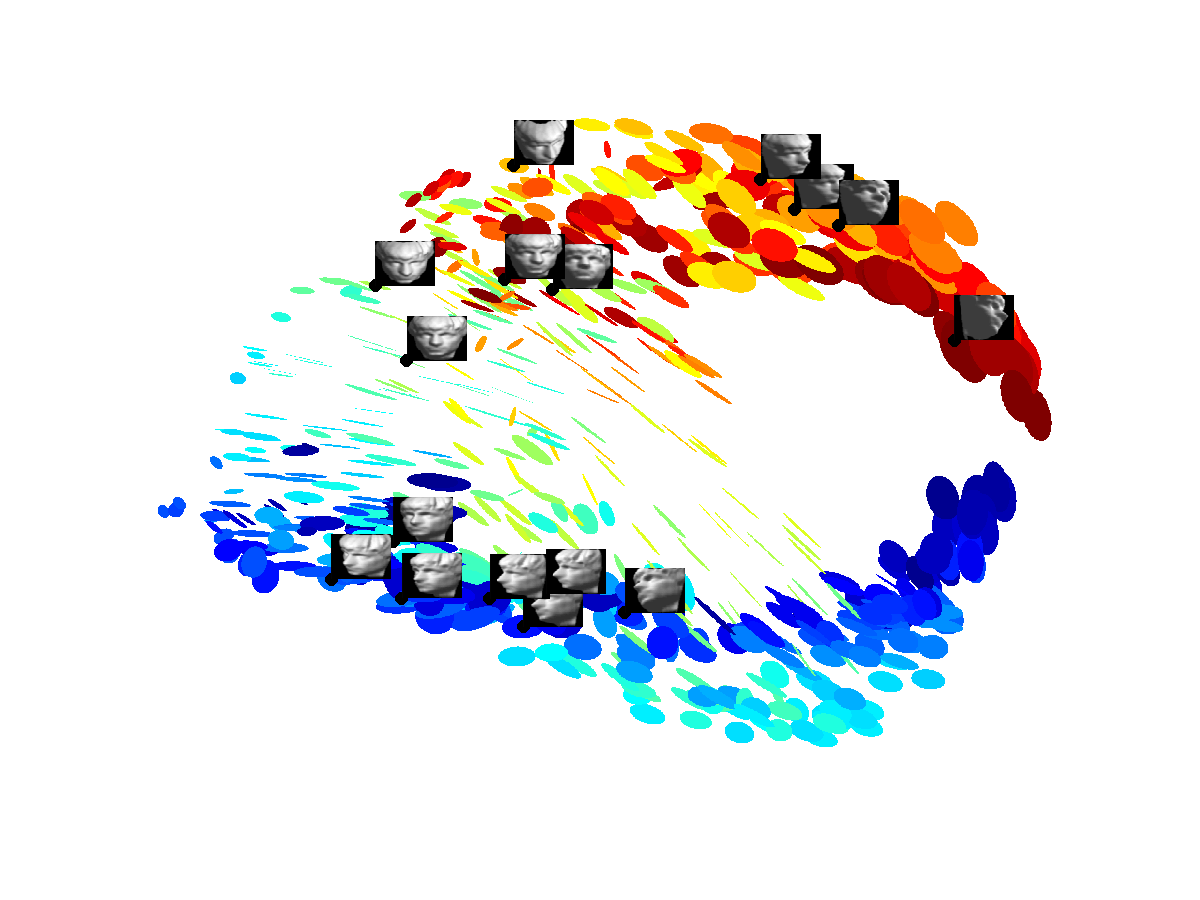} } \\
\multicolumn{2}{c}{ (c)}\\
\end{tabular}
\caption{Two-dimensional visualization of the faces manifold, along with embedding. The color scheme corresponds to the left-right motion of the faces. The embeddings shown are: (a) Isomap, (b) LTSA, and Diffusion Maps ($\lambda = 1$) (c). \label{fig:facesIsoLTSADM}}
\end{center}
\end{sidewaysfigure}

As expected given its focus on preserving the geodesic distances, the Isomap seems to recover the simple rectangular geometry of the data set, as Figure \ref{fig:facesIsoLTSADM} shows. LTSA, on the other hand, distorts the original data, particularly in the corners, where the Riemannian metric takes the form of thin ellipses. Diffusion Maps distorts the original geometry the most. The fact that the embedding for which we have theoretical guarantees of consistency causes the most distortion highlights, once more, that consistency provides no information about the level of distortion that may be present in the embedding geometry. 

\comment{Because one can only expect $\M$ to be a diffeomorphism of the parameter space, in figure \ref{fig:facesIsoLTSADM} pitch and yaw, it is important to stress that, {\em a priori}, it is not obvious which one of the embeddings recovers the geometry accurately. This is especially true when comparing Isomap and LTSA, since the fact that the parameter space (defined by the face and light source orientation) is a rectangular prism does not guarantee that the resulting manifold will have the exact same shape. Therefore, the Isomap and LTSA could easily be considered equally valid representations of $\M$. In fact, when LTSA was introduced in \cite{ZhangZ:04}, the authors considered this particular data set, and at no point did they suggest that LTSA's performance was inadequate in comparison to Isomap. It is truly only by computing $h_{\nsamp}$ that we can appreciate the distortion induced by LTSA on this data set. 
} 



\begin{figure} 
\begin{center}
\begin{tabular}{cc}
\includegraphics[trim= 3cm 2cm 3cm 2cm,clip=true,width=0.5\textwidth]{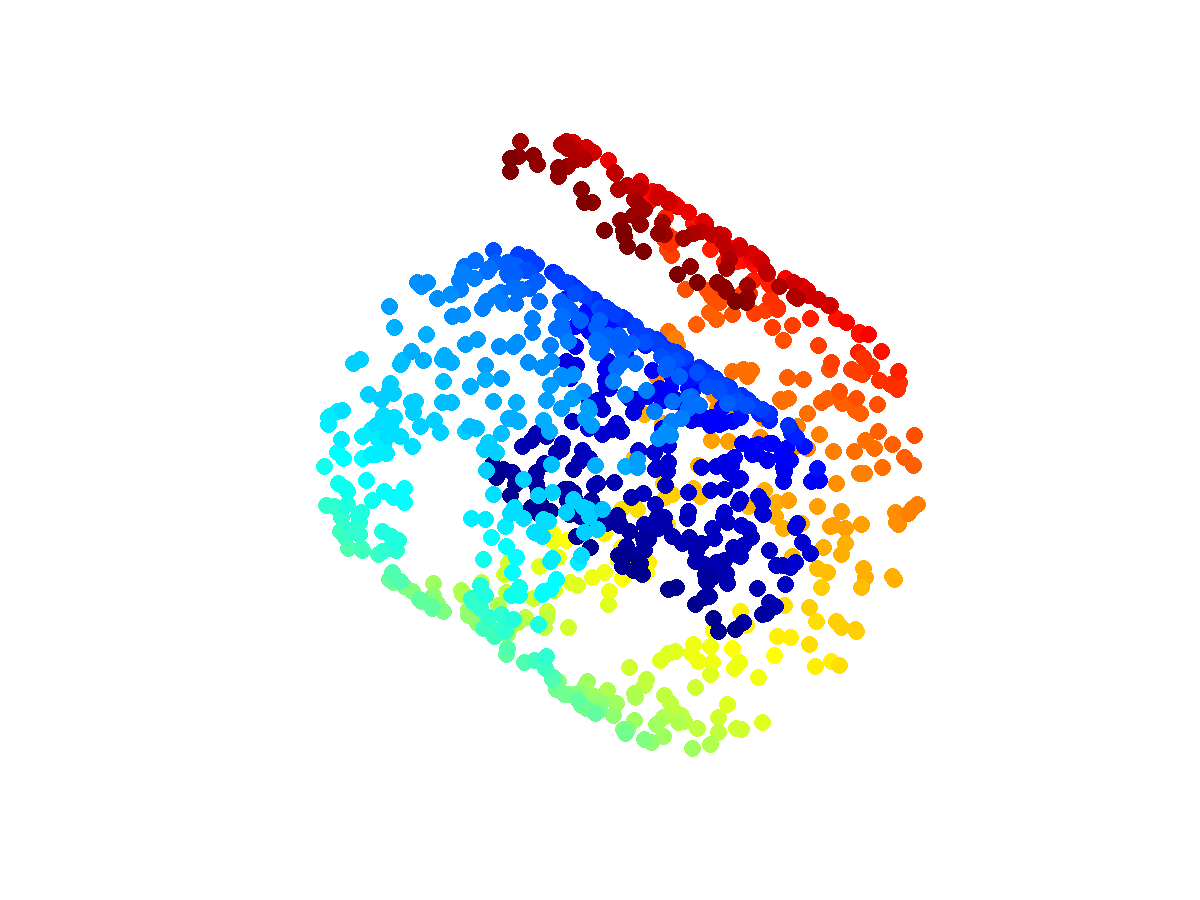} &
 \includegraphics[width=0.5\textwidth]{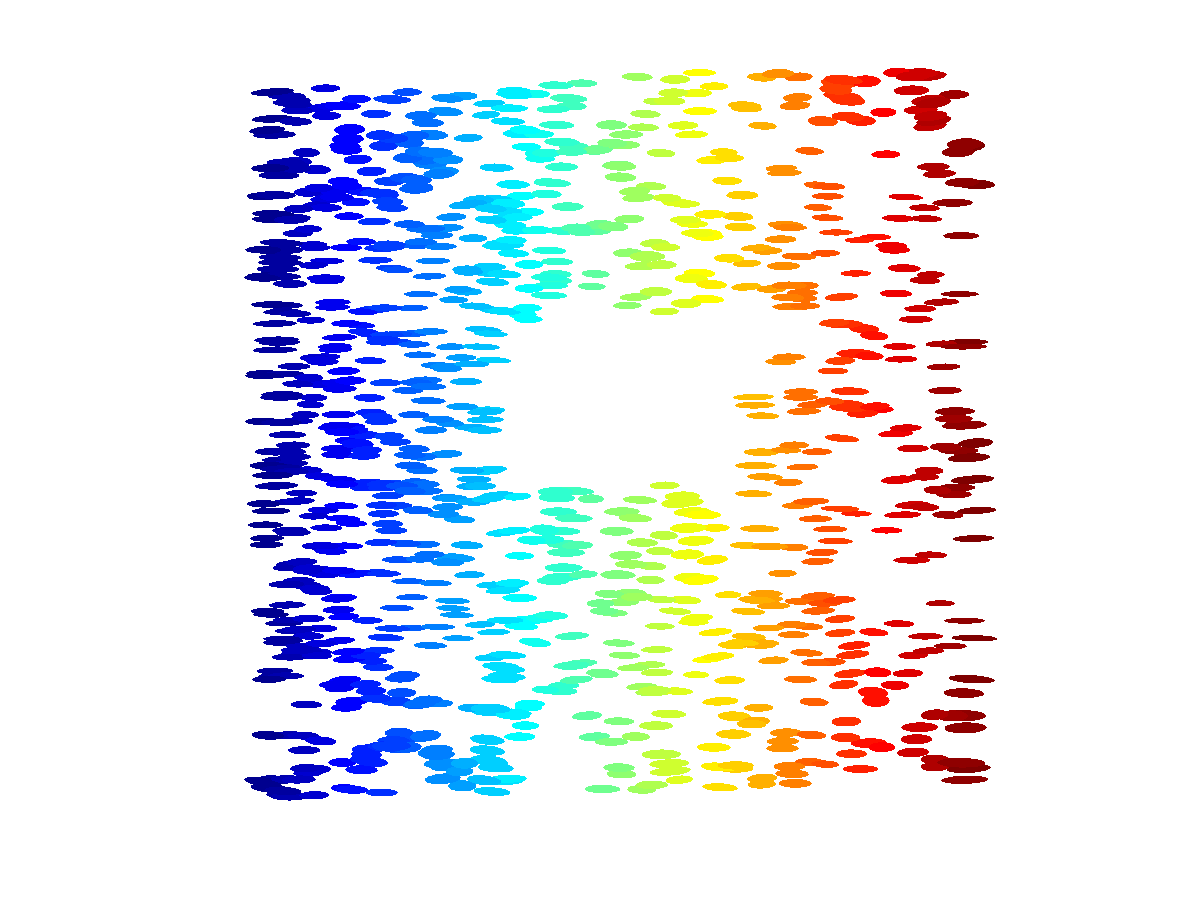} \\ 
(a) & (b) 
\end{tabular}
\end{center}
\caption{\label{fig:ltsaSwiss}
(a) Swissroll with a hole in $\rrr^3$. (b) LTSA embedding of the manifold in $\rrr^2$ along with metric. 
}
\end{figure}

Our next example, Figure \ref{fig:ltsaSwiss}, shows an almost isometric reconstruction of a common example, the Swiss roll with a rectangular hole in the middle. This is a popular test data set because many algorithms have trouble dealing with its unusual topology. In this case, the LTSA recovers the geometry of the manifold up to an affine transformation. This is evident from the deformation of the embedding metric, which is parallel for all points in Figure  \ref{fig:ltsaSwiss} (b). 

One would hope that such an affine transformation of the correct geometry would be easy to correct; not surprisingly, it is. In fact, we can do more than correct it: for any embedding, there is a simple transformation that turns the embedding into a local isometry. Obviously, in the case of an affine transformation, locally isometric implies globally isometric. We describe these transformations along with a few two-dimensional examples in the context of data visualization in the following section. 



\subsection{Locally Isometric Visualization}
\label{sec:visualization}

Visualizing a manifold in a way that preserves the manifold geometry
means obtaining an isometric embedding of the manifold in 2D or 3D. This is obviously not possible for all manifolds; in particular, only flat
manifolds with intrinsic dimension below 3 can be ``correctly
visualized'' according to this definition. This problem has been long known in cartography: a wide variety of {\em
cartographic projections} of the Earth have been developed to map parts of the 2D sphere onto a
plane, and each aims to preserve a different family of geometric
quantities. For example, projections used for navigational, meteorological or topographic charts focus on maintaining angular relationships and accurate shapes over small areas; projections used for radio and seismic mapping focus on maintaining accurate distances from the center of the projection or along given lines; and projections used to compare the size of countries focus on maintaining accurate relative sizes (\cite{Syd87}). 

While the \riememb~ algorithm is a general solution to preserving
intrinsic geometry for all purposes involving calculations of
geometric quantities, it cannot immediately give a general solution to
the visualization problem described above. 

However, it offers a natural way of producing {\em locally} isometric
embeddings, and therefore locally correct visualizations for two- or
three-dimensional manifolds. The procedure is based on the
transformation of the points that will guarantee that the embedding is
the identity matrix.

\begin{boxit}
\benum
\item[]\hspace{\backitem}{\bf Given} $(f_{\nsamp}(\dataset),h_{\nsamp}(\dataset))$ Metric Embedding of $\dataset$ 
\item Select a point $p\in \dataset$ on the manifold 
\item Transform coordinates
$\tilde{f}_{\nsamp}(p')\,\leftarrow\,h_{\nsamp}^{-1/2}(p)f_{\nsamp}(p')$ for all
$p'\in \dataset$
\item[]\hspace{\backitem}{\bf Display} $\dataset$ in coordinates
$\tilde{f}_{\nsamp}$
\eenum 
\end{boxit}

As mentioned above, the transformation $\tilde{f}_{\nsamp}$ ensures that the
embedding metric of $\tilde{f}_{\nsamp}$ is given by $\tilde{h}_{\nsamp}(p)=I_{\dembed}$,
i.e. the unit matrix at $p$\footnote{Again, to be accurate,
$\tilde{h}_{\nsamp}(p)$ is the restriction of $I_\dembed$ to
$T_{\tilde{f}_{\nsamp}(p)}\tilde{f}_{\nsamp}(\M)$.}. As $h$ varies smoothly on the
manifold, $\tilde{h}_{\nsamp}$ should be close to $I_{\dembed}$ at points
near $p$, and therefore the embedding will be approximately isometric
in a neighborhood of $p$. \comment{The generalization of the above
normalization to embeddings and embedding metrics $h_{\nsamp}$ is
straightforward.}

\begin{figure} 
\vspace{-0.75 in}
\hspace{-1em}\includegraphics[width=0.5\textwidth]{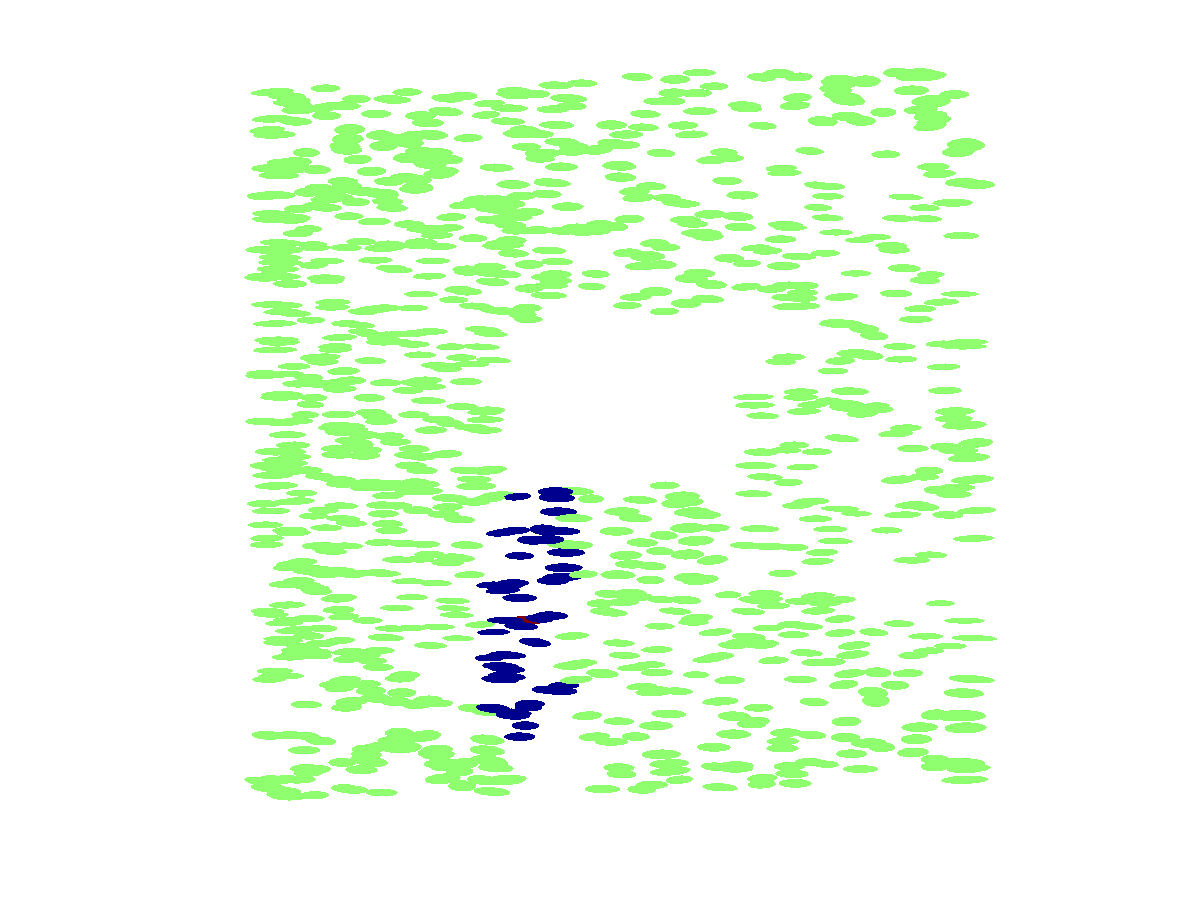}
\includegraphics[width=0.5\textwidth]{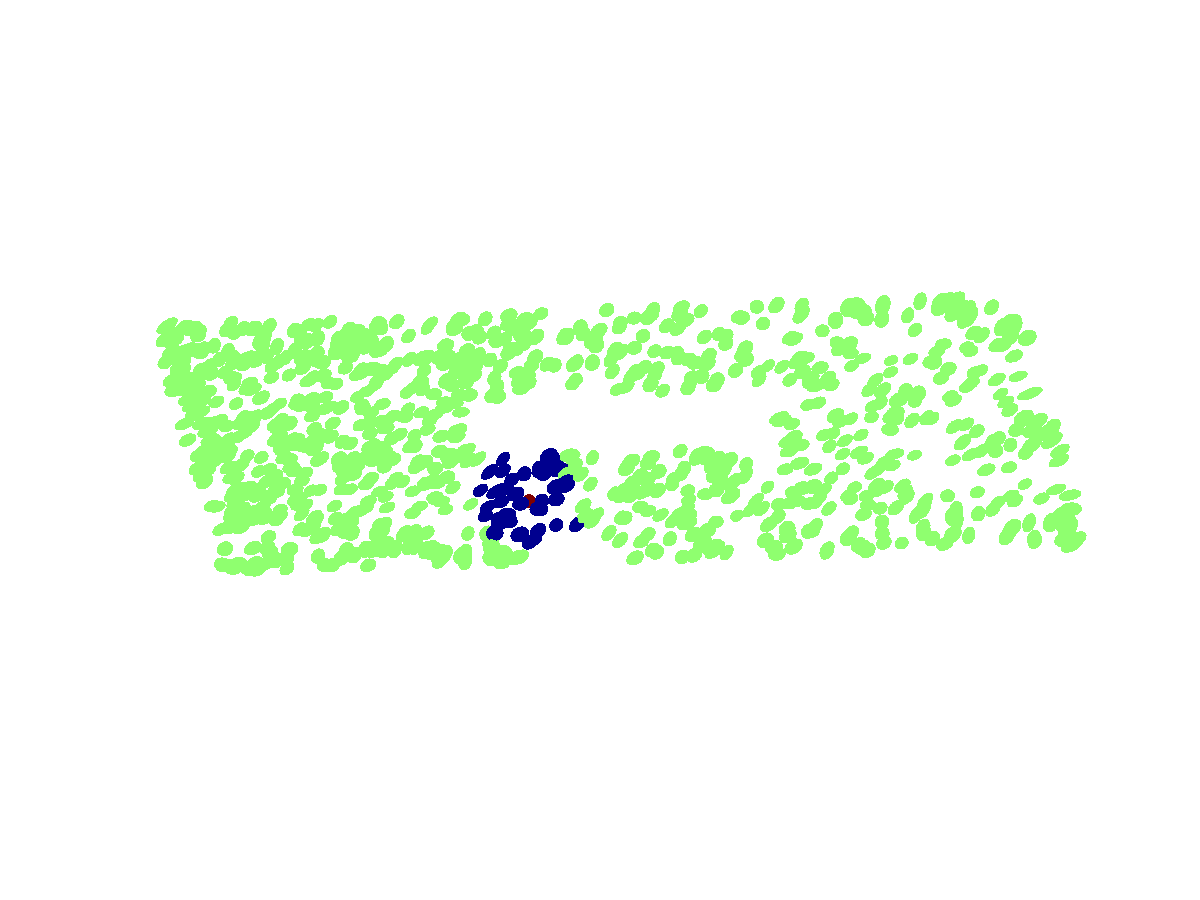}
\vspace{-0.25 in}
\begin{tabular}{cc}
\reflectbox{\includegraphics[width=0.45\textwidth]{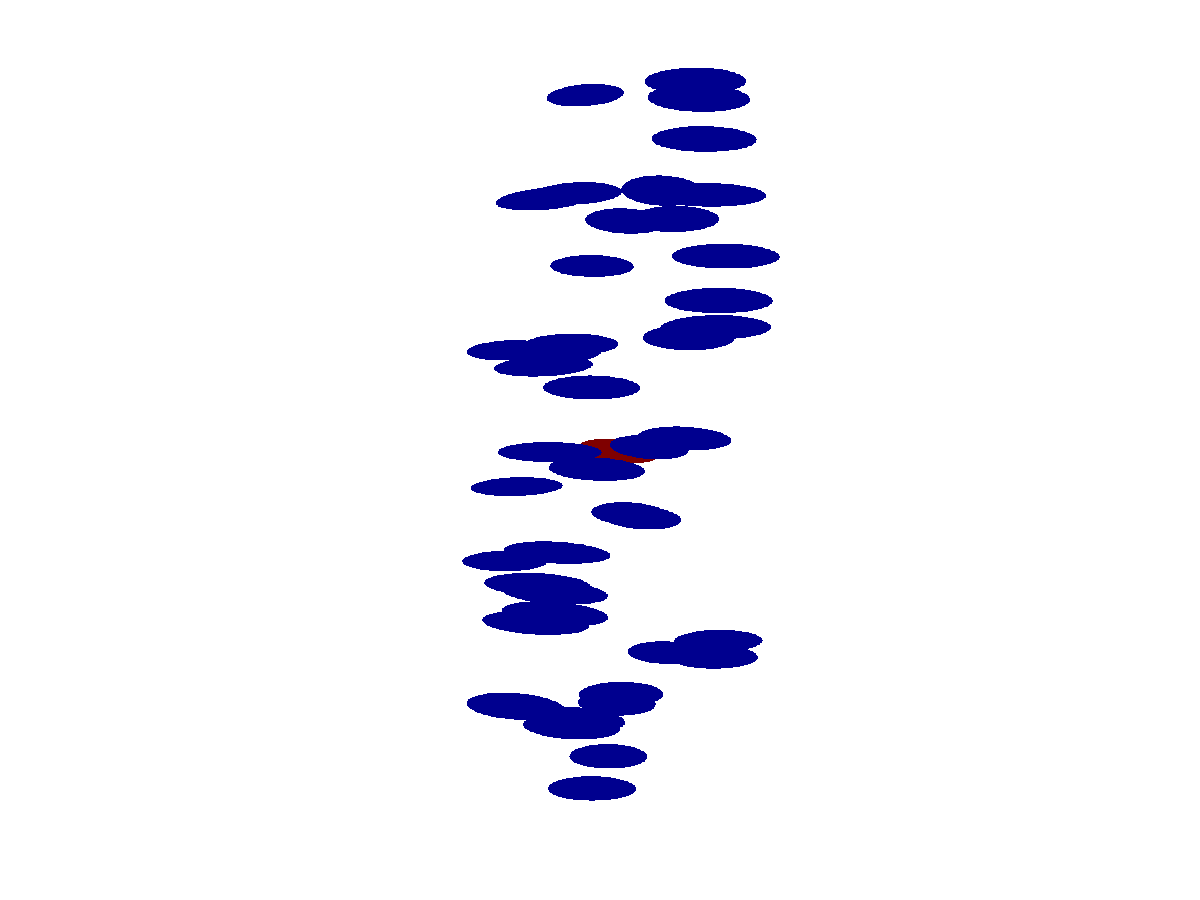}}&
\includegraphics[width=0.5\textwidth]{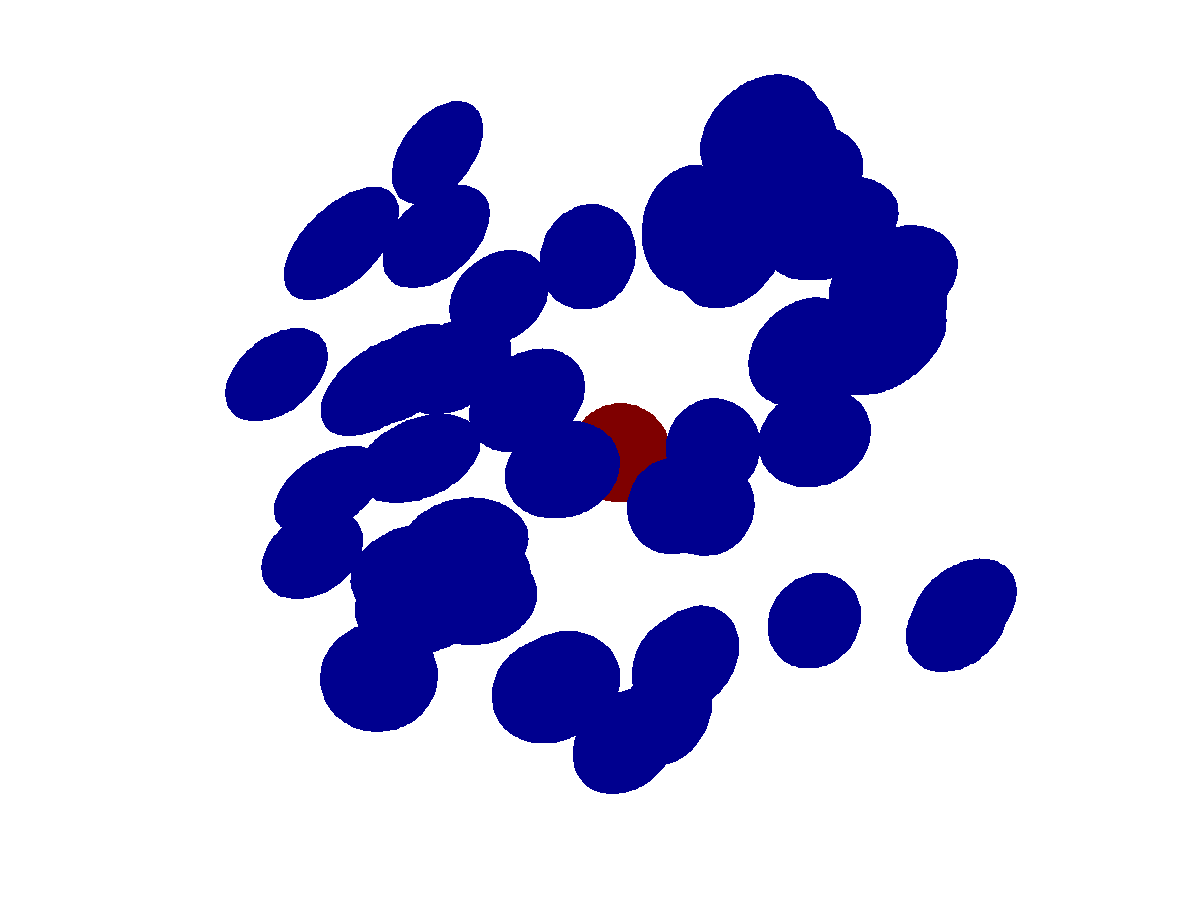}\\
\includegraphics[width=0.5\textwidth]{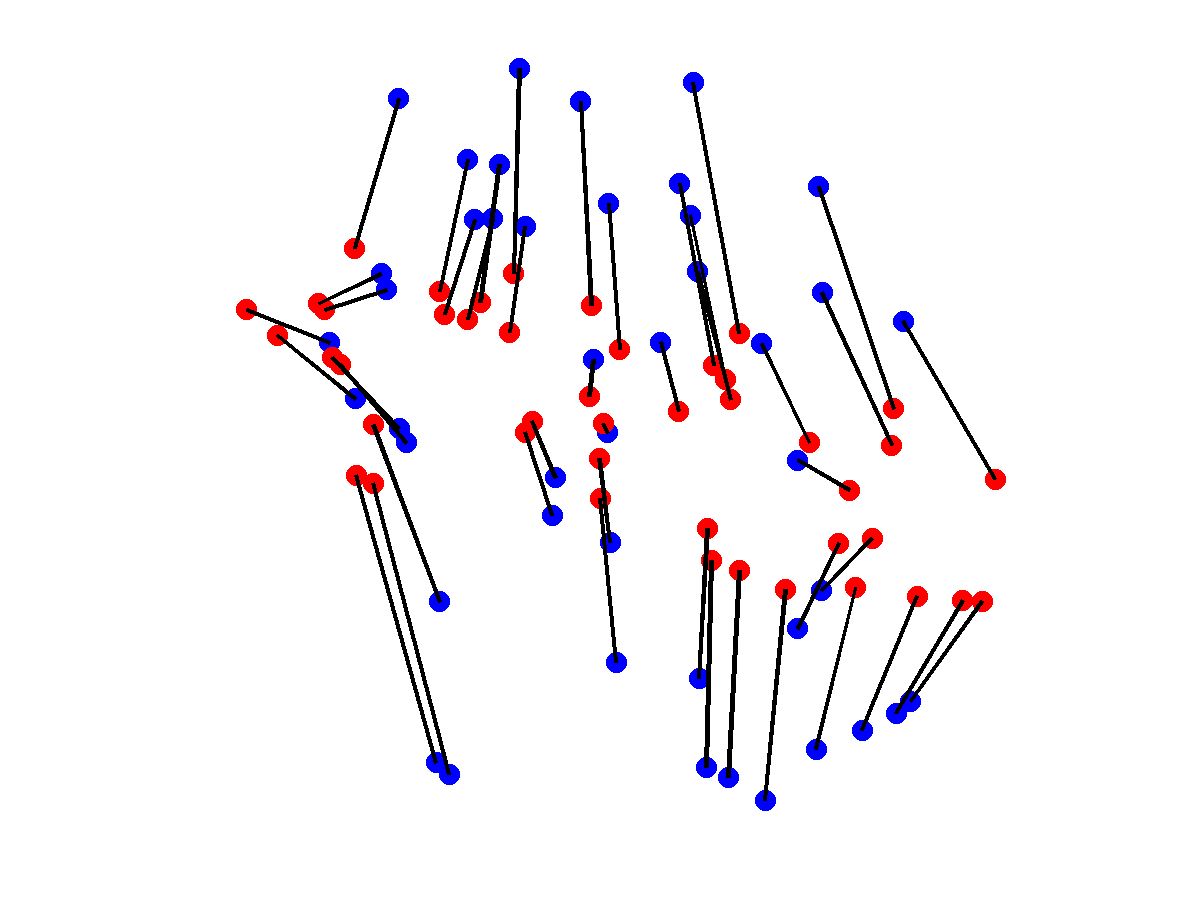}&
\includegraphics[width=0.49\textwidth]{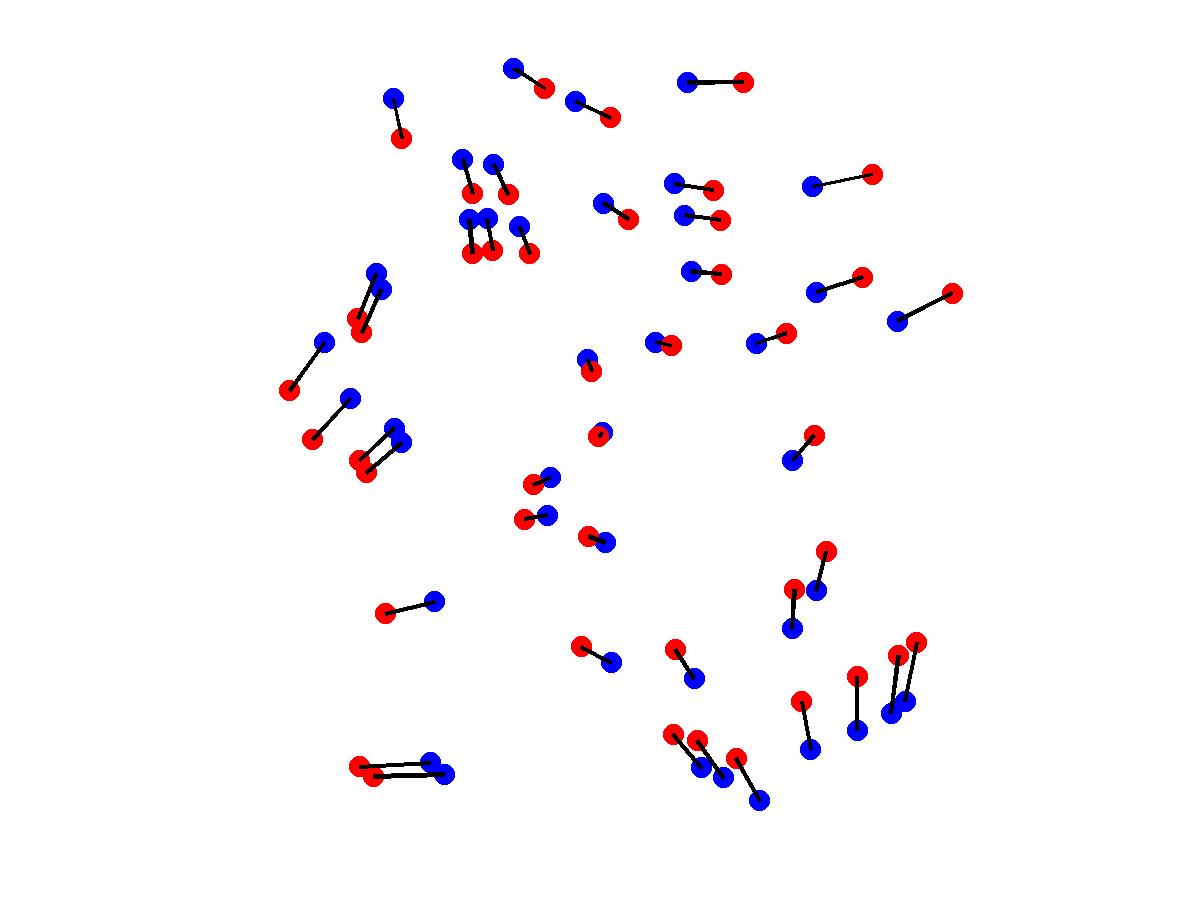}\\
\textcolor{blue}{$\bullet$} {\small original} &
\textcolor{red}{$\bullet$} {\small embedding} \\
\textcolor{white}{$\bullet$} {} &
\textcolor{white}{$\bullet$} {} \\

\end{tabular}

\caption{\label{fig:normalized-ltsa} 
Locally isometric visualization for the Swiss roll with a rectangular
hole, embedded in $\dintri=2$ dimensions by LTSA. Top left: LTSA
embedding with selected point $p$ (red) and its neighbors
(blue). Top right: locally isometric embedding. Middle left:
Neighborhood of $p$ for the LTSA embedding. Middle right:
Neighborhood of $p$ for the locally isometric embedding. Bottom left: Procrustes between the neighborhood of $p$ for the LTSA embedding and the original manifold projected on $T_p\M$; dissimilarity measure: $D=0.30$. Bottom right: Procrustes between the locally isometric embedding and the original manifold; dissimilarity measure: $D=0.02$.}
\end{figure}

\begin{figure} 
\vspace{-1 in}
\hspace{-0.25 in}
\includegraphics[width=0.5\textwidth]{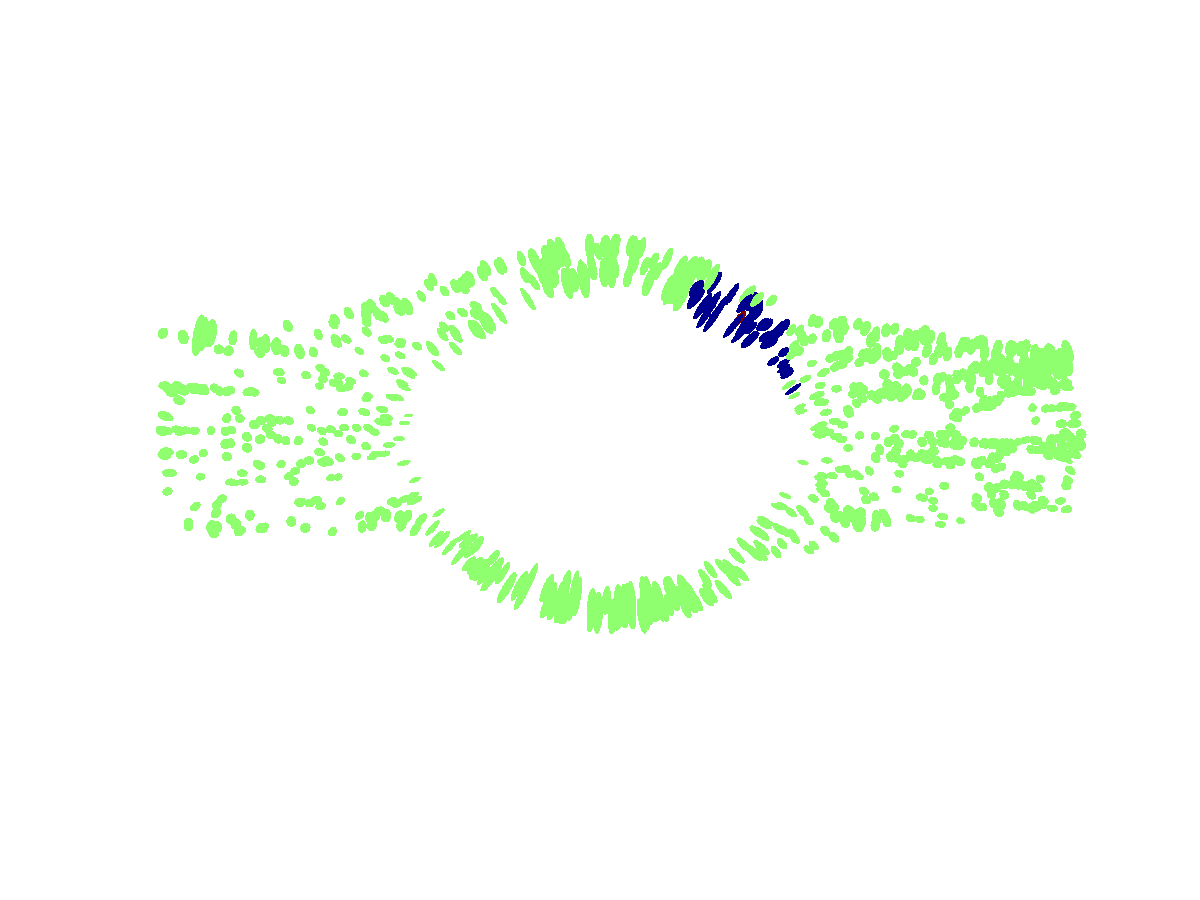}
\includegraphics[width=0.5\textwidth]{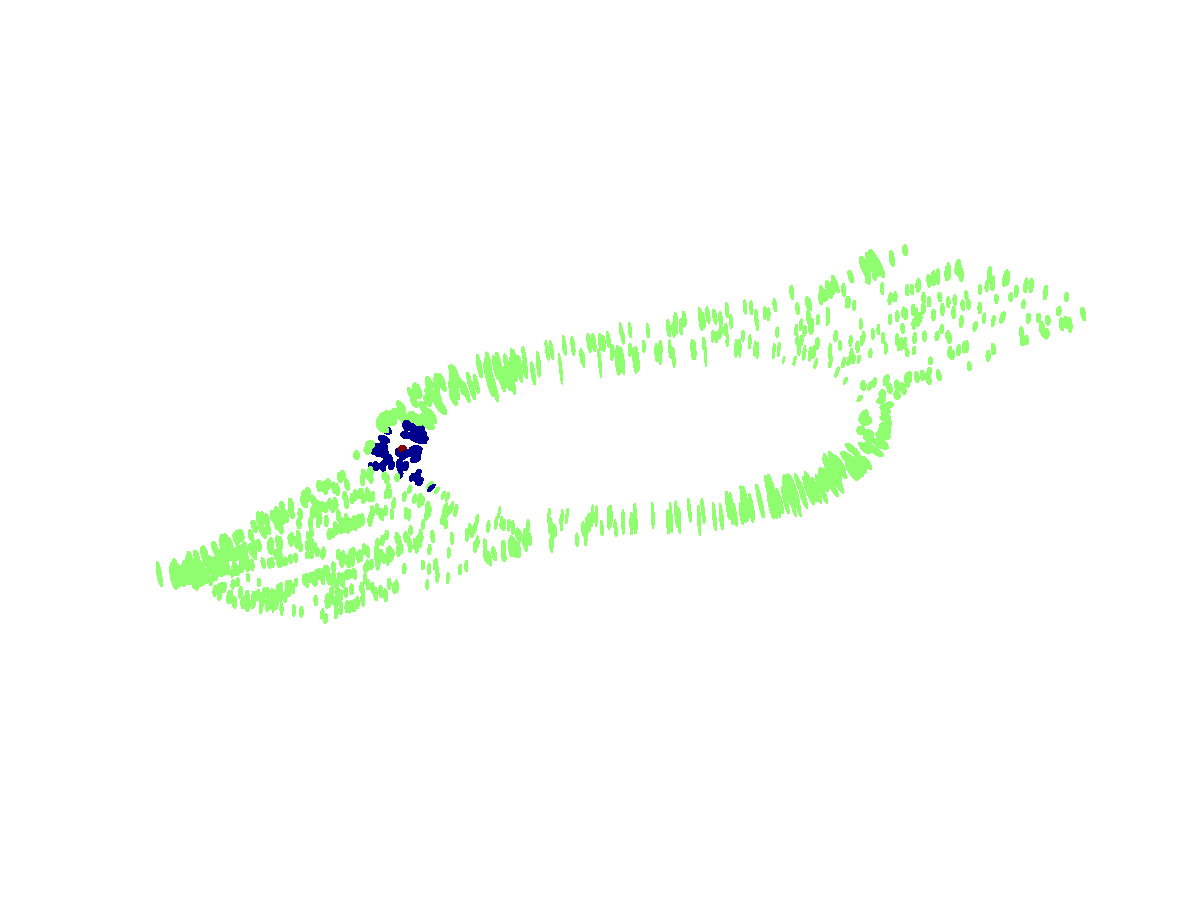}
\begin{tabular}{cc}
\reflectbox{\includegraphics[width=0.45\textwidth]{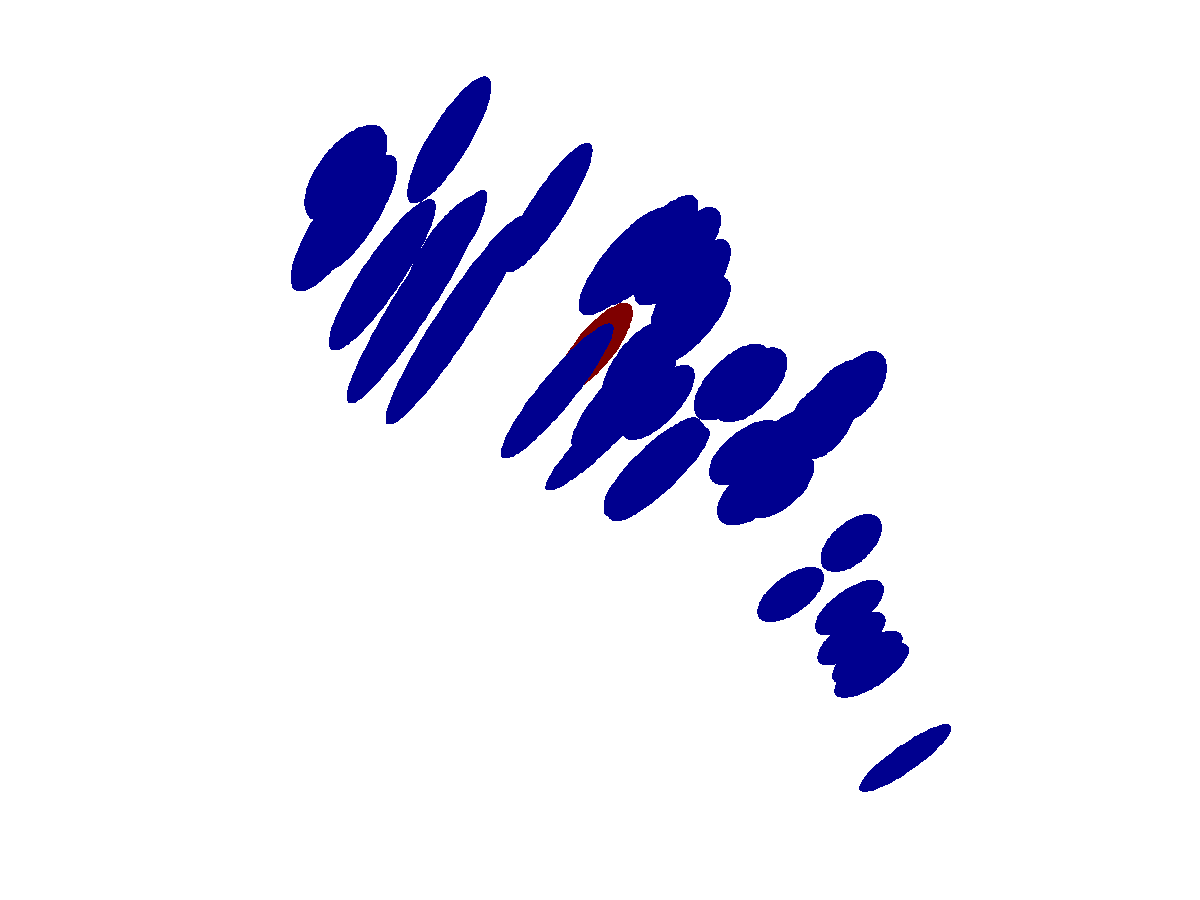}}&
\includegraphics[width=0.5\textwidth]{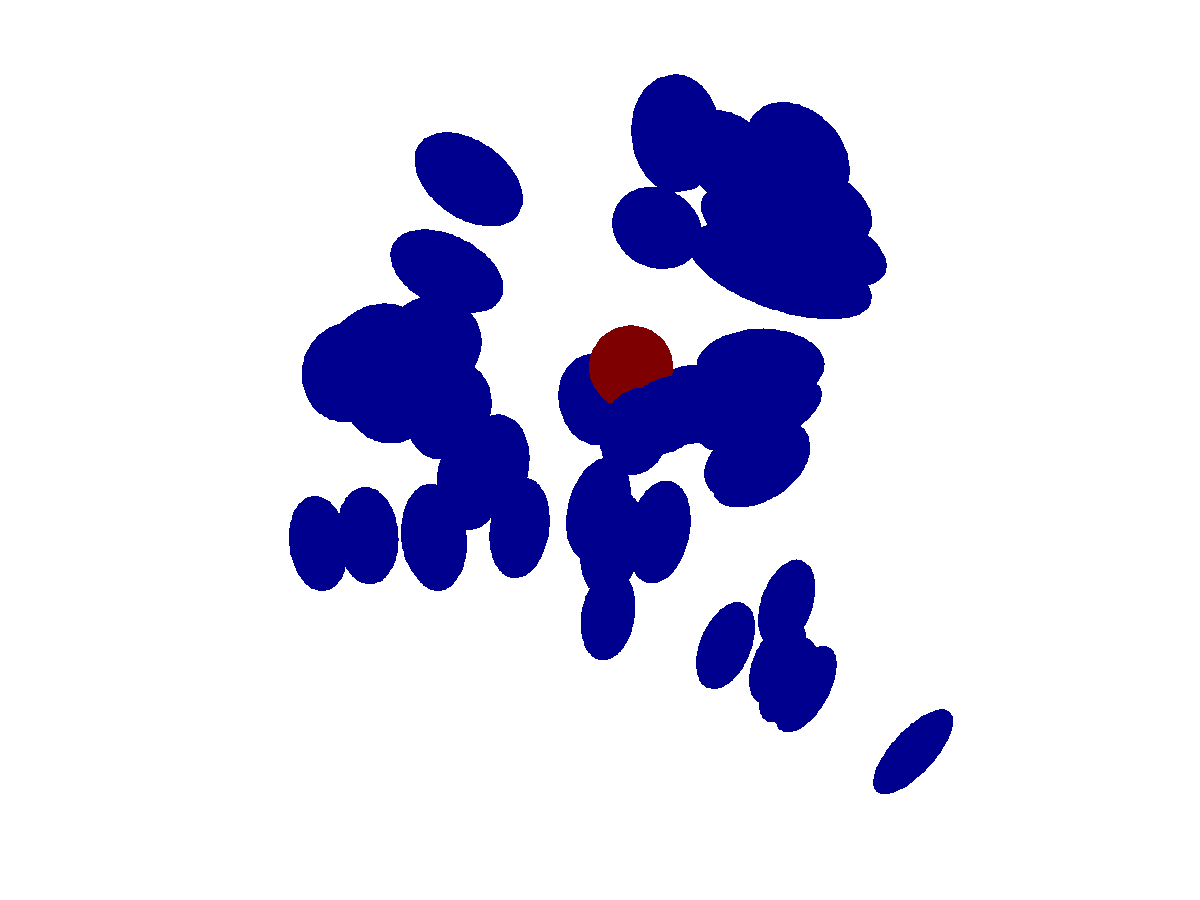}\\
\includegraphics[width=0.5\textwidth]{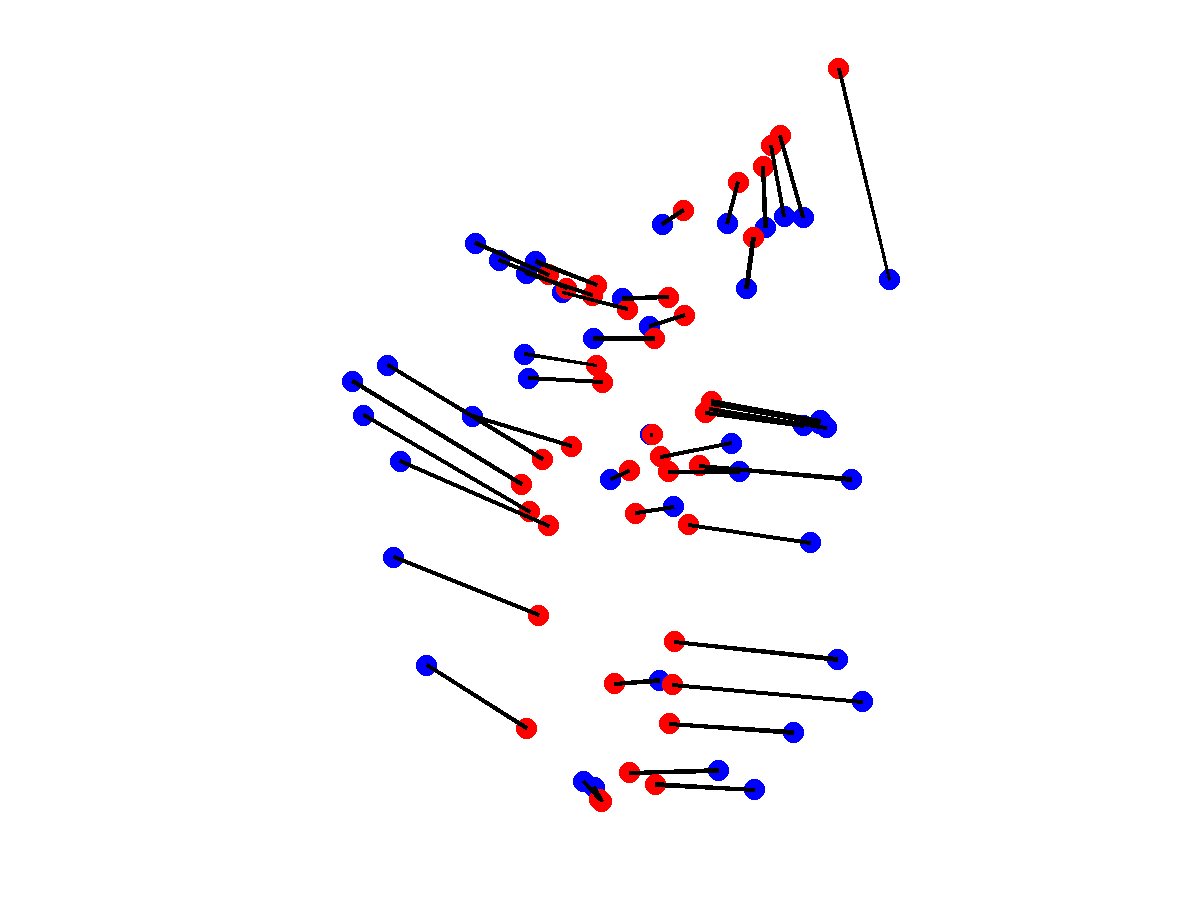}&
\includegraphics[width=0.5\textwidth]{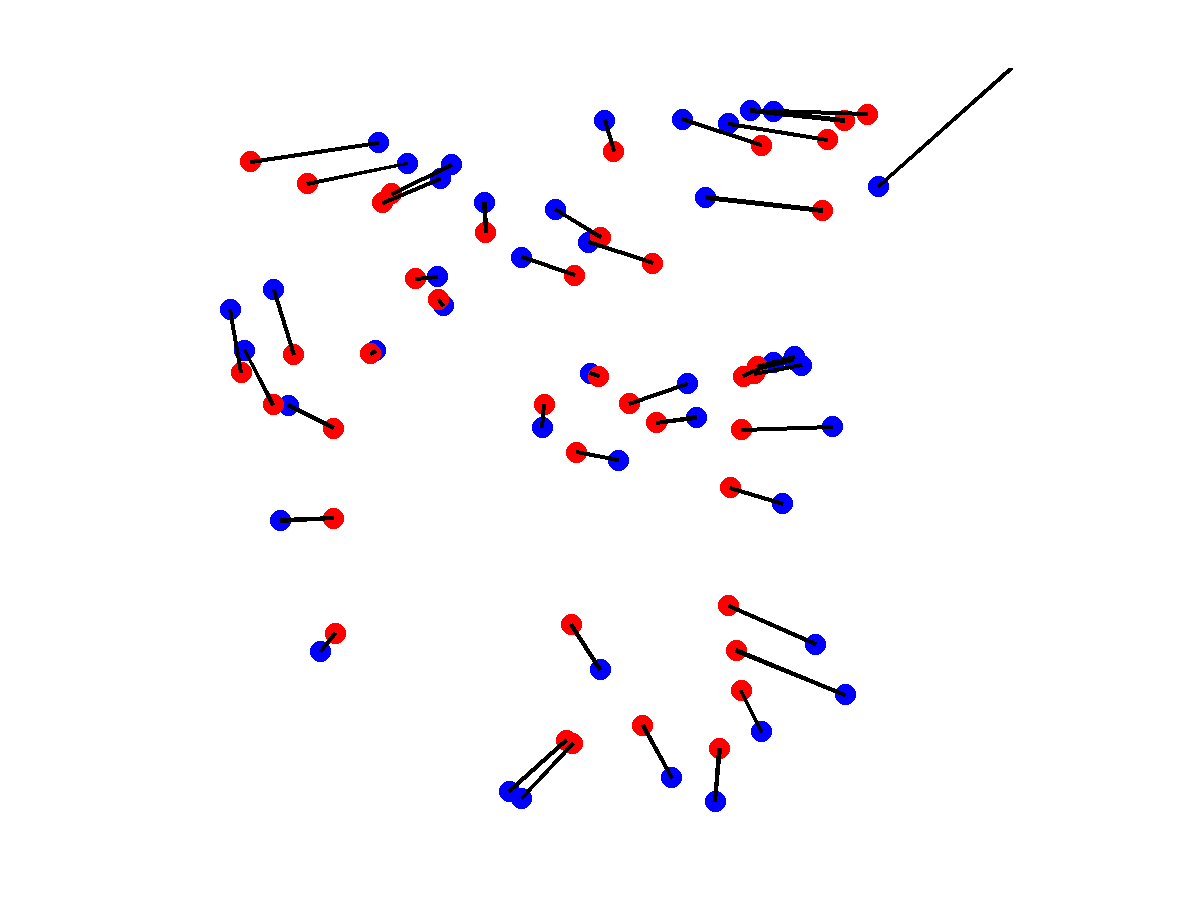}\\
\textcolor{blue}{$\bullet$} {\small original} &
\textcolor{red}{$\bullet$} {\small embedding} \\
\end{tabular}

\caption{\label{fig:normalized-isomap}
Locally isometric visualization for the Swiss roll with a rectangular
hole, embedded in $\dintri=2$ dimensions by Isomap. Top left: Isomap embedding with selected point $p$ (red), and its neighbors
(blue). Top right: locally isometric embedding. Middle left: Neighborhood of $p$ for the Isomap embedding. Middle right: Neighborhood of $p$ for the locally isometric embedding. Bottow left: Procrustes between the neighborhood of the $p$ for the Isomap embedding and the original manifold projected on $T_p\M$; dissimilarity measure: $D=0.21$. Bottom right: Procrustes between the locally isometric embedding and the original manifold; dissimilarity measure: $D=0.06$.}
\end{figure}

\begin{figure} 
\vspace{-1 in}
\hspace{-0.25 in}
\includegraphics[width=0.5\textwidth]{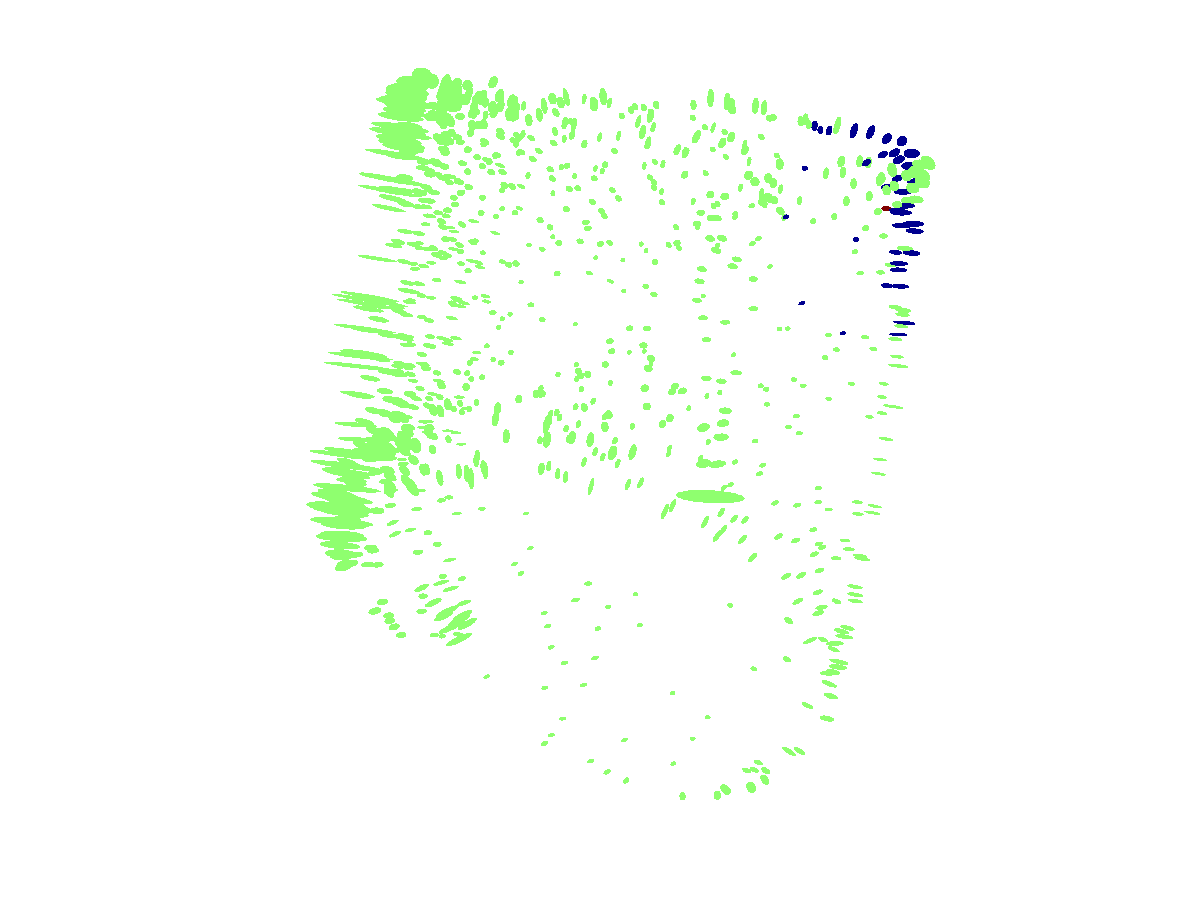}
\includegraphics[width=0.5\textwidth]{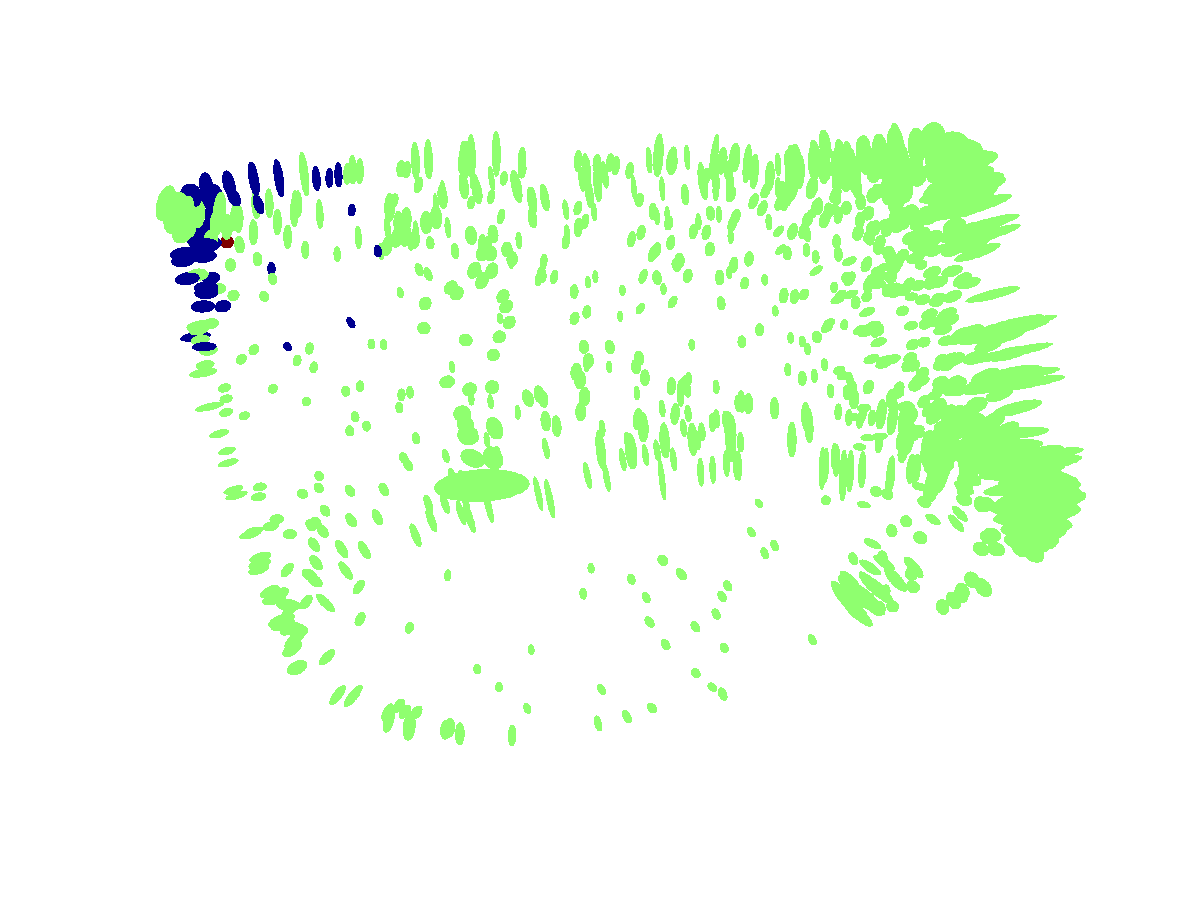}
\begin{tabular}{cc}
\reflectbox{\includegraphics[width=0.45\textwidth]{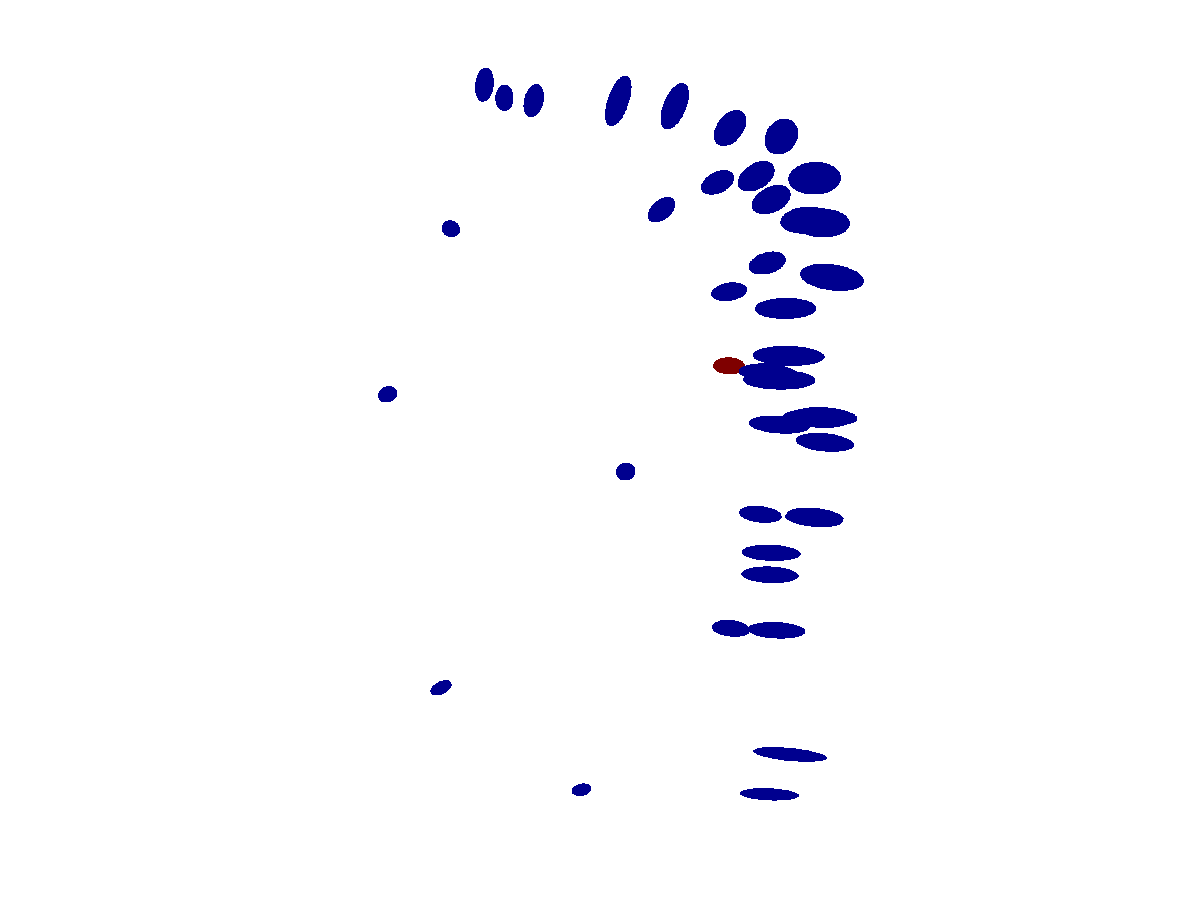}}&
\includegraphics[width=0.5\textwidth]{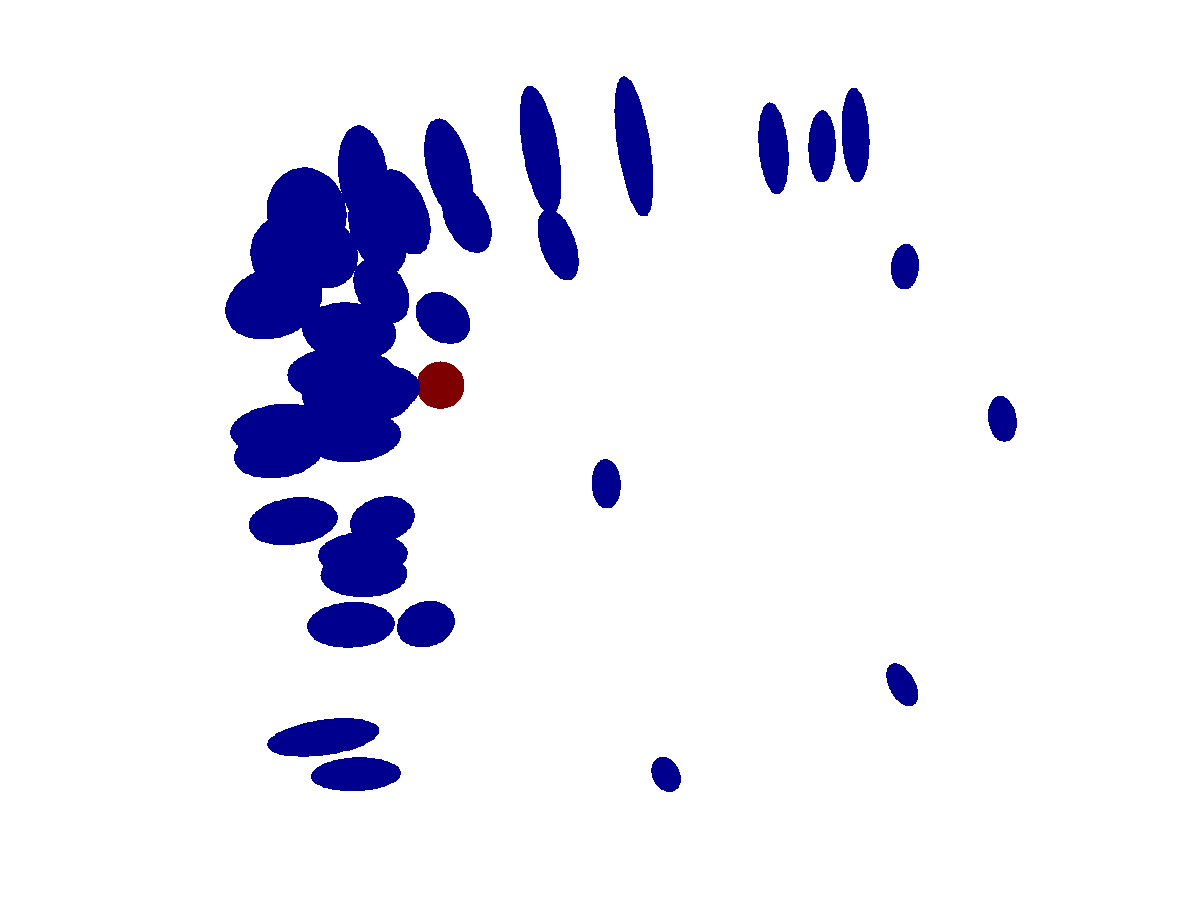}\\
\includegraphics[width=0.5\textwidth]{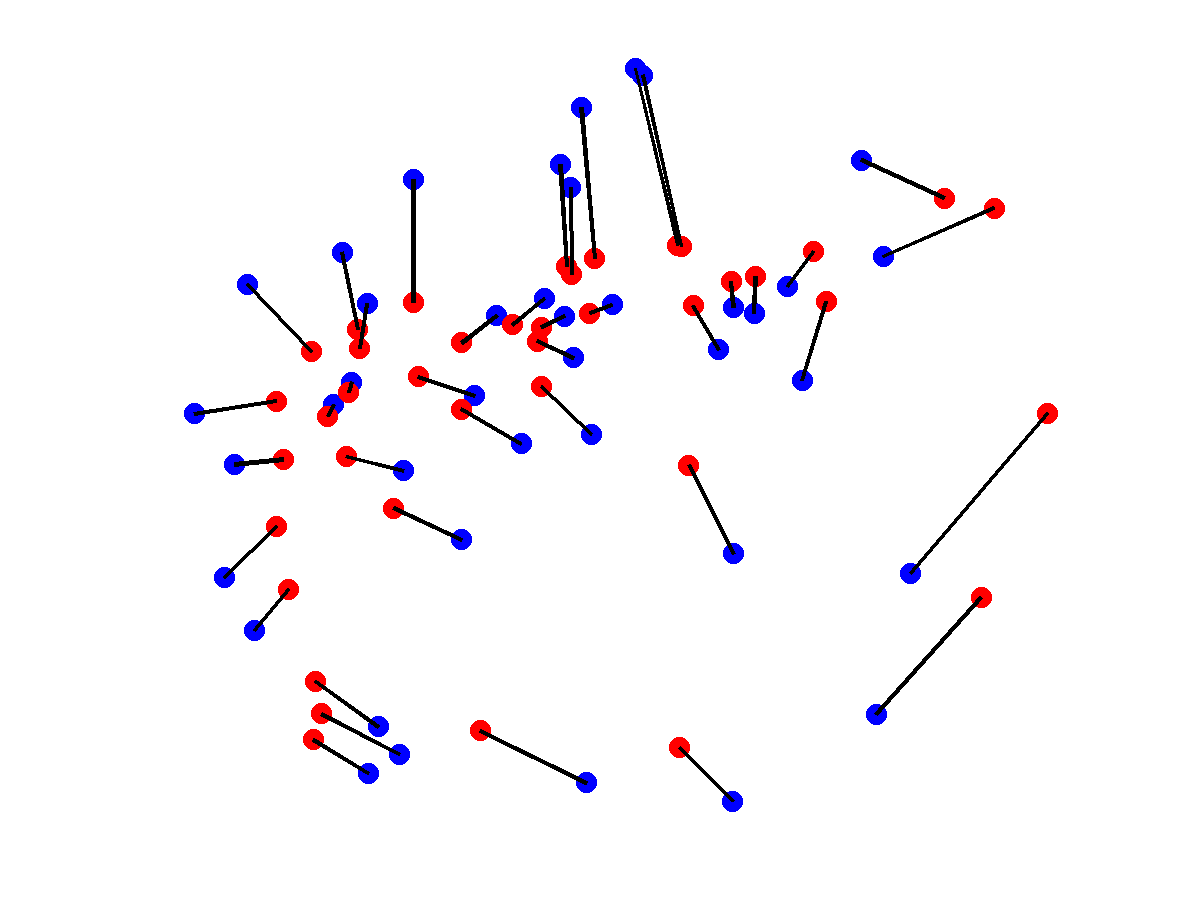}&
\includegraphics[width=0.5\textwidth]{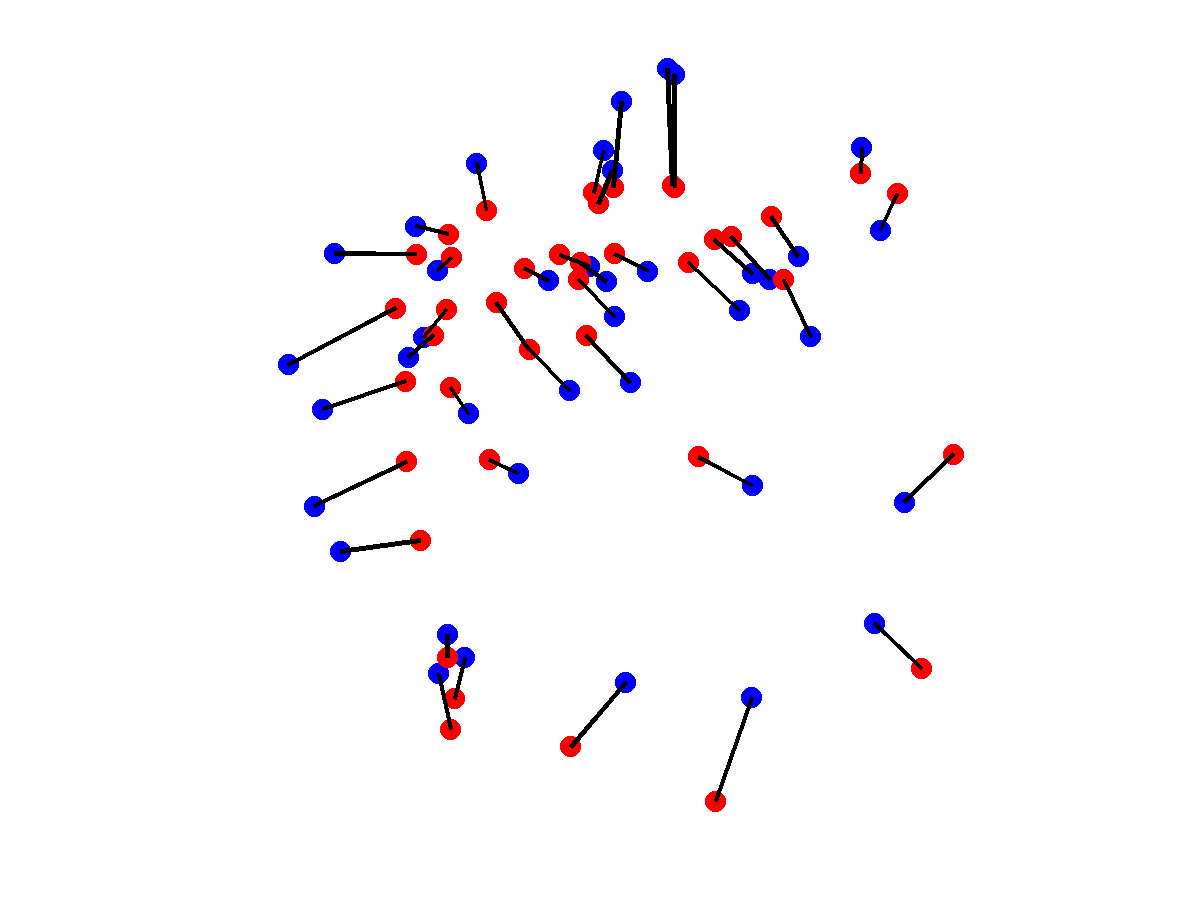}\\
\textcolor{blue}{$\bullet$} {\small original} &
\textcolor{red}{$\bullet$} {\small embedding} \\
\end{tabular}

\caption{\label{fig:normalized-le}
Locally isometric visualization for the Swiss roll with a rectangular
hole, embedded in $\dintri=2$ dimensions by Diffusion Maps ($\lambda = 1$). Top left: DM embedding with selected point $p$ (red), and its neighbors
(blue). Top right: locally isometric embedding. Middle left: Neighborhood of $p$ for the DM embedding. Middle right: Neighborhood of $p$ for the locally isometric embedding. Bottow left: Procrustes between the neighborhood of the $p$ for the DM embedding and the original manifold projected on $T_p\M$; dissimilarity measure: $D=0.10$. Bottom right: Procrustes between the locally isometric embedding and the original manifold; dissimilarity measure: $D=0.07$.}
\end{figure}

Figures \ref{fig:normalized-ltsa}, \ref{fig:normalized-isomap} and
\ref{fig:normalized-le} exemplify this procedure for the Swiss roll with a rectangular hole of Figure \ref{fig:ltsaSwiss} embedded respectively by LTSA, Isomap and Diffusion Maps. 
In these figures, we use the Procrustes method (\cite{GolRit09}) to align the original neighborhood of the chosen point $p$ with the same neighborhood in an embedding. The Procrustes method minimizes the sum of squared distances between corresponding points between all possible rotations, translations and isotropic scalings. 
The residual sum of squared distances is what we call the {\em Procrustes dissimilarity}. Its value is close to zero when the embedding is locally isometric around $p$. 

\subsection{Estimation of Geodesic Distances}
\label{sec:lengths}

The {\em geodesic distance} $d_\M(p,p')$ between two points
$p,p'\in\M$ is defined as the length of the shortest curve from $p$ to
$p'$ along manifold $\M$, which in our example is a half sphere of
radius 1. The geodesic distance $d$ being an intrinsic quantity, it
should evidently not change with the parametrization.

We performed the following numerical experiment. 
First, we sampled $\nsamp = 1000$ points uniformly on a half sphere. Second, we selected
two reference points $p,p'$ on the half sphere so that their geodesic distance would be $\pi/2$. We then proceeded to run three manifold learning algorithms on
$\dataset$, obtaining the Isomap, LTSA and DM embeddings. All the embeddings used the same 10-nearest neighborhood graph $G$. 

For each embedding, and for the original data, we calculated the naive
distance $||f_{\nsamp}(p)-f_{\nsamp}(p')||$. In the case of the original data, this
represents the straight line that connects $p$ and $p'$ and crosses
through the ambient space. For Isomap, $||f_{\nsamp}(p)-f_{\nsamp}(p')||$ should be
the best estimator of $d_\M(p,p')$, since Isomap embeds the data by
preserving geodesic distances using MDS. As for LTSA and DM, this
estimator has no particular meaning, since these algorithms are
derived from eigenvectors, which are defined up to a scale factor.

We also considered the {\em graph distance}, by which we mean the shortest path between the points in $G$, where the distance is given by $||f_{\nsamp}(q_i)-f_{\nsamp}(q_{i-1}')||$: 
\beq
{d}_\G(p,p')\;=\;\underset{\text{paths}}{\min} \{\sum_{i=1}^l
||f_{\nsamp}(q_i)-f_{\nsamp}(q_{i-1})||,
\,(q_0=p,\,q_1,\,q_2,\ldots \,q_l=p')\,
\mbox{\rm path in}\;\G\}.
\eeq
Note that although we used the same graph $\G$ to generate all the embeddings, the shortest path between points may be different in each embedding since the distances between nodes will generally not be preserved. 

The graph distance $d_\G$ is a good approximation for the geodesic
distance $d$ in the original data and in any isometric embedding, as it will closely follow the actual manifold rather then cross in the ambient space.

\begin{figure}
\begin{tabular}{ccc}
\includegraphics[trim=5cm 4cm 5cm 4cm,clip=true,width=0.33\textwidth]{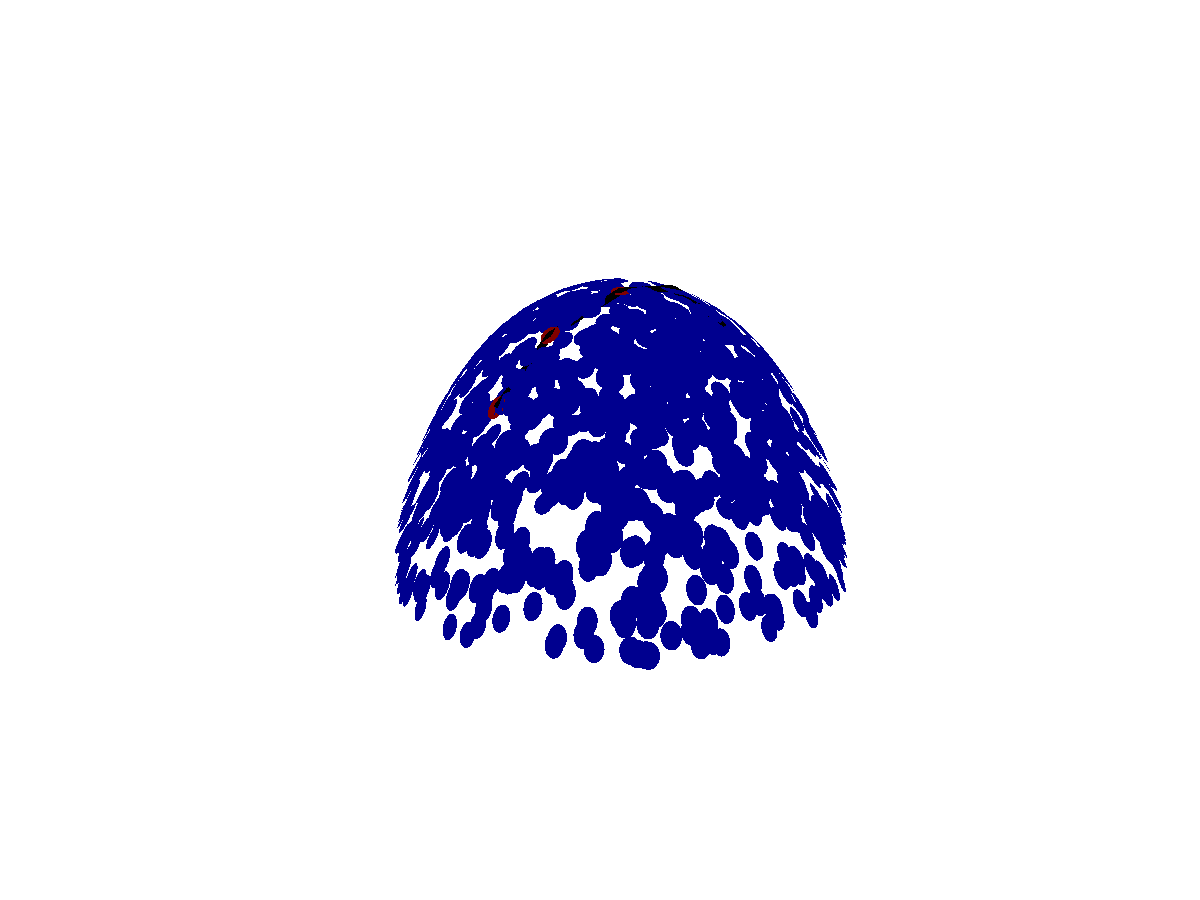} &
\includegraphics[trim=2cm 2cm 2cm 1cm,clip=true,width=0.33\textwidth]{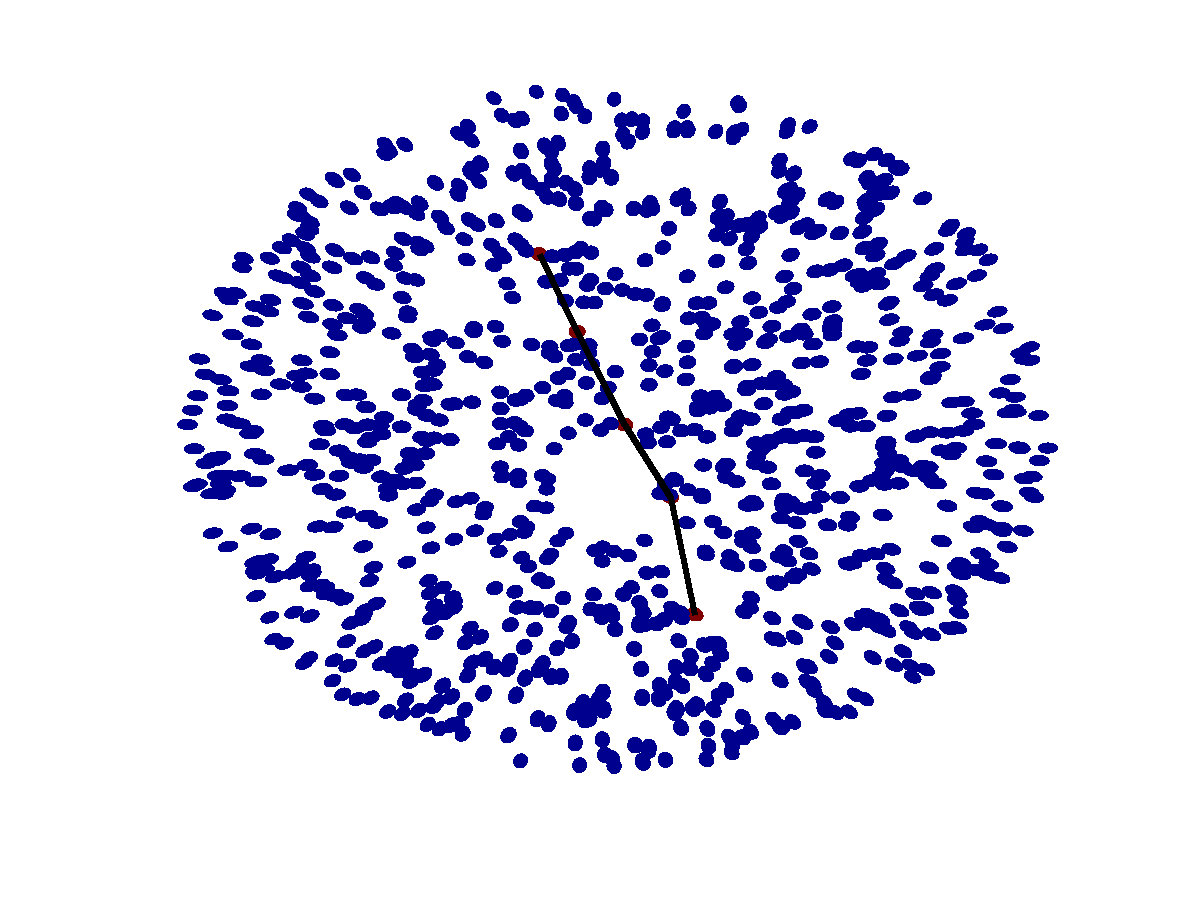} &
\includegraphics[trim=5cm 3cm 4cm 4cm,clip=true,width=0.33\textwidth]{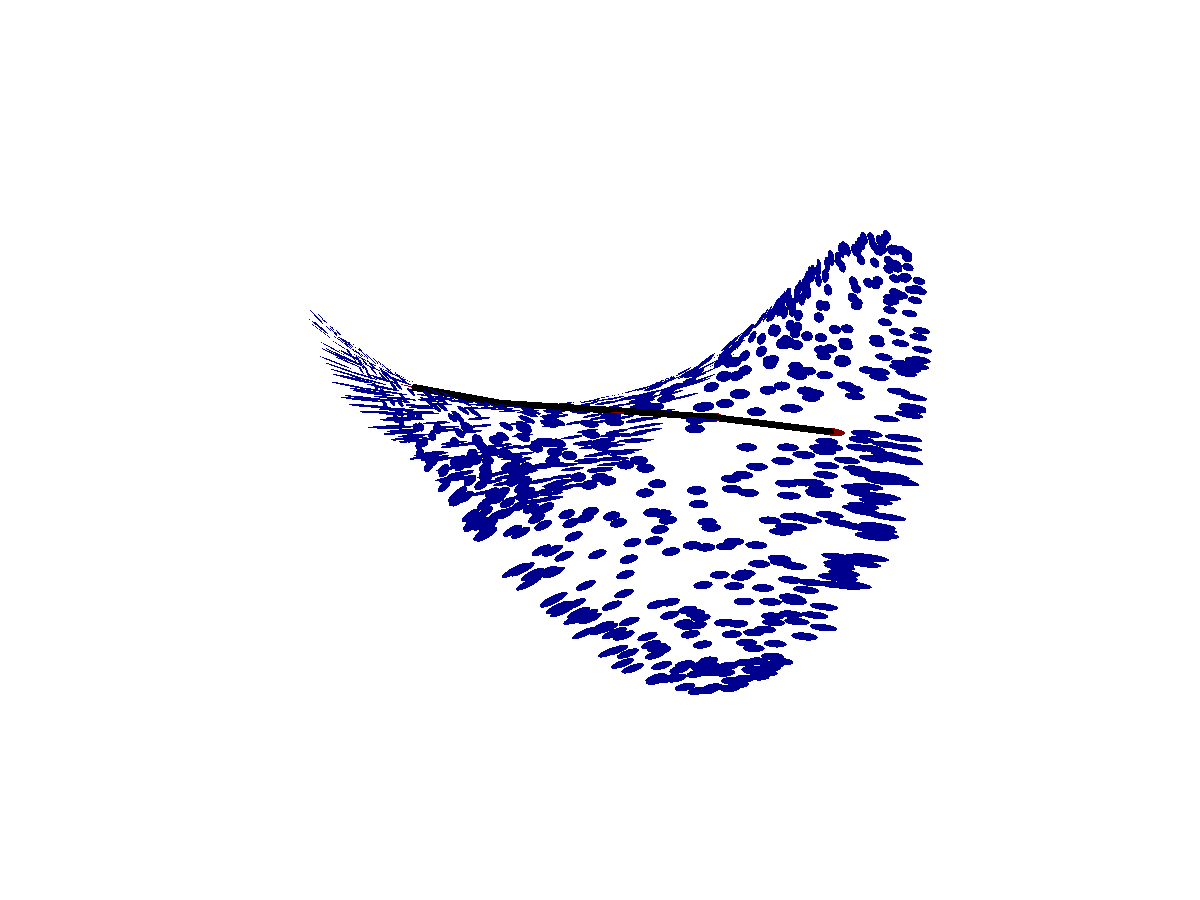}\\
(a) & (b) & (c) \\
\end{tabular}
\caption{\label{fig:half sphere}
Manifold and embeddings (black) used to compute the geodesic distance. Points that were part of the geodesic, including endpoints, are shown in red, while the path is shown in black. The LTSA embedding is not shown here: it is very similar to the Isomap. (a) Original manifold (b) Diffusion Maps (c) Isomap. 
}
\end{figure}

Finally, we computed the discrete minimizing geodesic as:
\beq \label{eq:geodesic-shortest-path}
\hat{d}_\M(p,p')\;=\;\underset{\text{paths}}{\min} \{\sum_{i=1}^l
\mathcal{H}(q_i,q_{i-1}),\,(q_0=p,\,q_1,\,q_2,\ldots \,q_l=p')\,
\mbox{\rm path in}\;\G\}
\eeq
where
\begin{eqnarray}
\mathcal{H}(q_i,q_{i-1})& =& \frac{1}{2}\sqrt{(f_{\nsamp}(q_i)-f_{\nsamp}(q_{i-1}))^t h_{\nsamp}(q_i)(f_{\nsamp}(q_i)-f_{\nsamp}(q_{i-1}))} \nonumber \\ 
& & +  \frac{1}{2}\sqrt{(f_{\nsamp}(q_i)-f_{\nsamp}(q_{i-1}))^t h_{\nsamp}(q_{i-1})(f_{\nsamp}(q_i)-f_{\nsamp}(q_{i-1}))}
\end{eqnarray}
is the discrete analog of the path-length formula \eqref{eq:l-curve} for the Voronoi tesselation of the space. By Voronoi tesselation, we mean the partition of the space into sets based on $\dataset$ such that each set consists of all points closest to a single point in $\dataset$ than any other. Figure \ref{fig:half sphere} shows the manifolds that we used in our
experiments, and Table \ref{tab:length-half sphere} displays the estimated distances.

\begin{table}
\begin{center}
\begin{tabular}{|c|c|c|c|c|}
\hline 
           &             & Shortest&     &\\
Embedding & $||f_{\nsamp}(p)-f_{\nsamp}(p')||$ & Path $d_\G$ & Metric $\hat{d}$ &
$\hat{d}$ Relative Error\\
\hline
Original data & 1.412 & 1.565 $\pm$ 0.003  & 1.582 $\pm$ 0.006 & 0.689 \%\\
Isomap $\dembed=2$ & 1.738 $\pm$ 0.027 & 1.646 $\pm$ 0.016 & 1.646 $\pm$ 0.029 & 4.755\%\\
LTSA $\dembed=2$ & 0.054 $\pm$ 0.001 & 0.051 $\pm$ 0.0001 & 1.658 $\pm$ 0.028 & 5.524\%\\
DM $\dembed=3$ & 0.204 $\pm$ 0.070 & 0.102 $\pm$ 0.001 & 1.576 $\pm$ 0.012 & 0.728\%\\
\hline
\end{tabular}
\caption{\label{tab:length-half sphere}
Distance estimates (mean and standard deviation) for a sample size of $n=2000$ points was used for all embeddings while the standard deviations were estimated by repeating the experiment 5 times. The relative errors in the last column were computed with respect to the true distance
$d=\pi/2\simeq$1.5708.
}
\end{center}
\end{table}
As expected, for the original data, $||p-p'||$ necessarily underestimates $d_\M$, while $d_\G$ is a very good approximation of $d_\M$, since it follows the manifold more closely. Meanwhile, the opposite is true for Isomap. The naive distance $||f_{\nsamp}(p)-f_{\nsamp}(p')||$ is close to the geodesic by construction, while $d_\G$ overestimates $d_\M$ since $d_\G \geq ||f_{\nsamp}(p)-f_{\nsamp}(p')||$ by the triangle inequality. Not surprisingly, for LTSA and Diffusion Maps, the estimates $||f_{\nsamp}(p)-f_{\nsamp}(p')||$ and $d_\G$ have no direct correspondence with the distances of the original data since these algorithms make no attempt at preserving absolute distances. 


However, the estimates $\hat{d}$ are
quite similar for all embedding algorithms, and they provide a good approximation for the true geodesic distance. It is interesting to note that $\hat{d}$ is the best estimate of the true geodesic distance even for the Isomap, whose focus is specifically to preserve geodesic distances. In fact, the only estimate that is better than $\hat{d}$ for any embedding is the graph distance on the original manifold.  


\subsection{Volume Estimation}
\label{sec:area}

The last set of our experiments demonstrates the use of the
Riemannian metric in estimating two-dimensional volumes: areas. We used an experimental
procedure similar to the case of geodesic distances, in that we
created a two-dimensional manifold, and selected a set $W$ on it. We then estimated the area of this set by generating a sample from the manifold, embedding the sample, and computing the area in the embedding space using a discrete form of \eqref{eq:Vol}.

One extra step is required when computing areas that was optional when computing distances: we need to construct coordinate chart(s) to represent the area of interest. Indeed, to make sense of the Euclidean volume element $dx^1\ldots dx^\dintri$, we need to work in $\rrr^\dintri$. Specifically, we resort to the idea expressed at the end of Section \ref{sub:H}, which is to project the embedding on its tangent at the point $p$ around which we wish to compute $dx^1\ldots dx^\dintri$. This tangent plane $T_{f(p)}f(\M)$ is easily identified from the SVD of $h_{\nsamp}(p)$ and its singular vectors with non-zero singular values. It is then straightforward to use the projection $\Pi$ of an open neighborhood $f(U)$ of $f(p)$ onto $T_{f(p)}f(\M)$ to define the coordinate chart $(U,x=\Pi\circ f)$ around $p$. Since this is a new chart, we need to recompute the embedding metric ${h_n}$ for it. 

By performing a tessellation of $(U,x=\Pi\circ f)$ (we use the Voronoi tesselation for simplicity), we are now in position to compute $\triangle x^1\ldots \triangle x^\dintri$ around $p$ and multiply it by $\sqrt{\det{({h_{\nsamp}})}}$ to obtain $\triangle \text{Vol} \simeq d\text{Vol}$. Summing over all points of the desired set gives the appropriate estimator:
\beq \label{eq:hVol}
\hat{\text{Vol}}(W) = \sum_{p\in W} \sqrt{\det{({h_{\nsamp}(p)})}}\triangle x^1(p)\ldots \triangle x^\dintri(p)\, .
\eeq


\begin{figure}

\begin{center}

\begin{tabular}{ccc} 

\includegraphics[trim=2cm 2cm 2cm 2cm,clip=true,width=0.33\textwidth]{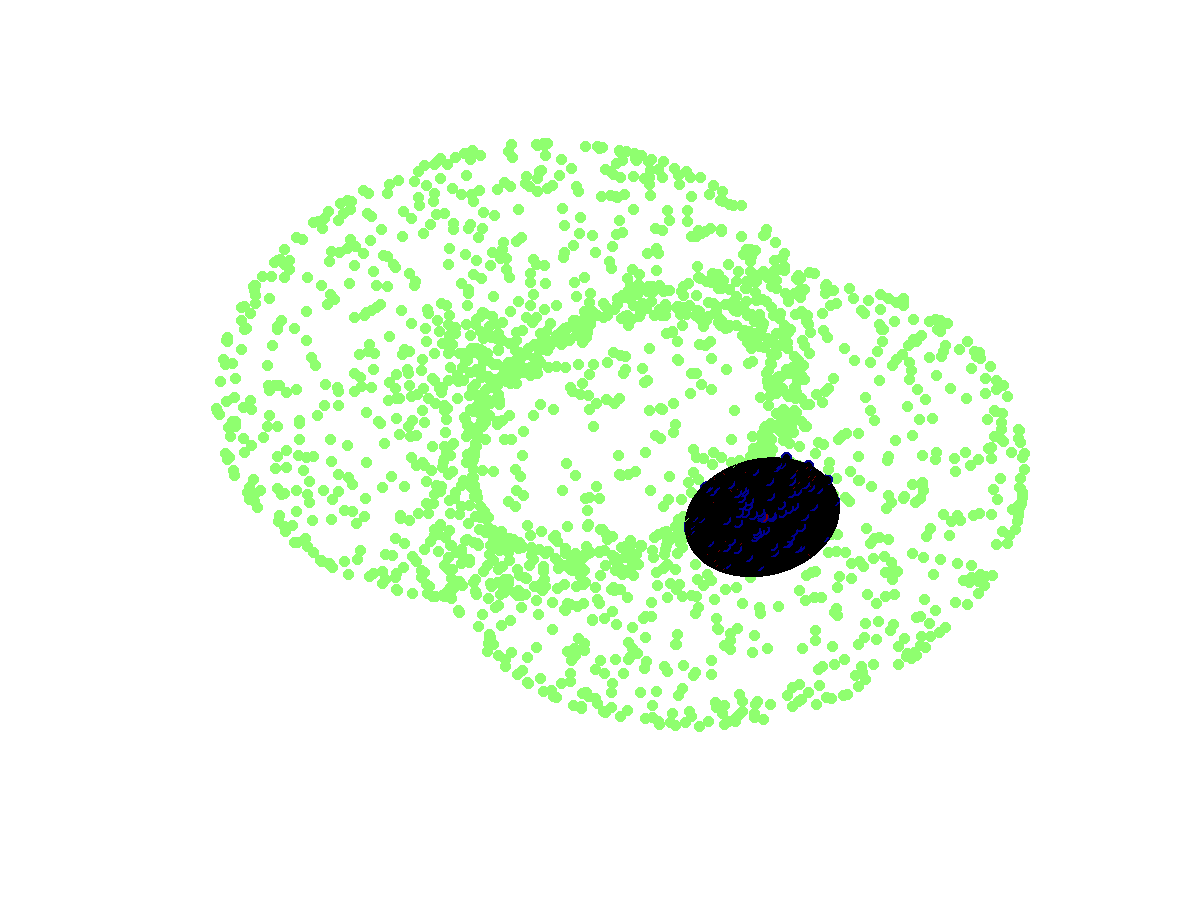} &
\includegraphics[trim=4cm 4cm 4cm 3cm,clip=true,width=0.33\textwidth]{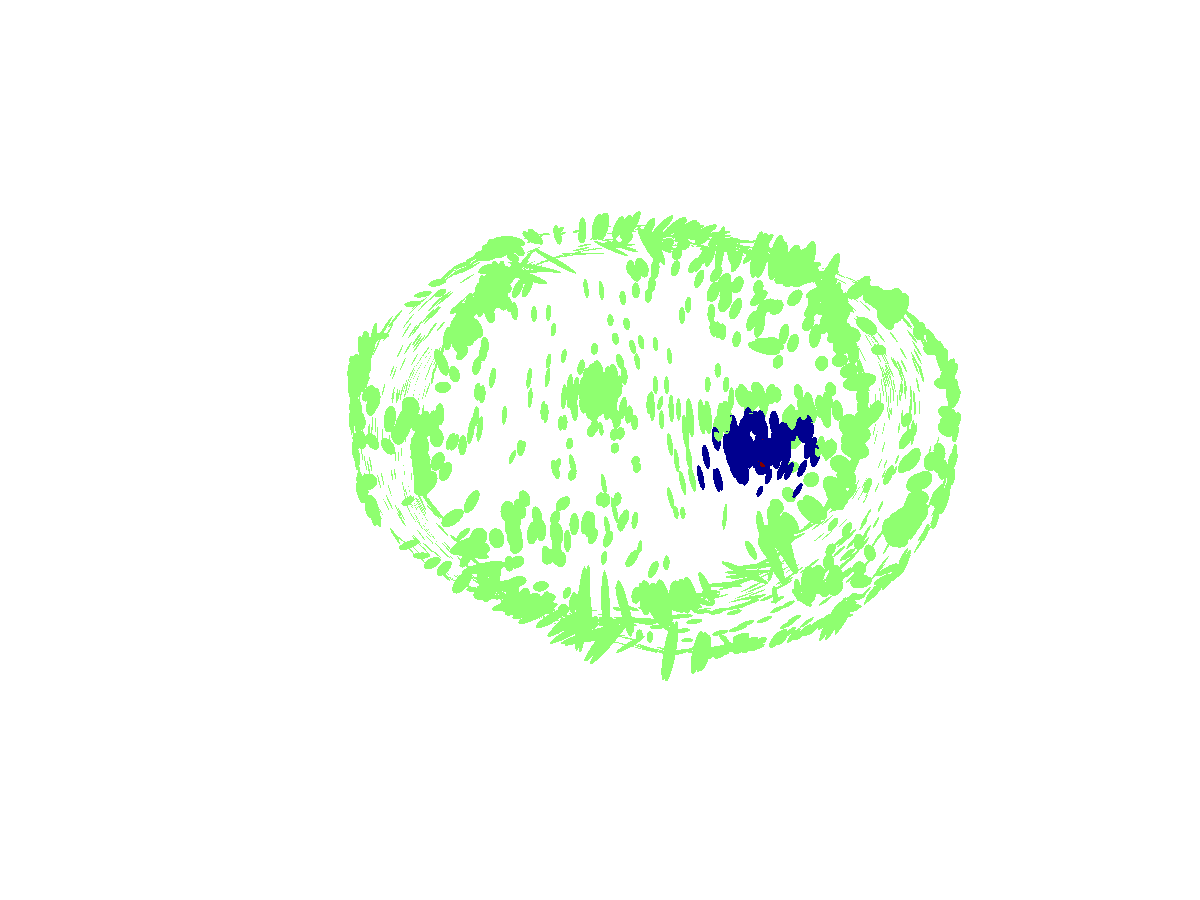} &
\includegraphics[trim=1cm 1cm 1cm 1cm,clip=true,width=0.33\textwidth]{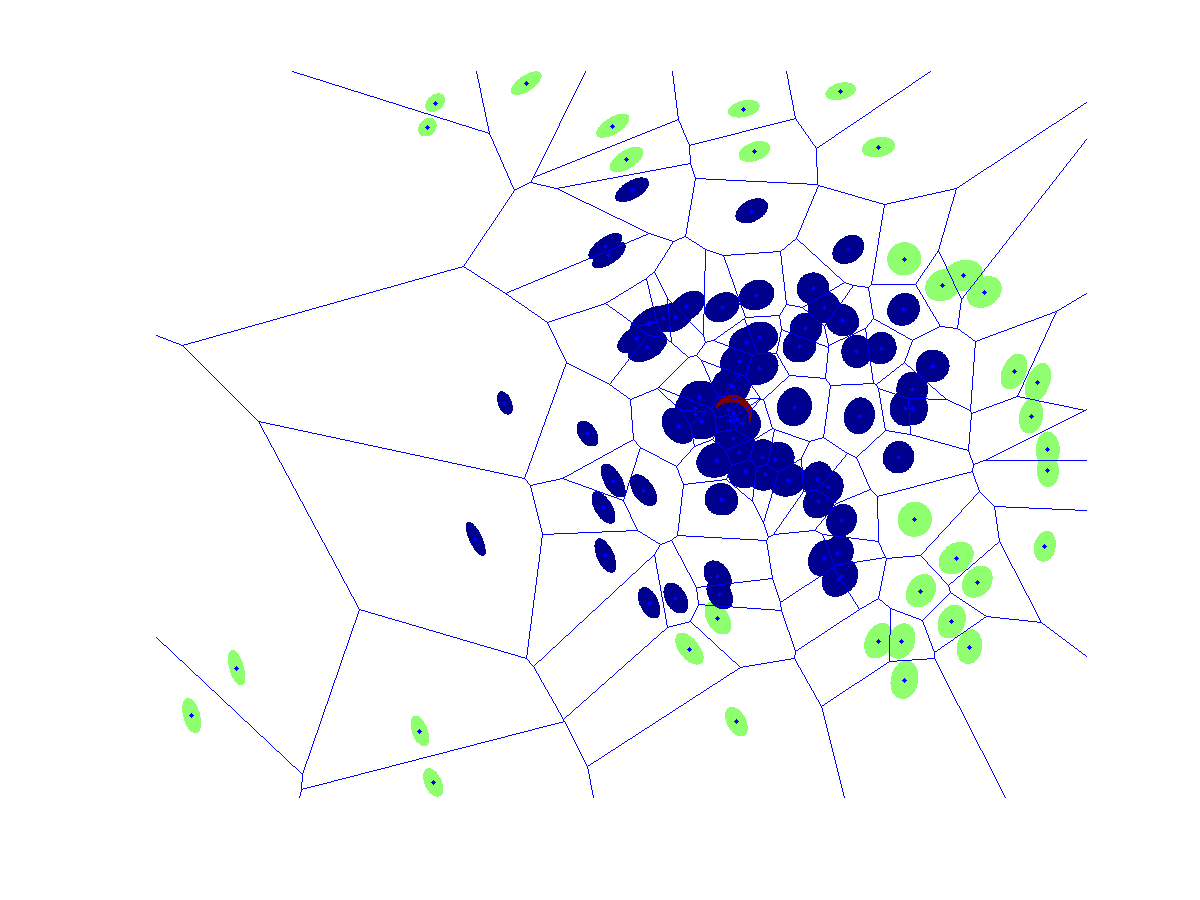}\\
(a) & (b) & (c) \\

\end{tabular}

\caption{\label{fig:area-dumbbell}
(a) Manifold along with $W$, the computed area in black. (b) Diffusion Maps ($\lambda = 1$) embedding with embedding metric $h$. (c) A locally isometric coordinate chart constructed from the Diffusion Maps along with the Voronoi tessellation. For Figures (b) and (c), the point at the center of $W$ is in red, the other points in $W$ are in blue and the points not in $W$ are in green.
The sample size is $\nsamp=1000$.}
\end{center}

\end{figure}

\begin{table}
\begin{center}
\begin{tabular}{|c|c|c|c|} 
\hline 
Embedding & Naive Area of $W$ & $\hat{\text{Vol}}(W)$ & $\hat{\text{Vol}}(W)$ Relative Error\\
\hline
Original data & 2.63 $\pm$ 0.10$^{\dagger}$ & 2.70 $\pm$ 0.10 & 2.90\%\\
Isomap & 6.53 $\pm$ 0.34$^{\dagger}$ & 2.74 $\pm$ 0.12 & 3.80\%\\
LTSA & 8.52e-4 $\pm$ 2.49e-4 & 2.70 $\pm$ 0.10 & 2.90 \%\\
DM & 6.70e-4 $\pm$ 0.464e-04$^{\dagger}$ &  2.62 $\pm$ 0.13 & 4.35 \%\\
\hline

\end{tabular}
\caption{\label{tab:Area} Estimates of the volume of $W$ on the hourglass depicted in Figure \ref{fig:area-dumbbell}, based on 1000 sampled points. The experiment was repeated five times to obtain a variance for the estimators. $^{\dagger}$The naive area/volume estimator is obtained by projecting the manifold or embedding on $T_p\M$ and $T_{f(p)}f(\M)$, respectively. This requires manually specifying the correct tangent planes, except for LTSA, which already estimates $T_{f(p)}f(\M)$. Similarly to LTSA, $\hat{\text{Vol}}(W)$ is constructed so that the embedding is automatically projected on  $T_{f(p)}f(\M)$. Here, the true area is $2.658$ } 
\end{center}
\end{table}

Table \ref{tab:Area} reports the results of our comparison of the performance of $\hat{\text{Vol}}(W)$, described in \ref{eq:hVol} , and a ``naive'' volume estimator that computes the area on the Voronoi tessellation once the manifold is projected onto the tangent plane. We find that $\hat{\text{Vol}}(W)$ performs better for all embeddings, as well as for the original data. The latter is explained by the fact that when we project the set $W$ onto the tangent plane $T_{f(p)}f(\M)$, we induce a fair amount of distortion, and the naive estimator has no way of correcting for it. 

The relative error for LTSA is similar 
to that for the
original data and larger than for the other methods. One possible
reason for this is the error in estimating the tangent plane ${\cal
T}_p\M$, which, in the case of these two methods, is done by local
PCA. 

\comment{The analysis of the asymptotic error of the tangent plane
estimation via SVD on local data was done \cite{} \mmp{ in Aswani,
Bickel and Tomlin 2010, Regression on manifolds and Singer and Wu
2011, Vector diffusion map and connection Laplacian. FINISH THIS}}


\section{Conclusion and Discussion}
\label{sec:discussion}


In this article, we have described a new method for preserving the important geometrical information in a data manifold embedded using any embedding algorithm. We showed that the Laplace-Beltrami operator can be used to augment {\em any} reasonable embedding so as to allow for the correct computation of geometric values of interest in the embedding's own coordinates.

Specifically, we showed that the Laplace-Beltrami operator allows us to recover a Riemannian manifold $(\mathcal{M},g)$ from the data and express the Riemannian metric $g$ in any desired coordinate system. We first described how to obtain the Riemannian metric from the mathematical, algorithmic and statistical points of view. Then, we proceeded to describe how, for any mapping produced by an existing manifold learning algorithm, we can estimate the Riemannian metric $g$ in the new data coordinates, which makes the geometrical quantities like distances and angles of the mapped data (approximately) equal to their original values, in the raw data. We conducted several experiments to demonstrate the usefulness of our method.

Our work departs from the standard manifold learning paradigm. While existing manifold learning algorithms, when faced with the impossibility of mapping curved manifolds to Euclidean space, choose to focus on
distances, angles, or specific properties of local neighborhoods and thereby settle for trade-offs, our method allows for dimensionality reduction without sacrificing {\em any} of these data properties. Of course, this entails recovering and storing more information than the coordinates alone. The information stored under the Metric Learning algorithm is of order $\dembed^2$ per point, while the coordinates only require $\dembed$ values per point. 

Our method essentially frees users to select their preferred embedding algorithm based on considerations
unrelated to the geometric recovery; the metric learning algorithm then
obtains the Riemannian metric corresponding to these coordinates through the Laplace-Beltrami operator.
Once these are obtained, the distances, angles, and other geometric
quantities can be estimated in the embedded manifold by standard
manifold calculus. These quantities will preserve their values from
the original data and are thus
embedding-invariant in the limit of $\nsamp \rightarrow \infty$.

Of course, not everyone agrees that the original geometry is interesting in and of itself; sometimes, it should be discarded in favor of a new geometry that better highlights the features of the data that are important for a given task. For example, clustering algorithms stress the importance of the dissimilarity (distance) between different clusters regardless of what the original geometry dictates. This is in fact one of the arguments advanced by \cite{NadLafCoi05} in support of spectral clustering which pulls points towards regions of high density.

Even in situations where the new geometry is considered more important, however, understanding the relationship between the original and the new geometry using Metric Learning - and, in particular, the pullback metric \cite{Lee03} - could be of value and offer further insight. Indeed, while we explained in Section \ref{sec:App} how the embedding metric $h$ can be used to infer how the original geometry was affected by the map $f$, we note at this juncture that the pullback metric, i.e. the geometry of $(f(\M),\iembed)$ pulled back to $\M$ by the map $f$, can offer interesting insight into the effect of the transformation/embedding.\footnote{One caveat to this idea is that, in the case where $\dhigh >> 1$, computing the pullback will not be practical and the pushforward will remain the best approach to study the effect of the map $f$. It is for the case where $\dhigh$ is not too large and $\dhigh \sim \dembed$ that the pullback may be a useful tool.} In fact, this idea has already been considered by \cite{Burges99} in the case of kernel methods where one can compute the pullback metric directly from the definition of the kernel used. In the framework of Metric Learning, this can be extended to any transformation of the data that defines an embedding. 






			    


\comment{
\section*{Acknowledgements}
We thank Jack Lee for many fruitful discussions on the topic of
manifolds and Riemannian geometry
This research was partly supported by NSF awards IIS-0313339 and IIS-0535100.
}

\bibliography{RMetric-arxiv}

\begin{thebibliography}{30}
\providecommand{\natexlab}[1]{#1}
\providecommand{\url}[1]{\texttt{#1}}
\expandafter\ifx\csname urlstyle\endcsname\relax
  \providecommand{\doi}[1]{doi: #1}\else
  \providecommand{\doi}{doi: \begingroup \urlstyle{rm}\Url}\fi

\bibitem[Behmardi and Raich(2010)]{BehRai10}
B.~Behmardi and R.~Raich.
\newblock Isometric correction for manifold learning.
\newblock In \emph{AAAI Symposium on Manifold Learning}, 2010.

\bibitem[Belkin and Niyogi(2002)]{BelNiy02}
M.~Belkin and P.~Niyogi.
\newblock Laplacian eigenmaps for dimensionality reduction and data
  representation.
\newblock \emph{Neural Computation}, 15:\penalty0 1373--1396, 2002.

\bibitem[Belkin and Niyogi(2007)]{belniy07}
M.~Belkin and P.~Niyogi.
\newblock Convergence of laplacians eigenmaps.
\newblock In \emph{Advances in Neural Information Processing Systems (NIPS)},
  2007.

\bibitem[Belkin et~al.(2009)Belkin, Sun, and Wang]{BelSunWan08}
M.~Belkin, J.~Sun, and Y.~Wang.
\newblock Constructing laplace operator from point clouds in rd.
\newblock In \emph{ACM-SIAM Symposium on Discrete Algorithms}, pages
  1031--1040, 2009.

\bibitem[Ben-Israel and Greville(2003)]{BenGre03}
A.~Ben-Israel and T.~N.~E. Greville.
\newblock \emph{Generalized inverses: {T}heory and applications}.
\newblock Springer, New York, 2003.

\bibitem[Bernstein et~al.(2000)Bernstein, {deSilva}, Langford, and
  Tennenbaum]{BernEtAL:00}
M.~Bernstein, V.~{deSilva}, J.~C. Langford, and J.~Tennenbaum.
\newblock Graph approximations to geodesics on embedded manifolds, 2000.
\newblock URL \url{http://web.mit.edu/cocosci/isomap/BdSLT.pdf}.

\bibitem[Borg and Groenen(2005)]{BorGro05}
I.~Borg and P.~Groenen.
\newblock \emph{Modern Multidimensional Scaling: Theory and Applications}.
\newblock Springer-Verlag, 2nd edition, 2005.

\bibitem[Burges(1999)]{Burges99}
C.~J.~C. Burges.
\newblock Geometry and invariance in kernel based methods.
\newblock \emph{Advances in Kernel Methods - Support Vector Learning}, 1999.

\bibitem[Coifman and Lafon(2006)]{CoiLaf06}
R.~R. Coifman and S.~Lafon.
\newblock Diffusion maps.
\newblock \emph{Applied and Computational Harmonic Analysis}, 21\penalty0
  (1):\penalty0 6--30, 2006.

\bibitem[Dreisigmeyer and Kirby(2007, retrieved June 2010)]{DreKir07}
D.~W. Dreisigmeyer and M.~Kirby.
\newblock A pseudo-isometric embedding algorithm, 2007, retrieved June 2010.
\newblock URL \url{http://www.math.colostate.edu/~thompson/whit_embed.pdf}.

\bibitem[Gin\'{e} and Koltchinskii(2006)]{GinKol06}
E.~Gin\'{e} and V.~Koltchinskii.
\newblock Empirical {G}raph {L}aplacian {A}pproximation of {L}aplace-{B}eltrami
  {O}perators: {L}arge {S}ample results.
\newblock \emph{High Dimensional Probability}, pages 238--259, 2006.

\bibitem[Goldberg and Ritov(2009)]{GolRit09}
Y.~Goldberg and Y.~Ritov.
\newblock {Local procrustes for manifold embedding: a measure of embedding
  quality and embedding algorithms}.
\newblock \emph{Machine Learning}, {77}\penalty0 ({1}):\penalty0 {1--25}, 2009.

\bibitem[Goldberg et~al.(2008)Goldberg, Zakai, Kushnir, and
  Ritov]{GolZakKusRit08}
Y.~Goldberg, A.~Zakai, D.~Kushnir, and Y.~Ritov.
\newblock {Manifold Learning: The Price of Normalization}.
\newblock \emph{Journal of Machine Learning Research}, {9}:\penalty0
  {1909--1939}, 2008.

\bibitem[Hein et~al.(2007)Hein, Audibert, and von Luxburg]{HeiAudLux07}
M.~Hein, J.-Y. Audibert, and U.~von Luxburg.
\newblock Graph {L}aplacians and their {C}onvergence on {R}andom {N}eighborhood
  {G}raphs.
\newblock \emph{Journal of Machine Learning Research}, 8:\penalty0 1325--1368,
  2007.

\bibitem[Lee(1997)]{Lee97}
J.~M. Lee.
\newblock \emph{Riemannian {M}anifolds: {A}n {I}ntroduction to {C}urvature}.
\newblock Springer, New York, 1997.

\bibitem[Lee(2003)]{Lee03}
J.~M. Lee.
\newblock \emph{Introduction to {S}mooth {M}anifolds}.
\newblock Springer, New York, 2003.

\bibitem[Nadler et~al.(2006)Nadler, Lafon, and Coifman]{NadLafCoi05}
B.~Nadler, S.~Lafon, and R.~R. Coifman.
\newblock Diffusion maps, spectral clustering and eigenfunctions of
  fokker-planck operators.
\newblock In \emph{Advances in Neural Information Processing Systems (NIPS)},
  2006.

\bibitem[Nash(1956)]{Nas56}
J.~Nash.
\newblock The imbedding problem for {R}iemannian manifolds.
\newblock \emph{Annals of Mathematics}, 63:\penalty0 20--63, 1956.

\bibitem[Ram et~al.(2009)Ram, Lee, March, and Gray]{ram2010ltaps}
P.~Ram, D.~Lee, W.~March, and A.~G. Gray.
\newblock Linear-time algorithms for pairwise statistical problems.
\newblock In \emph{Advances in Neural Information Processing Systems (NIPS)},
  2009.

\bibitem[Rosenberg(1997)]{Ros97}
S.~Rosenberg.
\newblock \emph{The {L}aplacian on a {R}iemannian {M}anifold}.
\newblock Cambridge University Press, 1997.

\bibitem[Saul and Roweis(2003)]{SauRow03}
L.~Saul and S.~Roweis.
\newblock Think globally, fit locally: unsupervised learning of low dimensional
  manifold.
\newblock \emph{Journal of Machine Learning Research}, 4:\penalty0 119--155,
  2003.

\bibitem[Snyder(1987)]{Syd87}
J.~P. Snyder.
\newblock \emph{Map Projections: A Working Manual}.
\newblock United States Government Printing, 1987.

\bibitem[Tenenbaum et~al.(2000)Tenenbaum, deSilva, and Langford]{TenDeS00}
J.~Tenenbaum, V.~deSilva, and J.~C. Langford.
\newblock A global geometric framework for nonlinear dimensionality reduction.
\newblock \emph{Science}, 290:\penalty0 2319--2323, 2000.

\bibitem[Ting et~al.(2010)Ting, Huang, and Jordan]{TingHJ10}
D.~Ting, L~Huang, and M.~I. Jordan.
\newblock An analysis of the convergence of graph laplacians.
\newblock In \emph{International Conference on Machine Learning}, pages
  1079--1086, 2010.

\bibitem[von Luxburg et~al.(2008)von Luxburg, Belkin, and
  Bousquet]{LuxBelBou08}
U.~von Luxburg, M.~Belkin, and O.~Bousquet.
\newblock Consistency of spectral clustering.
\newblock \emph{Annals of Statistics}, 36(2):\penalty0 555--585, 2008.

\bibitem[Weinberger and Saul(2006)]{weinberger06unsupervised}
K.Q. Weinberger and L.K. Saul.
\newblock Unsupervised learning of image manifolds by semidefinite programming.
\newblock \emph{International Journal of Computer Vision}, 70:\penalty0 77--90,
  2006.

\bibitem[Wittman(2005, retrieved 2010)]{Wit05}
T.~Wittman.
\newblock Manifold learning matlab demo, 2005, retrieved 2010.
\newblock URL \url{http://www.math.umn.edu/\textasciitilde wittman/mani/}.

\bibitem[Zha and Zhang(2003)]{ZhaZ:03}
H.~Zha and Z.~Zhang.
\newblock Isometric embedding and continuum isomap.
\newblock In \emph{International Conference on Machine Learning}, pages
  864--871, 2003.

\bibitem[Zhang and Zha(2004)]{ZhangZ:04}
Z.~Zhang and H.~Zha.
\newblock Principal manifolds and nonlinear dimensionality reduction via
  tangent space alignment.
\newblock \emph{Society for Industrial and Applied Mathematics Journal of
  Scientific Computing}, 26\penalty0 (1):\penalty0 313--338, 2004.

\bibitem[Zhou and Belkin(2011)]{ZhouBelkin11}
X.~Zhou and M.~Belkin.
\newblock Semi-supervised learning by higher order regularization.
\newblock In \emph{The 14th International Conference on Artificial Intelligence
  and Statistics}, 2011.

\end{thebibliography}


\end{document}